\documentclass[a4paper,fleqn]{cas-sc}

\usepackage[numbers]{natbib}
\usepackage[ruled,vlined]{algorithm2e}
\usepackage{float}
\usepackage{subfigure}

\def\tsc#1{\csdef{#1}{\textsc{\lowercase{#1}}\xspace}}
\tsc{WGM}
\tsc{QE}
\tsc{EP}
\tsc{PMS}
\tsc{BEC}
\tsc{DE}

\begin{document}
	
\let\WriteBookmarks\relax
\def\floatpagepagefraction{1}
\def\textpagefraction{.001}
\shorttitle{Three-way Imbalanced Learning based on Fuzzy Twin SVM}
\shortauthors{Wanting Cai et~al.}
\let\printorcid\relax

\title [mode = title]{Three-way Imbalanced Learning based on Fuzzy Twin SVM}                      
\author[1]{Wanting Cai}[style=chinese]
\ead{cwanting@hnu.edu.cn}
\address[1]{School of Mathematics, Hunan University, Changsha, Hunan 410082, P.R. China}

\author[1]{Mingjie Cai}[style=chinese]
\ead{cmjlong@163.com}
\cormark[1]

\author[1]{Qingguo Li}[style=chinese]
\ead{liqingguoli@aliyun.com}

\author[1]{Qiong Liu}[style=chinese]
\ead{qliu6666@163.com}

\cortext[cor1]{Corresponding author}

\begin{abstract}
Three-way decision (3WD) is a powerful tool for granular computing to deal with uncertain data, commonly used in information systems, decision-making, and medical care. Three-way decision gets much research in traditional rough set models. However, three-way decision is rarely combined with the currently popular field of machine learning to expand its research. In this paper, three-way decision is connected with SVM, a standard binary classification model in machine learning, for solving imbalanced classification problems that SVM needs to improve. A new three-way fuzzy membership function and a new fuzzy twin support vector machine with three-way membership (TWFTSVM) are proposed. The new three-way fuzzy membership function is defined to increase the certainty of uncertain data in both input space and feature space, which assigns higher fuzzy membership to minority samples compared with majority samples. To evaluate the effectiveness of the proposed model, comparative experiments are designed for forty-seven different datasets with varying imbalance ratios. In addition, datasets with different imbalance ratios are derived from the same dataset to further assess the proposed model's performance. The results show that the proposed model significantly outperforms other traditional SVM-based methods.
\end{abstract}



\maketitle

\section{Introduction}
Three-way decision \cite{Yao2010} is a powerful tool for granular computing developed in recent years to deal with uncertain data. The core idea of three-way decision is to divide a unified set into three mutually disjoint paired regions and for each region to define a corresponding decision strategy \cite{Yao2016}.
Three-way decision is typically used in traditional rough set models, where decision rules are generated using a combination of specific information within three regions of the positive and negative domain boundary domains. Yao combines three-way decision and rough set model to enrich theoretical knowledge of rough set theory significantly and makes it practical in applications because its model establishes a link between tolerance for error and cost of lousy decisions \cite{Yao2010}. Liang tackles the complex decision problem by adding loss functions to three-way decisions in decision-theoretic rough sets to make rational decisions in an intuitionistic fuzzy environment \cite{Liang2015}. Zhao studies three-way decisions in interval-valued fuzzy probabilistic rough sets based on decision-theoretic rough set approach, extending the application area of interval-valued fuzzy probability sets \cite{Zhao2016}. Zhang proposes a new model for utility-based scoring functions by considering differences between equivalence classes in classical three-way decision model \cite{Zhang2018}. Deng presents a more appropriate loss function for three-way decision in multiscale information systems to facilitate multi-attribute decision-making \cite{Deng2022}. Su introduces a new three-way decision model based on incomplete information systems that fill in missing information and makes it easier to calculate conditional probabilities and loss functions \cite{Su2023}.
For many three-way decision models, Hu \cite{Hu2014} systematically investigated three-way decision space problems and unified several existing and more representative types of three-way decision models into a mathematical theoretical framework.

Although there are many applications of three-way decision in information, decision-making, healthcare, and engineering \cite{Liu2013,Liu 2022}, there needs to be more integration with machine learning. In the hot topic prediction problem of machine learning, SVM is a vital learning technique introduced by Vapnik \cite{Vapnik1995}. SVM is widely applied in various fields such as image classification \cite{Sun2021,Mishra2021}, economy \cite{Do2009}, medical \cite{Shia2021,Wang2020}, and so on.
The imbalanced data means that there are significant differences between the numbers of different classes \cite{Sun2009}. Generally, a class with larger samples is called a negative class, while a class with smaller samples is called a positive class. IR, imbalance ratio, is defined to evaluate the degree of imbalance \cite{Newby2013}, which means the ratio of the number of negative classes to the number of positive classes. 
For the past few years, class imbalance learning (CIL) problems have been widely concerned in many areas, including e-mail folding \cite{Zhang2016}, face recognition \cite{Hupont2019}, fault diagnosis \cite{Wang2016} and et al.

In the past, the standard method of SVM to solve the imbalance problem was to artificially change data from imbalanced to balanced data with the help of ROS, RUS and SMOTE methods and then use SVM model for classification.
Three-way decision can extract decision rules from uncertain data and divide data into positive, negative, and boundary regions, resulting in two types of specific data and a few uncertain data. 
We consider whether the combination of three-way decision and SVM can be used to solve some practical problems to facilitate the practical application of three-way decision.
To tackle CIL problems, many methods are proposed to promote the performance of classifiers. These can be divided into three broad categories: data-level methods, algorithm-level methods and combination of data-level processing and algorithm-level methods.  
The data level methods, mainly under-sampling and over-sampling, use several techniques to turn imbalanced data into balanced data.
Under-sampling discards part of the negative samples to balance the number of positive and negative samples. Random under-sampling (RUS) \cite{Batista2004} is the most popular method, which turns the IR of data to 1 by randomly removing part of the negative samples. Due to the random deletion of negative samples, it is easy to lose important information and overfit. Yen \cite{Yen2009} proposed under-sampling based on clustering (SBC) method to delete some negative samples around the cluster of positive samples.
Over-sampling equals number of negative and positive samples by duplicating some positive samples. The classic method is random over-sampling (ROS) \cite{Batista2004}, whose idea is randomly replicating some positive samples. Because there are many repeatable samples in training data, it amplifies noises in positive samples. Synthetic minority over-sampling technique (SMOTE) \cite{Bowyer2002} proposed by Bowyer analyzes the information of positive samples to synthesize new samples.
Algorithm-level methods change existing algorithms to solve CIL problems effectively. Based on being cost-sensitive, Zong \cite{Zong2010} proposed a weighted limit learning machine (WELM) that assigns different weights to different samples to change the penalty factor. Thus, WELM could effectively reduce the misclassification probability of positive samples. Cheng \cite{Cheng2016} put forward a cost-sensitive large margin distribution machine (CS-LDM), which combines marginal distribution and cost sensitivity to improve the accuracy of the positive class.
In combining data and algorithm levels, SMOTEBoost \cite{Chawla2003} combines SMOTE technology and Boosting. SMOTEBoost changes initial data distribution and indirectly boosts positive samples' update weights; likewise, RUSBoosting \cite{Seiffrt2010}. EasyEnsemble \cite{Liu2009} has better generalization ability, using RUS technique to generate multiple balanced subsets to replace the base classifier of Boosting with Adaboost \cite{Freund1997}.

Replication and deletion of data at data level will result in some redundancy and missing data. The SVM should be improved at an algorithmic level by retaining all information in the original data without altering data. The improved model is guaranteed to make full use of original data information while maintaining the SVM framework. The SVM should be improved with three-way decision to solve the imbalance classification problem.
SVM regards each sample points as equally important \cite{TD2016}. However, Lin \cite{Lin2002}proposed a classic fuzzy SVM (FSVM) model, defining a fuzzy membership function for assigning different fuzzy membership to each sample in input space, making fuzzy membership of noises or outliers significantly lower than other samples. When FSVM learns the hyperplane, different samples have different contributions.
In FSVM, the critical problem is determining the fuzzy membership function for giving appropriate fuzzy membership to each sample. 

Lin \cite{Lin2002} proposed three fuzzy membership functions for different situations. For sequential learning problems, a quadric function of time makes the last sample the most important and the first sample the least important. If some problems focus on the accuracy of one-class classification, the function will make the fuzzy membership of samples belonging to a specific class 1 and 0.1 for others. The fuzzy membership is set by a function of Euclidean distance between a sample and its class center to overcome the effects of noises and outliers for common classification problems. The function assigns lower fuzzy membership to a sample far from its class center, considered a noise or outlier.

When the training set is not spherical, the fuzzy membership function cannot assign appropriate fuzzy membership for each sample, affecting the optimal hyperplane's learning. 
Many other metric methods are proposed to solve this problem to improve the classic FSVM model. Zhou \cite{Zhou2009} introduced density and dual membership to the fuzzy membership function in FSVM. Samples have different densities in different classes, so two classes of samples of membership function use two different design methods. The function can accurately reflect the influence of different samples on the optimal hyperplane. Ha \cite{Ha2013} put forward a score function to assign fuzzy membership of each training sample based on an intuitionistic fuzzy set and kernel function. This model can distinguish noises and outliers among training data very well. Yang \cite{Yang2011} applied fuzzy $c$-means (FCM) clustering method to cluster two classes from training data in FSVM, which was successfully used in fault diagnosis of wind turbine field \cite{Hang2016}. Wu \cite{Wu2014} used the partition index maximization (PIM) clustering method to calculate more valid and robust fuzzy membership in FSVM. Wang \cite{Wang2005} gave each sample two different fuzzy membership calculated by basic credit scoring methods to discriminate good creditors from bad ones. Li \cite{Li2015} proposed a regularized monotonic FSVM model, which incorporates the prior knowledge of monotonicity to define fuzzy membership of each sample. In \cite{An2013}, within-class scatter (WCS) is introduced to define a new fuzzy membership function to consider affinity among samples in FSVM.
Faced with divisible data only in high-dimensional space, these FSVM models could not calculate reasonable fuzzy membership in input space. 
Many scholars improve models in feature space to address the drawback of the above FSVM models. Jiang \cite{Jiang2006} defined a fuzzy membership function based on kernel function in high-dimensional space for nonlinear classification problems. Tang \cite{Tang2011} proposed a new fuzzy membership function that calculated membership in input and feature space.	
Lin \cite{Lin2004} proposed a new FSVM model, which calculated the similarity between sample and its label as fuzzy membership using kernel target alignment (KTA).
Wang \cite{Wang2020CKA} introduced the concept of centered kernel alignment (CKA) in FSVM. CKA was widely used in kernel learning, and selection \cite{Niazmardi}. CKA-FSVM calculated the dependence of samples and others in the same classes as fuzzy membership through CKA.
In high-dimensional space, these FSVM models become more robust for classification, mainly avoiding the influence of noises and outliers. However, defining the fuzzy membership function to calculate the most appropriate membership is still a problem worth considering.

To construct a proper fuzzy membership function, this paper considers introducing a three-way decision in FSVM. Three-way decision can provide a fuzzy membership function with a solid mathematical basis and has a prosperous theoretical basis for comparison with those fuzzy membership functions based on data structure. In order to solve model quickly, this paper introduces a twin support vector machine (TSVM) \cite{Jayadeva2007}. 
TSVM converts one significant quadratic programming problem (QPP) into two smaller QPPs \cite{Shao2011}, and the size of QPP has been changed from $m$  to $m/2$, so TSVM is four times faster than SVM and FSVM.

In this paper, we propose a new three-way fuzzy membership function and a new fuzzy twin support vector machine with three-way membership to solve CIL problems. TWFTSVM calculates membership of each sample using three-way decision with $k$-nearest neighbors of the sample in  input space. For a sample with more similar samples, membership is higher in boundary region from three-way decision, and vice versa. In feature space, TWFTSVM computes membership by using CKA \cite{Wang2020CKA} function and $k$-nearest neighbor information of the sample. The final three-way fuzzy membership function is the union of membership of input space and feature space, which assigns 1 as fuzzy membership of positive samples and assigns a valid fuzzy membership to each negative sample. TWFTSVM identifies the class of samples either belongs to the positive or negative class very well.
In summary, this paper proposes a new model called TWFTSVM, and the contributions are as follows:

\begin{itemize}
	\item TWFTSVM model inherits the advantages of TSVM in substantially improving the efficiency of classification and fuzzy membership function of FSVM, which could balance the amount of information in different categories by assigning different membership.
	
	\item TWFTSVM constructs a new three-way fuzzy membership function with a solid theoretical basis rather than simply exploiting the structural distribution of data and entirely using all information in original and mapped data.
	
	\item TWFTSVM broadens the application area of three-way decision and performs well when compared with other models in different UCI datasets with different imbalance ratios and different imbalance ratio datasets derived from the same dataset.
\end{itemize} 

The rest of this paper is organized as follows: \autoref{Section2} briefly introduces the structure of three-way decision, SVM and FSVM. Then, we describe three-way fuzzy membership function and TWFTSVM model in detail in \autoref{Section3}. Then, the comparative experiments and statistical test results are shown in \autoref{Section4}. Finally, \autoref{Section5} concludes this paper and puts forward future research.

\section{Preliminaries}\label{Section2}
This section briefly describes the structure of three-way decision, SVM and FSVM. 

\subsection{Three-way decision}
Three-way decision is divided into a generalized three-way and narrow three-way decision models. This subsection focuses on three-way decision based on fuzzy sets in a generalized three-way decision model. For any mapping of a fuzzy set X to closed interval 0 to 1, the formula is as follows:
\begin{equation}
	\left\{\begin{array}{l}
		\mu_{\tilde{A}}: X \rightarrow[0,1], \\
		x \rightarrow \mu_{\tilde{A}},
	\end{array}\right.
\end{equation}
set X has a fuzzy number $\tilde{A}$, $\mu_{\tilde{A}}$ is the fuzzy membership function of $\tilde{A}$, $\mu_{\tilde{A}}(x)$ is the fuzzy membership of $x$ to $\tilde{A}$. Then, the fuzzy number $\tilde{A}$ can be written as:
\begin{equation}
	\tilde{A}=\left\{\left(x, \mu_{\tilde{A}}(x)\right) \mid x \in X\right\}.
\end{equation}
Given a pair of thresholds $(\alpha, \beta)$ with $0 \leqslant \beta < \alpha \leqslant 1$ , three-way decision based on fuzzy set is as follows:
\begin{equation}
	\begin{aligned}
		& R_{1(\alpha, \beta)}\left(\mu_{\tilde{A}}\right)=\left\{x \in U \mid \mu_{\tilde{A}}(x) \geqslant \alpha\right\}, \\
		& R_{2(\alpha, \beta)}\left(\mu_{\tilde{A}}\right)=\left\{x \in U \mid \beta<\mu_{\tilde{A}}(x)<\alpha\right\}, \\
		& R_{3(\alpha, \beta)}\left(\mu_{\tilde{A}}\right)=\left\{x \in U \mid \mu_{\tilde{A}}(x) \leqslant \beta\right\}.
	\end{aligned}
\end{equation}
If $\mu_{\tilde{A}}(x) \geqslant \alpha$, then $x$ belongs to $\tilde{A}$. If $\mu_{\tilde{A}}(x) \leqslant \beta$, then $x$ does not belong to $\tilde{A}$. If $\beta < \mu_{\tilde{A}}(x) < \alpha $, then the state of $x$ with respect to $\tilde{A}$ is indeterminate.

\subsection{Support vector machine}
SVM is a traditional binary classification model. Suppose $S=\{(x_1, y_1), (x_2, y_2), \dots, (x_n, y_n)\}$ is training set with  $x_i \in R^d$ being a $d$-dimensional input vector and $y_i = \{-1, +1\}$ being the corresponding target class. SVM endeavors to find an optimal separation hyperplane $w^Tx+b=0$, where $w\in R^d$ is a parametric matrix as weight, and $b\in R$ is a constant as bias term. In inear SVM, the optimal separation hyperplane can be obtained by solving the following optimization problem:
\begin{equation}\label{eq1}
	\begin{aligned}
		\min \quad  &\frac{1}{2} ||w||^T + C\sum_{i=1}^{n}\xi _i,\\
		s.t. \quad &y_i(w\cdot x_i+b) \geqslant 1-\xi _i,\\
		&\xi _i \geqslant 0 , i= 1,2,\dots, n,
	\end{aligned}	
\end{equation}
where $\xi _i(i= 1,2,\dots, n)$ are slack variables, C is a penalty parameter, $n$ is the number of training samples.

The \autoref{eq1} cannot be solved directly, Therefore, we could solve its Lagrange dual problem:
\begin{equation}\label{eq2}
	\begin{aligned}
		\min \limits_{\alpha} \quad  &\frac{1}{2} \sum_{i=1}^{n} \sum_{i=1}^{n} \alpha _i \alpha_j y_i y_j (x_i\cdot x_j) - \sum_{i=1}^{n}\alpha_i,\\
		s.t. \quad &\alpha_i y_i=0, \\
		&\alpha _i \geqslant 0 , i= 1,2,\dots, n,
	\end{aligned}	
\end{equation}
where $\alpha_{is}$ are Lagrange multipliers. SVM optimal hyperplane can distinguish different classes of samples. For sample $x$, if $w^Tx+b \geqslant 0$, the label is positive, and on the contrary, the label is negative. 

\subsection{Fuzzy support vector machine}
FSVM introduces a fuzzy membership function, train set $S=\{(x_1, y_1, s_1), (x_2, y_2, s_2), \dots, (x_n, y_n, s_n)\}$, where $s_i\in \left( 0, 1\right]$ is fuzzy membership to measure the importance of  sample $x_i$ to its label $y_i$. We define a mapping function $\phi $ to map sample ${x}_{i}$ from input space $R^d$ to a feature space, and the value of the sample is turned into $\phi ({x}_{i})$. Similar to SVM, the target of FSVM is also to find an optimal hyperplane $ w^T \phi (x) + b = 0$, which can be achieved by the following:
\begin{equation}\label{eq3}
	\begin{aligned}
		\min \quad &\frac{1}{2}\|{w}\|^{2}+C \sum_{i=1}^{n} s_{i} \xi_{i},\\
		s.t. \quad &y_{i} \left((w^T \phi ({x}_{i}) + b\right) \geqslant 1-\xi_{i},\\
		& \xi_{i} \geqslant 0, i=1,2, \cdots, n,
	\end{aligned}	
\end{equation}
where $\xi _i(i= 1,2,\dots, n)$ are slack variables that make the sample separable, and C is the penalty parameter to ensure the hyperplane interval is as large as possible. The number of misclassified samples is as tiny as possible. Since $s_i$ is fuzzy membership, $s_i\xi_{i}$ assigns different weights to various samples. The misclassification cost $s_iC$ will be higher for necessary samples than those less critical samples.

Since the original FSVM problem is not a convex optimization problem, firstly, we introduce Lagrange multipliers to each sample, and \autoref{eq3} is constructing Lagrangian as follows:
\begin{equation}\label{eq4}
	\begin{aligned}
		&L(w, b, \xi, \alpha, \beta)=\frac{1}{2} \|{w}\|^{2}+C \sum_{i=1}^{n} s_{i} \xi_{i} -\sum_{i=1}^{n} \alpha_{i}\left(y_{i}\left(w \cdot \phi (x_i )+b\right)-1+\xi_{i}\right)-\sum_{i=1}^{n} \beta_{i} \xi_{i}.
	\end{aligned}
\end{equation}
Secondly, the Lagrangian $L(w, b, \xi, \alpha, \beta)$ is derived separately for  $w, b, \xi_{i}$,
\begin{equation}\label{eq5}
	\begin{aligned}
		\frac{\partial L(w, b, \xi, \alpha, \beta)}{\partial w} &=w-\sum_{i=1}^{n} \alpha_{i} y_{i} z_{i}=0, \\
		\frac{\partial L(w, b, \xi, \alpha, \beta)}{\partial b} &=-\sum_{i=1}^{n} \alpha_{i} y_{i}=0, \\
		\frac{\partial L(w, b, \xi, \alpha, \beta)}{\partial \xi_{i}} &=s_{i} C-\alpha_{i}-\beta_{i}=0.
	\end{aligned}
\end{equation}
Finally, the dual of original problem is yielded by applying \autoref{eq5} to \autoref{eq3} and \autoref{eq4}. The convex optimization problem is as follows:
\begin{equation}\label{eq6}
	\begin{aligned}
		\min \quad &\frac{1}{2}\sum_{i=1}^{n}\sum_{j=1}^{N}y_iy_j\alpha_i \alpha_jk(x_i, x_j) - \sum_{i=1}^{n}\alpha_i,\\
		s.t. \quad &\sum_{i=1}^{N}\alpha_i y_i = 0,\\
		& 0 \leqslant \alpha_i \leqslant s_iC, i=1,2, \cdots, n,
	\end{aligned}
\end{equation}
where the kernel function $k(x_i, x_j) = \phi (x_i)^T \phi (x_j )$ is defined by mapping function $\phi$ from input space to feature space. Through solving optimization problem \autoref{eq6}, the decision hyperplane can be expressed formally as
\begin{equation}\label{eq7}
	f(\boldsymbol{x})=\operatorname{sign}\left(\sum_{i=1}^{n} \alpha_{i} y_{i} k\left(\boldsymbol{x}_{i}, \boldsymbol{x}\right)+b\right).
\end{equation}
For sample $x_i$, if $f(x_i) > 0$, $x_i$ will be classified as a positive class; on the contrary, the label of $x_i$ is negative. Sample $x_i$ is said to be a support vector when the corresponding Lagrange multiplier $\alpha_{i}$ satisfies $0<\alpha_{i}<s_iC$.

\section{TWFTSVM: Fuzzy twin support vector machine with three-way membership}\label{Section3}

In this section, a new three-way fuzzy membership function is first explained in detail in terms of input space and feature space. Then, the structure of TWFTSVM is introduced in depth. Finally, the linear and nonlinear models with introduction of kernel functions are discussed at length.

\subsection{Three-way fuzzy membership assignment}\label{Section3.1}

In imbalanced data, positive samples are more important than negative ones because positive samples have a smaller number and the same information as negative ones. Therefore, the principle of fuzzy membership function is that a higher fuzzy membership should be assigned to these positive samples and lower to negative ones. Based on this principle, we introduce three-way decision \cite{Yao2016} to construct a three-way fuzzy membership function that evaluates positive and negative samples and some uncertain samples between two types of samples. Then, uncertain samples are mapped from input space into feature space as positive and negative. Therefore, CKA and $k$-nearest neighbors are combined to calculate fuzzy membership of samples in feature space. Finally, the final fuzzy membership of samples is generated by considering input and feature space. 

\textbf{In input space}, three-way decision divide a unified set into three mutually disjoint paired regions, and for each region to define a corresponding decision strategy. For classification problems, the three regions correspond to data identified as belonging to this category, identified as not belonging to this category, and indeterminate samples, respectively. Information entropy is used to measure the uncertainty of information, and the larger entropy indicates that information is more uncertain. We use information entropy and $k$-nearest neighbors to pre-process uncertain data.  

Let training set $S=\{(x_1,y_1), (x_2, y_2), \dots, (x_m, y_m)\}$, where $x_i\in R^{m\times d}$ and $y_i = \{+1,-1\}$. In the set of $k$-nearest neighbors $\{x_{i1}, \dots ,x_{ik} \}$ of sample $x_i$, the number of samples in same class as $x_i$ is $num^+(x_i)$ and number of samples in different class as $x_i$ is $num^-(x_i)$. There is an equation $H(x_i)= -p^+(x_i) \log_2 \left(p^+(x_i)\right)-p^-(x_i) \log_2 	\left(p^-(x_i)\right)$ that maps $x_i$ to $(x_i, H_i)$, where $p^+(x_i) =num^+(x_i)/k$, and $p^-(x_i) =num^-(x_i)/k.$ The value of $H_i$, which is an measure of uncertainty for $x_i$, takes the range 0 to 1. We use a pair of thresholds $(\alpha, \beta)$ in three-way decision to turn uncertain data into certainty. For sample $(x_i, y_i, H_i)$, $H_i$ is called the membership of $x_i$ to $y_i$. Given a pair of thresholds $(\alpha, \beta)$ and $0 \leq \beta < \alpha \leq 1$, the model of three-way decision can be written as follows:
\begin{equation}\label{eq3.2}
	\begin{aligned}
		R_{1(\alpha, \beta)}(y_i) &= \{x_i \in U | H_i(x_i) \geq \alpha\},\\		R_{2(\alpha, \beta)}(y_i) &= \{x_i \in U | \beta < H_i(x_i) < \alpha\},\\
		R_{3(\alpha, \beta)}(y_i) &= \{x_i \in U | H_i(x_i) \leq \beta\}.
	\end{aligned}
\end{equation}
If $H_i(x_i) \geq \alpha$, $x_i$ maybe not belong to $y_i$ but its true label is $y_i$. If $H_i(x_i) \leq \beta$, $x_i$ belongs to $y_i$. If $\beta < H_i(x_i) < \alpha$, $x_i$ is uncertain relative to $y_i$. Setting definite parameter values for thresholds $(\alpha, \beta)$ are often difficult because of the diversity of data distribution. In this paper, we use the magnitude of $k$-values affecting the uncertainty in the information entropy formula to determine threshold values $(\alpha, \beta)$. Let $\alpha = k/k+1$, and $\beta = 1 - \alpha$. Therefore, three-way fuzzy membership function in input space (TWF) can be written as follows:
\begin{equation}\label{eq3.3}
	TWF(x_i)= 
	\begin{cases}
		0.01 & R_{1(\alpha, \beta)}(y_i) = \{x_i \in U | H_i(x_i) \geq \alpha\},\\
		e^{-\frac{(\alpha \forall \beta)H(x_i)}{\alpha - \beta }} & R_{2(\alpha, \beta)}(y_i) = \{x_i \in U | \beta < H_i(x_i) < \alpha\},\\
		1 & R_{3(\alpha, \beta)}(y_i) = \{x_i \in U | H_i(x_i) \leq \beta\}.
	\end{cases}
\end{equation}
where $\alpha \forall \beta$ is $\alpha$ if $p_{num+} < 0.5$, and $\alpha \forall \beta$ is $\beta$ if $p_{num+} \geq  0.5$. $TWF(x_i)\in \left(0,1\right]$ is fuzzy membership of sample $x_i$ in input space. It ensures that samples around similar samples will have a higher fuzzy membership. Suppose $k$-nearest neighbors of a sample containing the same positive and negative samples. In that case, a lower fuzzy membership is assigned to the sample, which can weaken its contribution to the optimal hyperplane. Suppose the $k$-nearest neighbors of a sample containing a much larger number of samples of different classes than same class, a lower fuzzy membership is assigned to the sample, which is considered an outlier or noise. The essence of the function $TWF$ is to attenuate the fuzzy membership of negative samples close to positive samples. TWFTSVM has two hyperplanes, and there is no effect on support vectors.

\begin{figure}[t]
	\centering
	\includegraphics{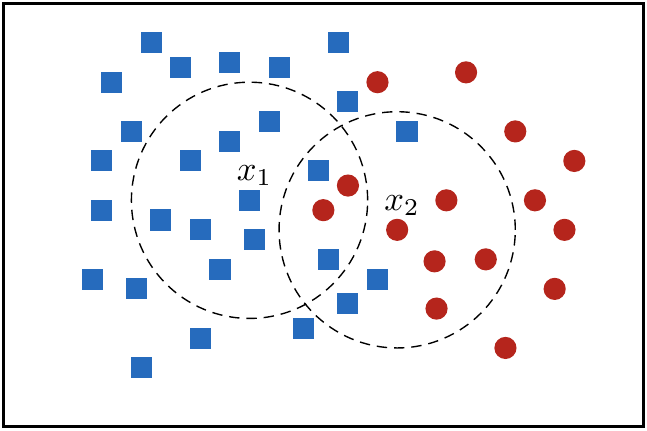}
	\caption{Demonstration on class probability evaluation.}
	\label{Demonstration on class probability evaluation.}
\end{figure}

\begin{table}[b]\footnotesize
	\caption{Entropy and fuzzymembership in original space of $x_1$ and $x_2$. } 
	\label{entropy}
	\setlength{\tabcolsep}{9mm}{
		\begin{tabular}{lllllll}
			\hline
			& $num^{+}$ & $num^{-}$ & $p^{+}$ & $p^{-}$ & H & TWF \\
			\hline
			$x_1$ & 9 & 2 & 9/11 & 2/11 & 0.68 & 0.93\\
			$x_2$ & 6 & 5 & 6/11 & 5/11 & 0.99 & 0.90 \\
			\hline
		\end{tabular}
	}
\end{table}

In \autoref{Demonstration on class probability evaluation.}, the blue squares represent negative samples, and the red dots represent positive samples. Let $k$ be 11 to calculate the information entropy. The 11-nearest neighbors of a sample $x_1$ and $x_2$ are in their circles. In \autoref{entropy}, it can be seen that $H(x_1)=0.68$ and $H(x_2)=0.99$, which shows sample $x_1$ is more certain than sample $x_2$. For sample $x_2$, $p^{+}(x_2)$ almost equal to $p^{-}(x_2)$, we consider $x_2$ be less important in computing the hyperplane of TWFTSVM and should be assigned a lower fuzzy membership. The $\alpha = 0.917$ is assigned by three-way fuzzy membership function, so $TWF(x_2)=0.90$ is less than $TWF(x_1)=0.93$. Generally, the lower information entropy a sample has, the higher fuzzy membership it assigns.

Based on Three-way decision, we use class information of sample and its $k$-nearest neighbors to design a three-way fuzzy membership function. For those samples with high information entropy, i.e. high uncertainty, three-way fuzzy membership function will assign them a lower membership. 

\textbf{In feature space}, we define a fuzzy membership function using CKA with $k$-nearest neighbors. CKA is a function that measures the relationship between two kernel matrices, which is defined as
\begin{equation}\label{eq3.4}
	\mathrm{CKA}\left(\mathbf{K}_{1}, \mathbf{K}_{2}\right)=\frac{\left\langle\overline{\mathbf{K}}_{1}, \overline{\mathbf{K}}_{2}\right\rangle_{\mathrm{F}}}{\sqrt{\left\langle\overline{\mathbf{K}}_{1}, \overline{\mathbf{K}}_{1}\right\rangle_{\mathrm{F}}\left\langle\overline{\mathbf{K}}_{2}, \overline{\mathbf{K}}_{2}\right\rangle_{\mathrm{F}}}},
\end{equation}
where $\left\langle\mathbf{\overline{K}_{1}}, \mathbf{\overline{K}}_{2}\right\rangle_{\mathrm{F}}=\sum_{i=1}^{m} \sum_{j=1}^{m} \overline{k}_{1}\left({x}_{i}, {x}_{j}\right) \overline{k}_{2}\left({x}_{i}, {x}_{j}\right)$,  $\overline{k}_1$ and $\overline{k}_2$ are two kernel functions. The $\overline{\mathbf{K}}=\mathbf{H K H}$ is centralised by centering matrix $\mathbf{H}=\mathbf{I}-\mathbf{e e}^{\mathrm{T}} / m \in \mathrm{R}^{m \times m}$, 
where $\mathbf{e}=(1, \cdots, 1)^{\mathrm{T}} \in \mathrm{R}^{m}$ and $\mathbf{I} \in \mathrm{R}^{m \times m}$ be identity matrix. $\mathbf{K_1} \in \mathbf{R}^{m \times m}$ is defined by $\mathbf{K_1}^{ij}=k\left(x_{i}, x_{j}\right)$ and $\mathbf{K_2} \in \mathbf{R}^{m \times m}$ is defined by $\mathbf{K_2}^{ij} = l\left(y_i, y_j \right)$, respectively. If $y_{i}=y_{j}$, the $l\left(y_{i}, y_{j}\right) = +1$, and on the contrary $l\left(y_i, y_j \right) = -1$. Let $\boldsymbol{y}=\left(y_{1}, \cdots, y_{m}\right)^{\mathrm{T}}$, the CKA can be rewritten as
\begin{equation}\label{eq3.5}
	\begin{aligned}
		\mathrm{CKA}(\mathbf{K_1}, \mathbf{K_2}) 
		&=\frac{\left\langle\overline{\mathbf{K}}_1, \boldsymbol{y} \boldsymbol{y}^{\mathrm{T}}\right\rangle_{\mathrm{F}}}{\sqrt{\langle\overline{\mathbf{K}}_1, \mathbf{K_1}\rangle_{\mathrm{F}}\langle\overline{\mathbf{K}}_2, \mathbf{K_2}\rangle_{\mathrm{F}}}} =\frac{\sum\limits_{i=1}^{n} \sum\limits_{j=1}^{n} y_{i} y_{j} \bar{k}\left(\boldsymbol{x}_{i} \boldsymbol{x}_{j}\right)}{\sqrt{\langle\overline{\mathbf{K}}_1, \mathbf{K_1}\rangle_{\mathrm{F}}\langle\overline{\mathbf{K}}_2, \mathbf{K_2}\rangle_{\mathrm{F}}}} \\
		&=\frac{\sum\limits_{y_{i}=y_{j}} \bar{k}\left(\boldsymbol{x}_{i}, \boldsymbol{x}_{j}\right)-\sum\limits_{y_{i} \neq y_{j}} \bar{k}\left(\boldsymbol{x}_{i}, \boldsymbol{x}_{j}\right)}{\sqrt{\langle\overline{\mathbf{K}}_1, \mathbf{K_1}\rangle_{\mathrm{F}}\langle\overline{\mathbf{K}}_2, \mathbf{K_2}\rangle_{\mathrm{F}}}}.
	\end{aligned}
\end{equation}

Based on this equation, we measure fuzzy membership of a sample with all similar and dissimilar samples in its $k$-nearest neighbors. For sample $x_i$, fuzzy membership function is defined as follow:
\begin{equation}\label{eq3.6}
	\operatorname{KF}\left(\mathbf{K_1}, \mathbf{K_2}, \boldsymbol{x}_{i}\right)=\sum_{y_{ik}=y_{i}} \bar{k}\left(\boldsymbol{x}_{i}, \boldsymbol{x}_{ik}\right)-\sum_{y_{ik} \neq y_{i}} \bar{k}\left(\boldsymbol{x}_{i}, \boldsymbol{x}_{ik}\right),
\end{equation}
where $k$-nearest neighbors of sample $x_i$ is $\{x_{i1}, \dots ,x_{ik} \}$, and $y_{ik}$ is the class of sample $x_{ik}$, and the kernel $\bar{k}\left(\boldsymbol{x}_{i}, \boldsymbol{x}_{ik}\right)$ measures the similarity between $x_i$ and $x_{ik}$. For convenience, $\operatorname{KF}\left(\mathbf{K_1}, \mathbf{K_2}, \boldsymbol{x}_{i}\right)$ is abbreviated as $\operatorname{KF}\left( \boldsymbol{x}_{i}\right)$. The $\operatorname{KF}\left( \boldsymbol{x}_{i}\right)$ assigns a higher fuzzy membership to sample $x_i$ with more samples of the same class in its $k$-nearest neighbors. $\operatorname{KF}\left( \boldsymbol{x}_{i}\right)$ not only assigns a valid fuzzy membership to each sample but also effectively mitigates the effects of outliers and noises. For negative sample $x_i$, we apply the linear normalization function to map fuzzy membership in unit interval \cite{Wang2005}.

Then, we combine and normalize $TWF(x_i)$ and $KF(x_i)$ as final fuzzy membership $IKF(x_i)$ for each negative sample, where $TWKF(x_i) = TWF(x_i) * {KF}(x_i)$. For imbalanced data, we should give higher fuzzy membership to positive samples than negative ones. In order to strengthen the importance of positive samples, final three-way fuzzy membership function gives 1 as fuzzy membership of positive samples. The fuzzy membership of negative samples is determined by $TWKF(x_i)$, which ensures that negative samples are less important than positive ones. So fuzzy membership of a training sample $x_i$ is assigned as follows:
\begin{equation}\label{eq3.7}
	s_{i}= \begin{cases}1.0 & \text { if } y_{i}=+1,\\ TWKF(x_i) & \text { if } y_{i}=-1. \end{cases}
\end{equation}
So far, fuzzy membership of all training samples is determined, and specific calculation steps are described in \autoref{algTWFTSVM}.

\begin{algorithm}[htbp] 
	\caption{Algorithm of final three-way fuzzy membership function}
	\label{algTWFTSVM}
	\LinesNumbered
	\KwIn{Training set $S=\{(x_1, y_1), (x_2, y_2), \dots, (x_m,y_m)\}$, parameters $\delta$, nearest neighbors $k$, threshold $\alpha$ and $\beta$, kernel function $\emph{ker}$.}
	\KwOut{Fuzzy membership $s_i$}
	
	\For{$x_i \in S$}{
		\uIf{$y_i = +1$}{
			$s_i = 1;$
		}
		\ElseIf{$y_i = -1$}{
			Calculate k nearest neighbors $\{x_{i1}, x_{i2}, \dots, x_{ik}\}$;
			
			Count $num^+(x_i)$ and $num^-(x_i)$ to calculate $p^+(x_i)$ and $p^-(x_i)$;
			
			Calculate the $TWF(x_i)$ according to \autoref{eq3.3};
			
			Calculate the $KF(x_i)$ according to \autoref{eq3.6} and normalize it;
			
			Calculate the $TWKF(x_i)$ and normalize it;
			
			$s_i = TWKF(x_i).$
		}
	}
\end{algorithm}

\subsection{Linear TWFTSVM}

We propose TWFTSVM, an applied three-way fuzzy membership function defined in \autoref{Section3.1} to solve imbalanced data. Suppose training set $S = \{\left(x_1, y_i, s_i \right))\}_{1}^{m}$ with $x_i\in R^d$, and $y_i\in \{+1, -1\}$ represents the corresponding class label ($+1$ for positive class, and $-1$ for negative class), and $s_i$ is fuzzy membership which is determined by \autoref{eq3.7}. The details of the linear kernel TWFTSVM model is as follows:
\begin{equation}\label{eq3.8}
	\begin{aligned}
		\min _{w_{1}, b_{1}, \xi_{2}} \quad &\frac{1}{2}\left\|A w_{1}+e_{1} b_{1}\right\|^{2}+ C_{2} s_{2}^{T} \xi_{2}, \\
		s.t. \quad &-\left(B w_{1}+e_{2} b_{1}\right)+\xi_{2} \geq e_{2},\\
		&\xi_{2} \geq 0,
	\end{aligned}
\end{equation}
and
\begin{equation}\label{eq3.9}
	\begin{aligned}
		\min _{w_{2}, b_{2}, \xi_{1}} \quad &\frac{1}{2}\left\|B w_{2}+e_{2} b_{2}\right\|^{2}+ C_{4} s_{1}^{T} \xi_{1}, \\
		s.t. \quad &-\left(A w_{2}+e_{1} b_{2}\right)+\xi_{1} \geq e_{1},\\
		&\xi_{1} \geq 0,
	\end{aligned}
\end{equation}
where $C_2$ and $C_4$ are penalty parameters, $\xi _1$ and $\xi_2$ are slack variables, $e_1$ and $e_2$ are vectors with appropriate dimensions, and $s_1\in R^{m_1}$ and $s_2\in R^{m_2}$ are fuzzy membership vectors of class. The $m_1$ is number of positive samples, and $m_2$ is the number of negative samples. To solve convex optimization problems better, Lagrangian of \autoref{eq3.8} and \autoref{eq3.9} is introduced by Lagrangian multipliers $\alpha_{1}, \beta_{1}, \alpha_{2}, \beta_{2}$ as follows:
\begin{equation}\label{eq3.10}
	\begin{aligned}
		&L(w_1, b_1, \xi _2, \alpha _1, \beta _1) = \frac{1}{2}\|Aw_1+e_1b_1\|^2 + C_2s_2^T\xi _2 -\alpha_1\left[ - \left( Bw_1+e_2b_1 \right) +\xi _2 - e_2 \right]- \beta_1\xi_2,
	\end{aligned}
\end{equation}
and
\begin{equation}\label{eq3.11}
	\begin{aligned}
		&L(w_2, b_2, \xi _1, \alpha _2, \beta _2) = \frac{1}{2}\|Bw_2+e_2b_2\|^2 + C_4s_1^T\xi _1 -\alpha_2\left[ - \left( Aw_1+e_2b_1 \right) +\xi _1 - e_1 \right]- \beta_2\xi_1,
	\end{aligned}
\end{equation}
where $\alpha_1=\left(\alpha_{11}, \dots ,\alpha_{1m_2} \right)^T$, $\beta_1=\left(\beta_{11}, \dots , \beta_{1m_2}\right)$, $\alpha_2=\left(\alpha_{21}, \dots ,\alpha_{2m_1} \right)^T$ and $\beta_2=\left(\beta_{21}, \dots , \beta_{2m_1}\right)$. K.K.T conditions of \autoref{eq3.10} can be obtained as follows:		
\begin{equation}\label{eq3.12}
	\begin{aligned}
		&\frac{\alpha L}{\partial w_{1}}=A^{\top}\left(A w_{1}+e_{1} b_{1}\right)+\alpha_{1} B, \\
		&\frac{\partial L}{b_{1}}=e_{1}^{\top}\left(A w_{1}+e_{1} b_{1}\right)+\alpha_{1} e_{2}, \\
		&\frac{\alpha L}{\partial \xi_{2}}=C_{2} s_{2}^{\top}-\alpha_{1}-\beta_{1}.
	\end{aligned}
\end{equation}
Then, we combine \autoref{eq3.12} to obtain
\begin{equation}\label{eq3.13}
	\left(\begin{array}{c}
		A^{T} \\
		e_{1}^{T}
	\end{array}\right)\left(A e_{1}\right)\left(\begin{array}{l}
		w_{1} \\
		b_{1}
	\end{array}\right)+\left(\begin{array}{c}
		B \\
		e_{2}
	\end{array}\right) \alpha_1 =0.
\end{equation} 
Define $H_1 = \left( A, e_1 \right)$, $G_2 = \left( B, e_2 \right)$ and $u_1 = \left( w_1, b_1 \right)^T$, naturally, \autoref{eq3.13} is transformed into the following form:
\begin{equation}\label{eq3.14}
	H_{1}^{T} H_{1} u_{1}+G_{2}^{T} \alpha=0 \Rightarrow u_{1}=-\left(H_{1}^{T} H_{1}\right)^{-1} G_{2}^{T} \alpha.
\end{equation}
To overcome the difficulty of calculating $H_1^TH_1$ directly, a regularization term $C_1I$ is introduced in \autoref{eq3.14}, where $I$ is an identity vector with the appropriate dimension. Thus,
\begin{equation}\label{eq3.15}
	u_{1}=-\left(H_{1}^{T} H_{1}+C_{1} I\right)^{-1} G_{2}^{T} \alpha _1.
\end{equation}
In the same way, we define $u_2 = \left( w_2\quad  b_2 \right)^T$ to get K.K.T conditions of \autoref{eq3.11} as:
\begin{equation}
	u_{2}=\left(G_{2}^{T} G_{2}+C_{3} I\right)^{-1} H_{1}^{T} \alpha _2.
\end{equation}
Using K.K.T conditions, the Wolfe dual problem can be written as:
\begin{equation}\label{eq3.17}
	\begin{aligned}
		\min \limits_{\alpha _1} \quad & \frac{1}{2}\alpha_{1}^T G_2\left(H_1^TH_1+C_1I\right)^{-1}G_2^T\alpha_{1} - e_2^T\alpha_{1},\\
		s.t. \quad & 0\leq \alpha_{1} \leq C_2s_2,
	\end{aligned}
\end{equation}
and
\begin{equation}\label{eq3.18}
	\begin{aligned}
		\min \limits_{\alpha _2} \quad & \frac{1}{2}\alpha_{1}^T H_1\left(G_2^TG_2+C_3I\right)^{-1}H_1^T\alpha_{2} - e_1^T\alpha_{2},\\
		s.t. \quad & 0\leq \alpha_{2} \leq C_4s_1.
	\end{aligned}
\end{equation}
Through solving the optimization problem \autoref{eq3.17} and \autoref{eq3.18}, decision hyperplane can be expressed formally as:
\begin{equation}\label{eq3.19}
	\begin{aligned}
		&f_1(x) = w_1^Tx +b_1 = 0,\\
		&f_2(x) = w_2^Tx +b_2 = 0,
	\end{aligned}
\end{equation}
where the $w_1,w_2\in R^d$ and $b_1,b_2\in R$. The class $y_i$ where a new sample $x$ belongs can be discerned by the discriminatory formula as follow: 
\begin{equation}\label{eq3.20}
	y_i = \arg \min _{j =1,2} \frac{|w_j^Tx+b_j|}{\|w_j\|},
\end{equation}	
where $|\cdot|$ is the symbol of absolute value. For the hyperplane $w_1^Tx +b_1 = 0$, its support vectors are all positive samples, and support vectors of hyperplane $w_2^Tx +b_2 = 0$ are all negative samples.

\subsection{Nolinear TWFTSVM}

To solve nonlinear classification problems, the structure of TWFTSVM is defined as follow:
\begin{equation}\label{eq3.21}
	\begin{aligned}
		\min _{w_{1}, b_{1}, \xi_{2}} \quad &\frac{1}{2}\left\|k\left(A, X^{T}\right) w_{1}+e_{1} b_{1}\right\|^{2}+C_{2} s_{2}^{T} \xi_{2}, \\
		s.t. \quad &-\left(k\left(B, X^{T}\right) w_{1}+e_{2} b_{1}\right)+\xi_{2} \geq e_{2},\\
		& \xi_{2} \geq 0,
	\end{aligned}
\end{equation}
and
\begin{equation}\label{eq3.22}
	\begin{aligned}
		\min _{w_{2}, b_{2}, \xi_{1}} \quad &\frac{1}{2}\left\|k\left(B, X^{T}\right) w_{2}+e_{2} b_{2}\right\|^{2}+C_{4} s_{1}^{T} \xi_{1}, \\
		s.t. \quad & \left(k\left(A, X^{T}\right) w_{2}+e_{1} b_{2}\right)+\xi_{1} \geq e_{1},\\
		& \xi_{1} \geq 0.
	\end{aligned}
\end{equation}
where kernel function is $k\left(x_{1}, x_{2}\right)=\left(\phi\left(x_{1}\right), \phi\left(x_{2} \right) \right)$. Same as the linear TWFTSVM model, the Lagrangian structure of \autoref{eq3.21} as below:
\begin{equation}\label{eq3.23}
	\begin{aligned}
		L\left(w_{1}, b_{1}, \xi_{2}, \alpha _1, \beta _1 \right) &=  \frac{1}{2}\left\|k\left(A, X^{T}\right) w_{1}+e_{1} b_{1}\right\|^{2} +C_{2} s_{2}^{T} \xi_{2}\\
		&+\alpha_1\left[\left(k\left(B, X^{T}\right) w_{1}\right.\right.
		\left.\left.+e_{2} b_{1}\right)-\xi_{2}+e_{2}\right] -\beta_1 \xi_{2}
	\end{aligned}		
\end{equation}
and 
\begin{equation}\label{eq3.24}
	\begin{aligned}
		L\left(w_{2}, b_{2}, \xi_{1}, \alpha _2, \beta _2 \right) &= \frac{1}{2}\left\|k\left(A, X^{T}\right) w_{2}+e_{2} b_{2}\right\|^{2} +C_{4} s_{1}^{T} \xi_{1} \\
		&+\alpha_2\left[\left(k\left(B, X^{T}\right) w_{1}\right.\right.
		\left.\left.+e_{1} b_{2}\right)-\xi_{1}+e_{1}\right] -\beta_2 \xi_{1}.
	\end{aligned}		
\end{equation}
K.K.T conditions of \autoref{eq3.21} are achieved as follows:
\begin{equation}\label{eq3.25}
	\begin{aligned}
		\frac{\partial L}{\partial w_{1}}= &k\left(A, X^{T}\right)^{T}\left(k\left(A, X^{T}\right) w_{1}+e_{1} b_{1}\right) +\alpha _1 k\left(B, X^{T}\right)=0, \\
		\frac{\partial L}{\partial b_{1}}= &e_{1}^{T}\left(k\left(A, X^{T}\right) w_{1}+e_{1} b_{1}\right)+\alpha _1 e_{2}=0, \\
		\frac{\partial L}{\partial \xi_{2}}= &C_{2} s_{2}^{T}-\alpha _1-\beta _1=0,
	\end{aligned}
\end{equation}
Then, all sub-equation in \autoref{eq3.25} are combined to obtain
\begin{equation}\label{eq3.26}
	\begin{aligned}
		&\left(\begin{array}{c}
			k\left(A, X^{T}\right)^{T} \\
			e_{1}^{T}
		\end{array}\right)\left(k\left(A, X^{T}\right) e_{1}\right)\left(\begin{array}{c}
			w_{1} \\
			b_{1}
		\end{array}\right) +\left(\begin{array}{c}
			k\left(B, X^{T}\right) \\
			e_{2}
		\end{array}\right) \alpha=0 .
	\end{aligned}
\end{equation}
Let $H_{1}^{*}=\left(k\left(A, X^{T}\right) \quad e_{1}\right)$, $G_{2}^{*}=\left(k\left(B, X^{T}\right) \quad e_{2}\right)$, and $u_1^* = (w_1\quad b_1)^T$, then \autoref{eq3.26} can be written as:
\begin{equation}\label{eq3.27}
	u_{1}^{*}=-\left(H_{1}^{T *} H_{1}^{*}+C_{1} I\right)^{-1} G_{2}^{T *} \alpha _1.
\end{equation}
In the same way, we can obtain 
\begin{equation}\label{eq3.28}
	u_{2}^{*}=\left(G_{2}^{T *} G_{2}^{*}+C_{3} I\right)^{-1} H_{1}^{T *} \alpha _2.
\end{equation} 
With the K.K.T conditions and Lagrangian method, the corresponding Wolfe dual can be written as:
\begin{equation}\label{eq3.29}
	\begin{aligned}
		\min \limits_{\alpha _1} \quad & \frac{1}{2}\alpha_{1}^T G_2^*\left(H_1^{T*}H_1^*+C_1I\right)^{-1}G_2^{T*}\alpha_{1} - e_2^T\alpha_{1},\\
		s.t. \quad & 0\leq \alpha_{1} \leq C_2s_2,
	\end{aligned}
\end{equation}
and 
\begin{equation}\label{eq3.30}
	\begin{aligned}
		\min \limits_{\alpha _2} \quad & \frac{1}{2}\alpha_{1}^T H_1^*\left(G_2^{T*}G_2^*+C_3I\right)^{-1}H_1^{T*}\alpha_{2} - e_1^T\alpha_{2},\\
		s.t. \quad & 0\leq \alpha_{2} \leq C_4s_1.
	\end{aligned}
\end{equation}
By solving the above optimal problems, two nonparallel hyperplanes are as follows:
\begin{equation}
	\begin{aligned}
		& k\left(x, X^{T}\right) w_{1}+b_{1}=0,\\
		& k\left(x, X^{T}\right) w_{2}+b_{2}=0.
	\end{aligned}
\end{equation}
For a new sample $x$, its class $y_i$ is determined by the following equation:
\begin{equation}
	k=\underset{i=1,2}{\arg \min }\left\{\frac{\left|w_{1}^{T} k\left(x, X^{T}\right)+b_{1}\right|}{\sqrt{w_{1}^{T} k\left(A, X^{T}\right) w_{1}}}, \frac{\left|w_{2}^{T} k\left(x, X^{T}\right)+b_{2}\right|}{\sqrt{w_{2}^{T} k\left(B, X^{T}\right) w_{2}}}\right\},
\end{equation}
where $|\cdot|$ is the symbol of absolute value.

\section{Experiments}\label{Section4}

\subsection{Datasets description and comparison algorithms}\label{Section4.1}

In order to measure the effectiveness and generalization ability of TWFTSVM in this paper, twelve benchmark datasets, including both binary and multiclassification datasets, are selected from UCI. The classified datasets are the heart-stalog, pima, breast-cancer-wisconsin, liver-disorders, wisc and haberman. For other multi-classified datasets, we transform them into classified datasets. For example, dataset ecoli126Vs345 groups classes 1,2 and 3 of dataset ecoli into positive class and classes 3,4 and 5 of dataset ecoli into negative class. \autoref{tabledatasets} shows the details of imbalanced datasets. 

Different IR datasets are obtained in experiments after processing benchmark datasets from UCI. In order to investigate the effect of TWFTSVM proposed in this paper, we design comparative experiments in different datasets containing low, medium, and high IR. SVM \cite{Vapnik1995}, ROS-SVM \cite{Batista2004}, RUS-SVM \cite{Batista2004}, SMOTE-SVM \cite{Bowyer2002}, FSVM \cite{Lin2002}, TSVM \cite{Jayadeva2007} and CKA-FSVM \cite{Wang2020CKA} as comparison algorithms in experiments. ROS-SVM, RUS-SVM, and SMOTE-SVM are  standard methods used by SVM to solve imbalance classification problem in the past, and CKA-FSVM is the state-of-the-art method.

We evaluate the generalization of different algorithms using 10-fold cross-validation. The RBF kernel function $ker(x_i, x_j) = exp^{\left(\|x_i-x_j\|^2/2\sigma ^2\right)}$ which has the well-mapped capability as kernel function of experiments, and kernel parameter $\sigma ^2 = \frac{1}{N^2}\sum_{i,j = 1}^{N}\|x_i - x_j\|^2$. In SVM, ROS-SVM, RUS-SVM, SMOTE-SVM, FSVM and  CKA-FSVM, parameter $\sigma ^2$ is selected from  $\{10^{-2}, 10^{-1}, 10^{0}, 10^{1}\}$ and penalty parameter $C$ is selected from $\{2^{-3}, 2^{-1}, \dots, 2^{10}\}$. In TSVM, postive penalty $C_1 = C_3$ and regularization parameter $C_2 = C_4$ are selected from $\{2^{-1}, 2^{-0}, \dots , 2^{5}\}$. For TWFTSVM, we let parameters $C1 = C3$ and $C2 = C4$ same as TSVM. The number $k$ of nearest neighbors is selected from $\{7, 9, 11, 13\}$, according to $\alpha = \frac{k}{k+1}$ and $\beta = 1 - \alpha$ so the range of thresholds $(\alpha, \beta)$ is located in the set of $\{(0.875, 0.125), (0.900, 0.100), (0.917, 0.083), (0.929, 0.071)\}$.

\begin{table}[b]\footnotesize
	\caption{Information on imbalanced datasets.} 
	\label{tabledatasets}
	\resizebox{\linewidth}{!}{
		\begin{tabular}{llllllllllll}
			\hline
			Dataset                 & Pos. & Neg. & Inst. & Dim. & IR    & Dataset       & Pos. & Neg. & Inst. & Dim. & IR    \\
			\hline
			\textbf{Low imbalance}(IR $\leq$ 4.0)          &      &      &       &      &       &               &      &      &       &      &       \\
			heart-statlog           & 120  & 150  & 270   & 13   & 1.25  & haberman      & 81   & 225  & 306   & 3    & 2.78  \\
			pima                    & 368  & 500  & 768   & 8    & 1.36  & glass015Vs7   & 29   & 83   & 112   & 9    & 2.86  \\
			liver-disorders         & 145  & 200  & 345   & 6    & 1.38  & vehicle1      & 218  & 628  & 846   & 18   & 2.88  \\
			zoo0                    & 41   & 60   & 101   & 16   & 1.46  & vehicle2      & 212  & 634  & 846   & 18   & 2.99  \\
			glass2                  & 76   & 138  & 214   & 9    & 1.82  & glass013Vs7   & 29   & 87   & 116   & 9    & 3.00  \\
			ecoli1Vs2               & 77   & 142  & 219   & 7    & 1.84  & ecoli1346Vs25 & 82   & 250  & 332   & 7    & 3.05  \\
			breast-cancer-wisconsin & 241  & 458  & 699   & 9    & 1.90  & glass025Vs7   & 29   & 89   & 118   & 9    & 3.07  \\
			wisc                    & 241  & 458  & 699   & 9    & 1.90  & zoo02356Vs14  & 24   & 77   & 101   & 16   & 3.21  \\
			seeds3                  & 70   & 140  & 210   & 7    & 2.00  & ecoli123Vs456 & 77   & 255  & 332   & 7    & 3.32  \\
			wine1                   & 59   & 119  & 178   & 13   & 2.02  & ecoli1        & 77   & 259  & 336   & 7    & 3.36  \\
			zoo02346Vs15            & 28   & 73   & 101   & 16   & 2.61  & glass026Vs7   & 29   & 98   & 127   & 9    & 3.38  \\
			wine                    & 48   & 130  & 178   & 13   & 2.71  & zoo01246Vs35  & 21   & 80   & 101   & 16   & 3.81  \\
			ecoli1Vs6               & 52   & 142  & 194   & 7    & 2.73  &               &      &      &       &      &       \\
			\textbf{Medium imbalance} (4.0 $<$ IR $\leq$ 7.0)       &      &      &       &      &       &               &      &      &       &      &       \\
			zoo0135Vs246            & 19   & 82   & 101   & 16   & 4.32  & glass026Vs3   & 17   & 85   & 102   & 9    & 5.00  \\
			glass02Vs3              & 17   & 76   & 93    & 9    & 4.47  & glass012Vs7   & 29   & 146  & 175   & 9    & 5.03  \\
			ecoli126Vs345           & 60   & 272  & 332   & 7    & 4.53  & ecoli1256Vs34 & 55   & 277  & 332   & 7    & 5.04  \\
			ecoli124Vs6             & 52   & 239  & 291   & 7    & 4.60  & glass017Vs3   & 17   & 99   & 116   & 9    & 5.82  \\
			zoo01456Vs23            & 18   & 83   & 101   & 16   & 4.61  & ecoli146Vs3   & 35   & 214  & 249   & 7    & 6.11  \\
			zoo01234Vs56            & 18   & 83   & 101   & 16   & 4.61  & glass6        & 29   & 185  & 214   & 9    & 6.38  \\
			zoo01256Vs34            & 17   & 84   & 101   & 16   & 4.94  &               &      &      &       &      &       \\
			\textbf{High imbalance} (IR $>$ 7.0)       &      &      &       &      &       &               &      &      &       &      &       \\
			ecoli1Vs4               & 20   & 142  & 162   & 7    & 7.10  & ecoli123Vs4   & 20   & 254  & 274   & 7    & 12.70 \\
			ecoli1246Vs35           & 40   & 292  & 332   & 7    & 7.30  & ecoli126Vs4   & 20   & 271  & 291   & 7    & 13.55 \\
			ecoli126Vs3             & 35   & 271  & 306   & 7    & 7.74  & glass4        & 13   & 201  & 214   & 9    & 15.46 \\
			ecoli3                  & 35   & 301  & 336   & 7    & 8.60  & ecoli5        & 20   & 312  & 332   & 7    & 15.60 \\
			ecoli1236Vs45           & 25   & 307  & 332   & 7    & 12.28 &               &      &      &       &      &       \\
			\hline	
		\end{tabular}
	}
\end{table}

Because correctly classified negative samples greatly influence accuracy, the accuracy evaluation in traditional classification models must be better suited to CIL problems. It is not easy to measure whether positive samples are correctly classified. For better measurement of the performance of TWFTSVM and compared algorithms on imbalanced datasets, G-Means $= \sqrt{SE\times SP}$ is widely used to evaluate CIL problems. The SE is the percentage of positive samples correctly identified, and SP is the percentage of negative samples correctly identified in the test set.

\subsection{Experiments on low imbalanced datasets}

In this section, TWFTSVM, with the comparison algorithms ROS-SVM, RUS-SVM, SMOTE-SVM, FSVM, TSVM and CKA-SVM, measured the effect on low imbalanced datasets. \autoref{tablelowIR} details experimental results of 25 low IR datasets. For each dataset, the best results of the different algorithms are shown in bold. The average rank of each algorithm across all datasets is listed in the last row. In \autoref{tablelowIR}, it is shown that TWFTSVM performs best overall 25 low IR datasets compared with ROS-SVM, RUS-SVM, SMOTE-SVM, FSVM, TSVM and CKA-SVM because it has the minor average rank of all compared algorithms. TWFTSVM is ranked 2nd in the ecoli1Vs6, haberman and ecoli123Vs456 datasets and 1st in all other datasets.  

\begin{table}[h]\footnotesize
	\caption{G-Means of the compared algorithms on low imbalanced datasets.(1 $<$ IR $\leq$ 4)} 
	\label{tablelowIR}
	\resizebox{\linewidth}{!}{
		\begin{tabular}{lllllllll}
			\hline
			& SVM             & ROS-SVM         & RUS-SVM         & SMOTE-SVM       & FSVM    & CKA-FSVM        & TSVM            & TWFTSVM           \\
			\hline
			heart-statlog           & 50.96$\pm$08.71  & 57.84$\pm$07.38  & 50.96$\pm$08.71  & 52.75$\pm$09.20   & 58.08$\pm$08.76  & 57.23$\pm$10.11 & 62.09$\pm$09.76  & \textbf{79.74$\pm$04.93}  \\
			pima            & 39.35$\pm$14.36 & 63.48$\pm$06.30   & 38.84$\pm$14.01 & 44.84$\pm$06.27  & 58.17$\pm$07.19  & 58.63$\pm$06.57  & 64.22$\pm$10.06 & \textbf{74.09$\pm$04.20}   \\
			liver-disorders         & 55.74$\pm$16.58 & 58.38$\pm$12.72 & 57.72$\pm$15.59 & 55.77$\pm$15.03 & 59.08$\pm$14.51 & 60.43$\pm$10.72 & \textbf{62.76$\pm$09.41}  & 62.32$\pm$05.75  \\
			zoo0                    & 98.66$\pm$04.02  & \textbf{100.00$\pm$00.00}   & 97.32$\pm$05.36  & \textbf{100.0$\pm$00.00}   & 97.32$\pm$05.36  & 93.05$\pm$09.56  & 98.66$\pm$04.02  & \textbf{100.0$\pm$00.00}   \\
			glass2                  & 60.99$\pm$26.03 & 63.45$\pm$15.06 & 54.73$\pm$30.05 & 61.39$\pm$25.80  & 55.43$\pm$29.08 & 58.76$\pm$24.95 & 56.54$\pm$29.80  & \textbf{65.79$\pm$17.72} \\
			ecoli1Vs2               & 93.61$\pm$14.79 & 95.43$\pm$09.50   & 95.06$\pm$09.39  & 95.06$\pm$09.39  & 92.56$\pm$14.50  & 93.61$\pm$14.79 & 95.68$\pm$08.77  & \textbf{96.55$\pm$04.28}  \\
			breast-cancer-wisconsin & 96.22$\pm$04.07  & 95.94$\pm$03.11  & 96.22$\pm$04.07  & 96.99$\pm$02.69  & 95.87$\pm$03.76  & 96.01$\pm$03.52  & 96.80$\pm$02.53   & \textbf{97.41$\pm$02.02}  \\
			wisc                    & 94.03$\pm$01.45  & 96.97$\pm$01.63  & 94.35$\pm$01.80   & 96.87$\pm$01.53  & 97.07$\pm$01.72  & 97.19$\pm$01.86  & 96.78$\pm$01.49  & \textbf{97.38$\pm$01.53}  \\
			seeds3                  & 86.93$\pm$12.95 & 90.88$\pm$06.79  & 86.93$\pm$12.95 & 86.93$\pm$12.95 & 89.19$\pm$07.43  & 91.06$\pm$08.07  & 90.94$\pm$09.29  & \textbf{93.04$\pm$07.58}  \\
			wine1                   & 75.66$\pm$11.13 & 84.14$\pm$07.66  & 75.66$\pm$11.13 & 82.8$0\pm$08.66   & 83.10$\pm$07.30   & 80.54$\pm$10.54 & 81.79$\pm$07.78  & \textbf{96.96$\pm$03.92}  \\
			zoo02346Vs15            & 96.33$\pm$07.34  & 98.16$\pm$05.51  & 95.77$\pm$12.68 & 95.77$\pm$12.68 & 98.16$\pm$05.51  & 84.21$\pm$15.57 & 97.42$\pm$05.70   & \textbf{100.00$\pm$00.00}   \\
			wine                    & 97.77$\pm$03.28  & 96.36$\pm$04.72  & 96.36$\pm$04.72  & 97.10$\pm$04.12   & 98.16$\pm$03.29  & 97.77$\pm$03.28  & 98.55$\pm$03.25  & \textbf{99.22$\pm$01.57}  \\
			ecoli1Vs6               & 95.23$\pm$04.04  & 94.86$\pm$04.09  & \textbf{96.29$\pm$03.75}  & 93.75$\pm$04.05  & 94.10$\pm$04.13   & 94.83$\pm$04.68  & 94.40$\pm$07.12   & \textbf{96.29$\pm$03.75}  \\
			haberman                & 48.80$\pm$24.67  & 59.08$\pm$12.05 & \textbf{61.52$\pm$11.78} & 47.08$\pm$21.16 & 49.71$\pm$19.48 & 45.45$\pm$25.28 & 49.51$\pm$10.28 & 60.75$\pm$12.41 \\
			glass015Vs7             & 83.59$\pm$29.09 & 81.95$\pm$30.15 & 81.20$\pm$30.11  & 76.13$\pm$29.50  & 82.76$\pm$29.32 & 72.06$\pm$37.29 & 87.58$\pm$15.69 & \textbf{90.92$\pm$10.42} \\
			vehicle1                & 55.65$\pm$08.35  & 72.66$\pm$07.93  & 54.48$\pm$07.87  & 60.67$\pm$08.53  & 73.85$\pm$06.71  & 74.48$\pm$06.58  & 95.56$\pm$02.31  & \textbf{96.77$\pm$01.89}  \\
			vehicle2                & 08.64$\pm$10.59  & 59.31$\pm$05.73  & 47.33$\pm$12.46 & 42.58$\pm$11.12 & 37.19$\pm$07.93  & 36.14$\pm$07.88  & 74.17$\pm$08.58  & \textbf{78.53$\pm$04.01}  \\
			glass013Vs7             & 85.77$\pm$31.24 & 88.16$\pm$29.89 & 88.16$\pm$29.89 & 88.16$\pm$29.89 & 93.94$\pm$13.25 & 85.77$\pm$31.24 & 88.16$\pm$29.89 & \textbf{98.16$\pm$05.51}  \\
			ecoli1346Vs25           & 76.10$\pm$20.49  & 78.92$\pm$28.64 & 79.63$\pm$28.76 & 80.27$\pm$28.90  & 76.87$\pm$29.82 & 77.49$\pm$29.02 & 79.81$\pm$16.96 & \textbf{83.11$\pm$23.82} \\
			glass025Vs7             & 78.16$\pm$39.46 & 90.94$\pm$08.56  & 84.42$\pm$29.08 & 82.01$\pm$28.32 & 76.33$\pm$38.82 & 76.33$\pm$38.82 & 89.44$\pm$13.47 & \textbf{90.89$\pm$13.05} \\
			zoo02356Vs14            & 98.45$\pm$04.65  & 99.26$\pm$02.23  & 98.45$\pm$04.65  & 94.49$\pm$09.73  & 96.33$\pm$08.82  & 96.33$\pm$08.82  & 97.07$\pm$08.79  & \textbf{99.26$\pm$02.23}  \\
			ecoli123Vs456           & \textbf{95.54$\pm$04.49}  & 94.69$\pm$03.25  & 95.07$\pm$03.13  & 95.25$\pm$02.92  & 95.09$\pm$03.74  & 95.28$\pm$03.50   & 93.74$\pm$06.05  & 95.28$\pm$03.50   \\
			ecoli1                  & 81.45$\pm$16.04 & 77.60$\pm$22.21  & 81.45$\pm$23.7  & 82.42$\pm$15.71 & 80.84$\pm$23.06 & 75.47$\pm$26.93 & 80.85$\pm$18.07 & \textbf{83.40$\pm$20.37}  \\
			glass026Vs7             & 87.80$\pm$16.23  & 85.76$\pm$17.64 & 89.44$\pm$13.47 & \textbf{91.47$\pm$08.54}  & 90.44$\pm$14.49 & 87.67$\pm$18.85 & 82.83$\pm$29.35 & 88.23$\pm$16.05 \\
			zoo01246Vs35            & 94.14$\pm$11.72 & 97.07$\pm$08.79  & 97.42$\pm$03.16  & 97.42$\pm$03.16  & 97.07$\pm$08.79  & 51.13$\pm$34.92 & 96.43$\pm$08.79  & \textbf{100.00$\pm$00.00}  \\
			\hline
			Avearge Rank            & 5.72        & 4.14        & 5.26        & 4.84        & 4.94         & 5.72        & 4.00           & 1.38         \\
			\hline
		\end{tabular}
	}	
\end{table}

In order to compare performance differences between different algorithms better, statistical hypothesis testing is used for further investigation. First of all, the Friedman test based on algorithmic ordering is used to evaluate whether the performance of algorithms is equal. The equation of the Friedman test is written as follows:
\begin{equation}
	\begin{aligned}
		F_F = \frac{(n-1)\tau_{{\chi}^2}}{n(k-1)-\tau_{{\chi}^2}},
	\end{aligned}
\end{equation}
where $k = 8$ is the number of compared algorithms and $n = 25$ is the number of low imbalance datasets, and $\tau_{\chi}$ is calculated by
\begin{equation}
	\begin{aligned}
		\tau_{{\chi}^2} = \frac{12n}{k(k+1)}\left(\sum_{i=1}^{k}r_i^2-\frac{k{(k+1)}^2}{4}\right).
	\end{aligned}
\end{equation}
where the rank $r_i^j$ of $i$-th algorithm in the $j$-th datasets, and the average rank $r_i = \frac{1}{n}\sum_{j-1}^{n}r_i^j $ can be calculated in \autoref{tablelowIR}. Assume that the performances of all algorithms are identical, and then the specific values can be calculated as follows:
\begin{equation}
	\begin{aligned}
		\tau_{{\chi}^2} &= \frac{12\times 25}{8\times 9}(5.72^2+4.14^2+5.26^2+4.84^2 +4.94^2+5.72^2+4.00^2+1.38^2-\frac{8\times 9^2}{4} ) = 58.24\notag
	\end{aligned}
\end{equation}
and
\begin{equation}
	\begin{aligned}
		F_F = \frac{24\times 58.24}{25\times 7-58.24}=11.97.\notag
	\end{aligned}
\end{equation}
Thus, $F_F$ obeys F distribution with degrees of freedom $(8-1) = 7$ and $(8-1)\times (25-1)=168$. When $\alpha = 0.1$, the critical value of the Friedman test with 8 algorithms and 25 datasets is 1.93. Because of $F_F = 11.97 > 1.93$, the assumption is rejected. It can be seen that there are significant differences in performance between different algorithms.

\begin{figure}[b]
	\centering
	\includegraphics{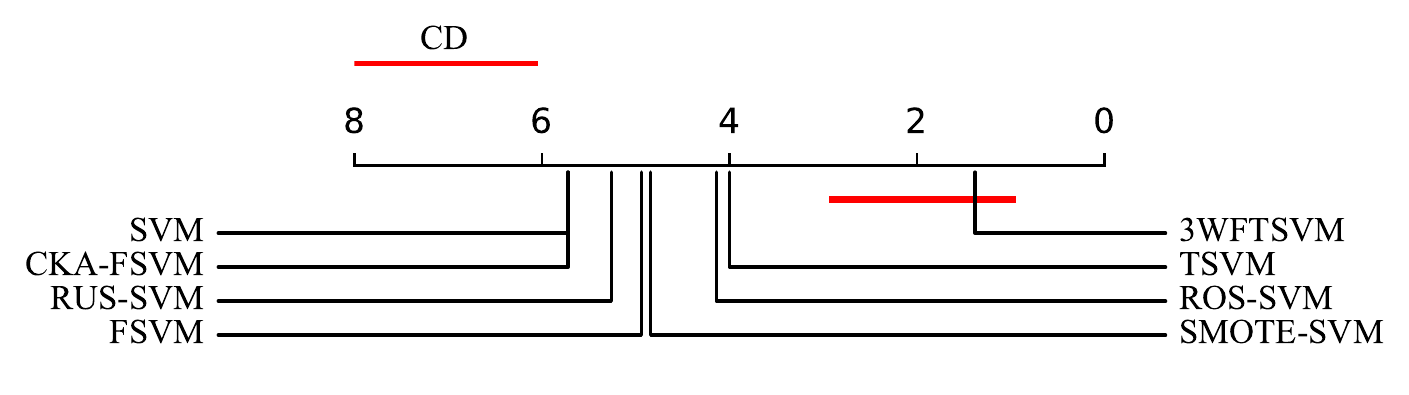}
	\caption{Nemenyi test on low imbalance datasets}
	\label{Nemenyi test on low imbalance datasets}
\end{figure}

To test whether TWFTSVM proposed in this paper is significantly different from compared algorithms, the Nemenyi test, as the post-got test, is used to further distinguish the differences between algorithms. The critical difference $CD_{\alpha } = q_{\alpha}\sqrt{\frac{k(k+1)}{6n}}$ is used to measure the difference of mean ordinal values, where the $q_\alpha$ is determined by $k=8$ and $\alpha = 0.1$. For $q_{\alpha} = 2.78$, we can get critical distance $CD_{\alpha } = 2.78\sqrt{\frac{8\times 9}{6\times 25}}=1.93$. In \autoref{Nemenyi test on low imbalance datasets}, the values in the coordinate axis represent the average rank of algorithms, and the length of red line indicates $CD_{\alpha }$ value if the line of different algorithms has an intersection, which indicates that there is no significant difference among different algorithms. \autoref{Nemenyi test on low imbalance datasets} is shown that TWFTSVM has no intersection with compared algorithms, which means that TWFTSVM outperforms other algorithms. The difference between TSVM ranked 2nd, and TWFTSVM is $4.00 - 1.38 = 2.62 > 1.93$, so the assumption that TWFTSVM and TSVM have the same performance can be rejected. In general, TWFTSVM proposed by this paper performs best in low IR datasets.

\subsection{Experiments on medium imbalanced datasets}

Experiments are designed to evaluate the effectiveness of the proposed TWFTSVM with comparison algorithms ROS-SVM, RUS-SVM, SMOTE-SVM, FSVM, CKA-SVM and TSVM on medium imbalanced datasets. \autoref{tablemediumIR} details the experimental results of 13 medium IR datasets. The best results of different algorithms are shown in bold for each dataset. The average ranks of each algorithm across all datasets are listed in the last row. In \autoref{tablemediumIR}, it is shown that TWFTSVM rank 1st in most datasets and the average rank (1.57) is smaller than other compared algorithms.

\begin{table}[h]\footnotesize
	\caption{G-Means of the compared algorithms on medium imbalanced datasets.(4 $<$ IR $\leq$ 7)} 
	\label{tablemediumIR}
	\resizebox{\linewidth}{!}{
		\begin{tabular}{lllllllll}
			\hline
			& SVM         & ROS-SVM     & RUS-SVM     & SMOTE-SVM   & FSVM & CKA-FSVM    & TSVM        & TWFTSVM       \\
			\hline
			zoo0135Vs246  & 94.14$\pm$11.72 & 89.63$\pm$12.38 & 92.85$\pm$11.35 & 91.21$\pm$13.42 & 88.28$\pm$14.35 & \textbf{97.37$\pm$04.40}   & 94.14$\pm$11.72 & \textbf{97.37$\pm$04.40} \\
			glass02Vs3    & 59.40$\pm$32.70   & 74.39$\pm$20.30  & 61.84$\pm$33.49 & 65.05$\pm$34.86 & 70.25$\pm$32.36 & 71.84$\pm$29.73 & 49.12$\pm$42.54 & \textbf{78.70$\pm$21.82}  \\
			ecoli126Vs345 & 74.34$\pm$16.04 & 84.30$\pm$10.53  & 76.52$\pm$17.31 & 82.07$\pm$18.40  & 80.81$\pm$15.96 & \textbf{86.55$\pm$09.93}  & 78.35$\pm$19.64 & 86.12$\pm$10.49 \\
			ecoli124Vs6   & 80.05$\pm$16.10  & 85.02$\pm$19.71 & 85.59$\pm$19.54 & \textbf{87.28$\pm$17.98} & 85.36$\pm$19.72 & 80.31$\pm$13.84 & 83.31$\pm$16.86 & 85.39$\pm$19.58 \\
			zoo01456Vs23  & 94.14$\pm$11.72 & 90.10$\pm$12.64  & 94.14$\pm$11.72 & 93.57$\pm$11.55 & 93.99$\pm$08.29  & 91.44$\pm$12.91 & 95.85$\pm$08.72  & \textbf{96.43$\pm$08.79}  \\
			zoo01234Vs56  & \textbf{100.00$\pm$00.00}   & \textbf{100.00$\pm$00.00}   & \textbf{100.00$\pm$00.00}   & \textbf{100.00$\pm$00.00}   & 87.07$\pm$30.31 & 77.07$\pm$39.50  & 97.07$\pm$00.00   & \textbf{100.00$\pm$00.00}   \\
			zoo01256Vs34  & \textbf{99.35$\pm$01.94}  & \textbf{99.35$\pm$01.94}  & \textbf{99.35$\pm$01.94}  & 98.71$\pm$02.58  & 96.43$\pm$08.79  & 96.43$\pm$08.79  & 96.43$\pm$08.79  & \textbf{99.35$\pm$01.94}  \\
			glass026Vs3   & 62.96$\pm$42.59 & 76.38$\pm$28.61 & 73.83$\pm$27.01 & 73.73$\pm$27.42 & 78.70$\pm$18.67  & 76.95$\pm$28.30  & 63.57$\pm$42.99 & \textbf{80.51$\pm$15.94} \\
			glass012Vs7   & 75.52$\pm$38.87 & 91.12$\pm$12.63 & 81.06$\pm$30.73 & 79.39$\pm$30.97 & 92.86$\pm$12.96 & 84.64$\pm$30.91 & 90.65$\pm$14.77 & \textbf{94.01$\pm$08.48}  \\
			ecoli1256Vs34 & 78.66$\pm$11.89 & 83.85$\pm$17.06 & 79.76$\pm$20.65 & 80.71$\pm$19.94 & 80.10$\pm$15.34  & 79.55$\pm$16.24 & 76.04$\pm$16.68 & \textbf{83.89$\pm$14.17} \\
			glass017Vs3   & 25.81$\pm$40.42 & 58.10$\pm$22.90   & 35.40$\pm$44.68  & 52.66$\pm$29.62 & 40.75$\pm$35.21 & 31.16$\pm$34.34 & 63.08$\pm$34.34  & \textbf{66.04$\pm$20.95} \\
			ecoli146Vs3   & 97.11$\pm$05.80   & 97.11$\pm$05.80   & 97.96$\pm$03.93  & 98.19$\pm$03.97  & 98.17$\pm$04.12  & \textbf{98.42$\pm$04.00}   & 97.11$\pm$05.80   & \textbf{98.42$\pm$04.00}   \\
			glass6        & 84.64$\pm$29.01 & 79.86$\pm$31.84 & 82.25$\pm$30.12 & 82.25$\pm$30.12 & 83.09$\pm$28.99 & 91.28$\pm$10.57 & 75.47$\pm$38.59 & \textbf{91.44$\pm$10.11} \\
			\hline
			Average Rank  & 5.88        & 4.34        & 4.73        & 4.50        & 4.69            & 4.53        & 5.73        & 1.57           \\
			\hline
		\end{tabular}
	}
\end{table}

\begin{figure}[b]
	\centering
	\includegraphics{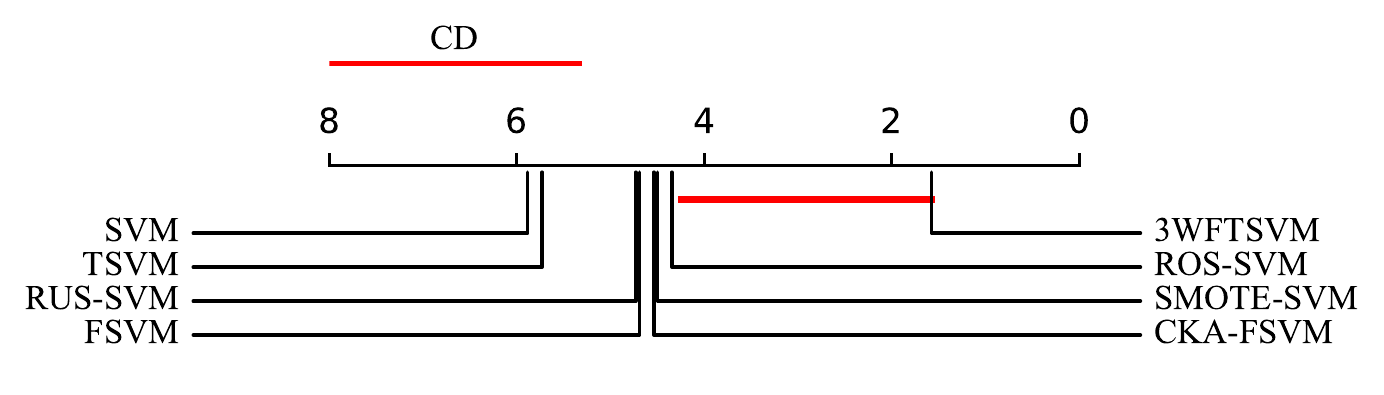}
	\caption{Nemenyi test on medium imbalance datasets}
	\label{Nemenyi test on medium imbalance datasets}
\end{figure}

The Friedman test is calculated as a Statistical test to explore performance differences between different algorithms further. We can calculate the
\begin{equation}
	\begin{aligned}
		\tau_{{\chi}^2} &= \frac{12\times 13}{8\times 9}(5.88^2+4.34^2+4.73^2+4.50^2 +4.69^2+4.53^2+5.73^2+1.57^2-\frac{8\times 9^2}{4}) = 26.19\notag
	\end{aligned}
\end{equation}
and
\begin{equation}
	\begin{aligned}
		F_F = \frac{12\times 26.19}{13\times 7-26.19}=4.85.\notag
	\end{aligned}
\end{equation}

Thus, $F_F$ obeys F distribution with degrees of freedom $(8-1) = 7$ and $(8-1)\times (13-1)=84$. When $\alpha = 0.1$, the critical value of the Friedman test with 8 algorithms and 13 datasets is 2.20. We can reject 
the assumption that different algorithms have the same performance because $F_F = 4.85 > 2.20$. The Nemenyi test, as the post-got test, is used to further distinguish the differences between algorithms. The critical difference $CD_{\alpha } = q_{\alpha}\sqrt{\frac{k(k+1)}{6n}}$ is used to measure the difference of mean ordinal values, where $q_\alpha$ is determined by $k=8$ and $\alpha = 0.1$. For $q_{\alpha} = 2.78$, we can get critical distance $CD_{\alpha } = 2.78\sqrt{\frac{8\times 9}{6\times 13}}=2.67$. In \autoref{Nemenyi test on medium imbalance datasets}, the values in the coordinate axis represent the average rank of algorithms, and the length of the red line indicates $CD_{\alpha }$ value. If the line of different algorithms has an intersection, which indicates that there is no significant difference among different algorithms. \autoref{Nemenyi test on medium imbalance datasets} shows that TWFTSVM has no intersection with other algorithms, which means that TWFTSVM outperforms other algorithms. For ROS-SVM rank 2nd, we can conclude that TWFTSVM performs better because the difference between the average rank of ROS-SVM and TWFTSVM is $4.34 - 1.57 = 2.77 > 2.67$. Therefore, TWFTSVM performs best than compared algorithms on 13 medium IR datasets under the statistical hypothesis test.

\subsection{Experiments on  high imbalanced datasets}

In order to investigate the effectiveness of the proposed TWFTSVM with comparison algorithms ROS-SVM, RUS-SVM, SMOTE-SVM, FSVM, CKA-SVM and TSVM, experiments are designed on high-imbalanced datasets.  \autoref{tablehighIR} details experimental results of 9 high IR datasets. The best results of different algorithms are shown in bold for each dataset. The average ranks of each algorithm across all datasets are listed in the last row. In \autoref{tablehighIR}, it is shown that TWFTSVM rank 1st in most datasets and the average rank (1.22) is smaller than other compared algorithms.

\begin{table}[htbp]\footnotesize
	\caption{G-Means of the compared algorithms on high imbalanced datasets.(IR $>$ 7)} 
	\label{tablehighIR}
	\resizebox{\linewidth}{!}{
		\begin{tabular}{lllllllll}
			\hline
			& SVM         & ROS-SVM     & RUS-SVM     & SMOTE-SVM   & FSVM & CKA-FSVM    & TSVM        & TWFTSVM       \\
			\hline
			ecoli1Vs4     & 97.07$\pm$08.79  & 98.96$\pm$01.59  & 98.96$\pm$01.59  & 98.96$\pm$01.59  & 85.10$\pm$14.92  & 85.10$\pm$14.92  & 97.07$\pm$08.79  & \textbf{100.00$\pm$00.00}   \\
			ecoli1246Vs35 & 65.77$\pm$24.96 & 78.20$\pm$29.45  & 84.37$\pm$22.42 & 82.34$\pm$22.67 & 83.26$\pm$19.92 & 78.58$\pm$16.54 & 84.50$\pm$13.37  & \textbf{87.83$\pm$15.87} \\
			ecoli126Vs3   &71.50$\pm$10.05  & 86.94$\pm$14.37 & 72.47$\pm$14.44 & 72.84$\pm$13.61 & 75.07$\pm$18.44 & 70.30$\pm$13.37 & 85.49$\pm$13.03 &\textbf{88.83$\pm$09.39} \\
			ecoli3        & 76.03$\pm$18.01 & 80.77$\pm$22.57 & 86.46$\pm$14.66 & 82.19$\pm$16.65 & 84.78$\pm$20.31 & 86.08$\pm$14.69 & 74.58$\pm$27.32 & \textbf{88.46$\pm$10.23} \\
			ecoli1236Vs45 & 90.85$\pm$14.65 & 92.94$\pm$10.44 & 89.17$\pm$14.68 & 94.04$\pm$09.50 & 95.32$\pm$06.94 & 95.89$\pm$06.33 & 89.41$\pm$15.48 & \textbf{98.49$\pm$02.05}  \\
			ecoli123Vs4   &94.14$\pm$11.72  & 96.68$\pm$08.70 & \textbf{99.60$\pm$00.79} & 99.21$\pm$00.97 & 98.61$\pm$01.26 & 98.81$\pm$00.97 & 97.07$\pm$08.79 &  99.10$\pm$02.31 \\
			ecoli126Vs4   & 90.84$\pm$13.22 & 87.54$\pm$13.02 & 88.46$\pm$14.56 & 92.81$\pm$11.30  & 93.87$\pm$09.87  & 94.45$\pm$09.97  & 87.91$\pm$14.08 & \textbf{98.86$\pm$02.30} \\
			glass4        & 64.63$\pm$43.28 & 76.51$\pm$38.67 & 84.42$\pm$29.56 & 87.07$\pm$30.31 & 90.00$\pm$30.00   & 77.07$\pm$39.50  & 77.07$\pm$39.50  & \textbf{90.61$\pm$11.83} \\
			ecoli5        & 87.96$\pm$14.11 & 93.58$\pm$12.09 & 86.45$\pm$12.52 & \textbf{94.16$\pm$09.27}  & 90.67$\pm$11.61 & 93.98$\pm$09.41  & 90.56$\pm$13.07 & 93.88$\pm$09.56  \\
			\hline
			Average Rank  & 6.83        & 5.44        & 4.66        & 3.66        & 4.16         & 4.44        & 5.55        & 1.22           \\
			\hline
		\end{tabular}
	}
\end{table}

\begin{figure}[b]
	\centering
	\includegraphics{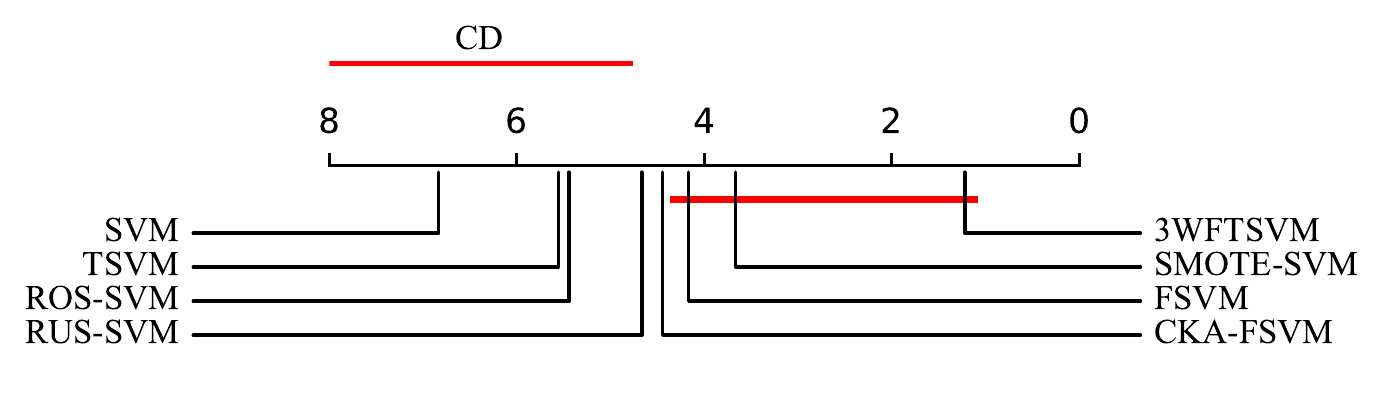}
	\caption{Nemenyi test on high imbalance datasets}
	\label{Nemenyi test on high imbalance datasets}
\end{figure}

To further explore performance differences between different algorithms, the Friedman test is used as calculated as Statistical test. We can calculated the
\begin{equation}
	\begin{aligned}
		\chi_F^2 &= \frac{12\times 7}{8\times 9}(6.83^2+5.44^2+4.66^2+3.66^2 +4.16^2+4.44^2+5.55^2+1.22^2-\frac{8\times 9^2}{4}) = 18.89\notag
	\end{aligned}
\end{equation}
and
\begin{equation}
	\begin{aligned}
		F_F = \frac{6\times 18.89}{7\times 7-18.89}=3.77.\notag
	\end{aligned}
\end{equation}

Thus, $F_F$ obeys F distribution with degrees of freedom $(8-1) = 7$ and $(8-1)\times (9-1)=56$. When $\alpha = 0.1$, the critical value of the Friedman test with 8 algorithms and 9 datasets is 2.47. We can reject the assumption that different algorithms perform similarly because $F_F = 3.77 > 2.47$. Nemenyi test as the post-goc test is used to further distinguish the differences between algorithms. The critical difference $CD_{\alpha } = q_{\alpha}\sqrt{\frac{k(k+1)}{6n}}$ is used to measure the difference of mean ordinal values, where $q_\alpha$ is determined by $k=8$ and $\alpha = 0.1$. For $q_{\alpha} = 2.78$, we can get critical distance $CD_{\alpha } = 2.78\sqrt{\frac{8\times 9}{6\times 9}}=3.21$. In \autoref{Nemenyi test on high imbalance datasets}, the values in the coordinate axis represent the average rank of algorithms, and the length of the red line indicates $CD_{\alpha }$ value. Suppose the line of different algorithms has an intersection, indicating no significant difference among different algorithms.

\autoref{Nemenyi test on high imbalance datasets} shows that TWFTSVM intersects with SMOTE-SVM and FSVM, meaning there is no significant difference in performance between TWFTSVM, SMOTE-SVM and FSVM. TWFTSVM has no intersection with ROS-SVM, TSVM and SVM. In conclusion, TWFTSVM performs best on 9 high IR datasets than ROS-SVM, TSVM and SVM  under the statistical hypothesis test.

\subsection{G-Means with different imbalance ratio in same dataset}

To further measure the performance of TWFTSVM, this paper reprocesses seven train sets with different IRs in the same dataset for comparison experiments. In each dataset, some positive and negative samples are randomly selected to make IR a specific value, and IR is selected from $\{2, 3, 4, 5, 6, 7, 8\}$. Due to space constraints, eight datasets in \autoref{tabledatasets} are selected for the presentation of comparison experiments.

\begin{figure}[htbp]\footnotesize
	\centering
	\subfigure[ecoli1246Vs35]{
		\includegraphics[width=8cm]{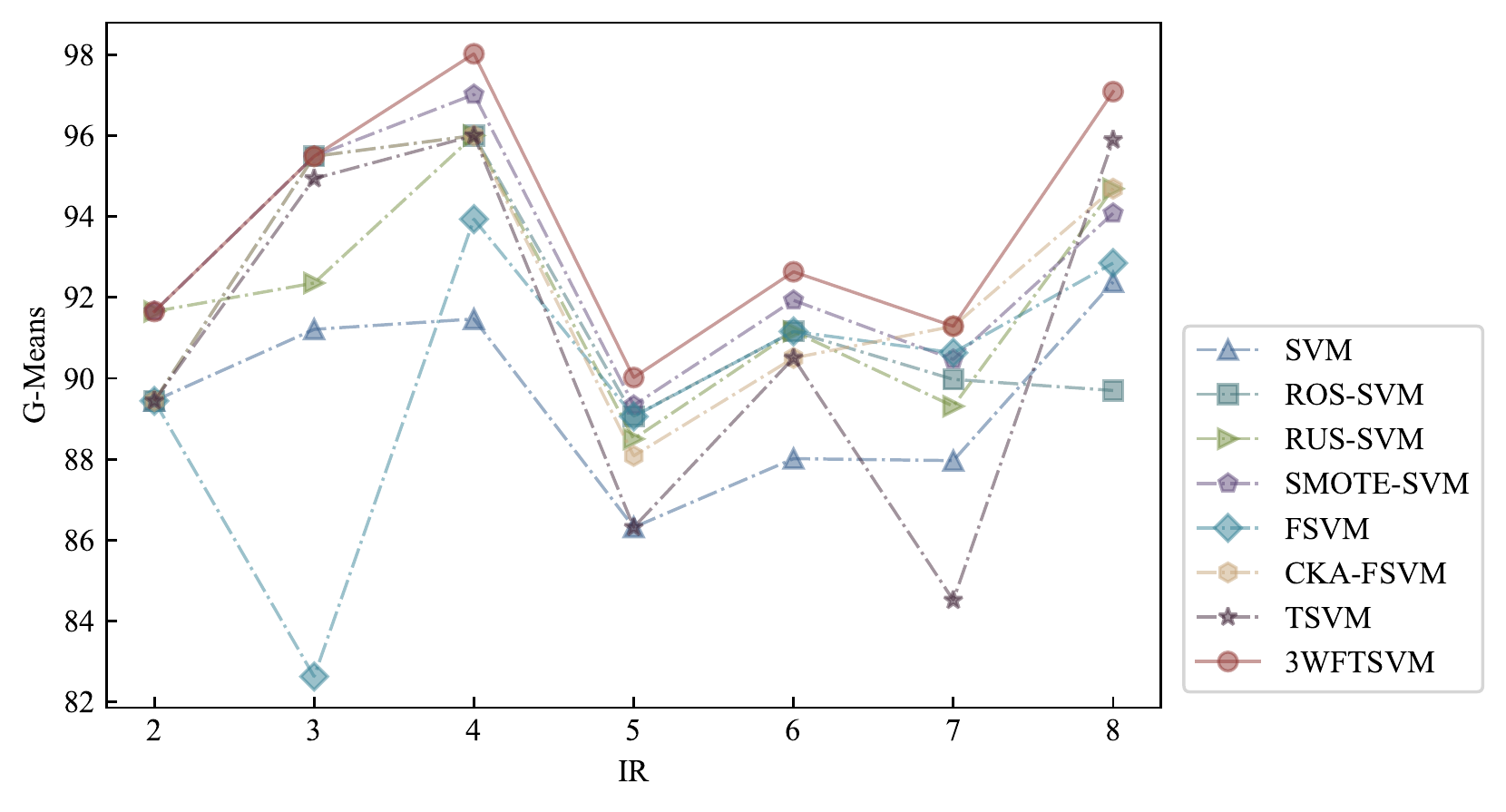}\label{ecoli246Vs35ir}
	}
	\hskip -10pt 
	\subfigure[glass2]{
		\includegraphics[width=8cm]{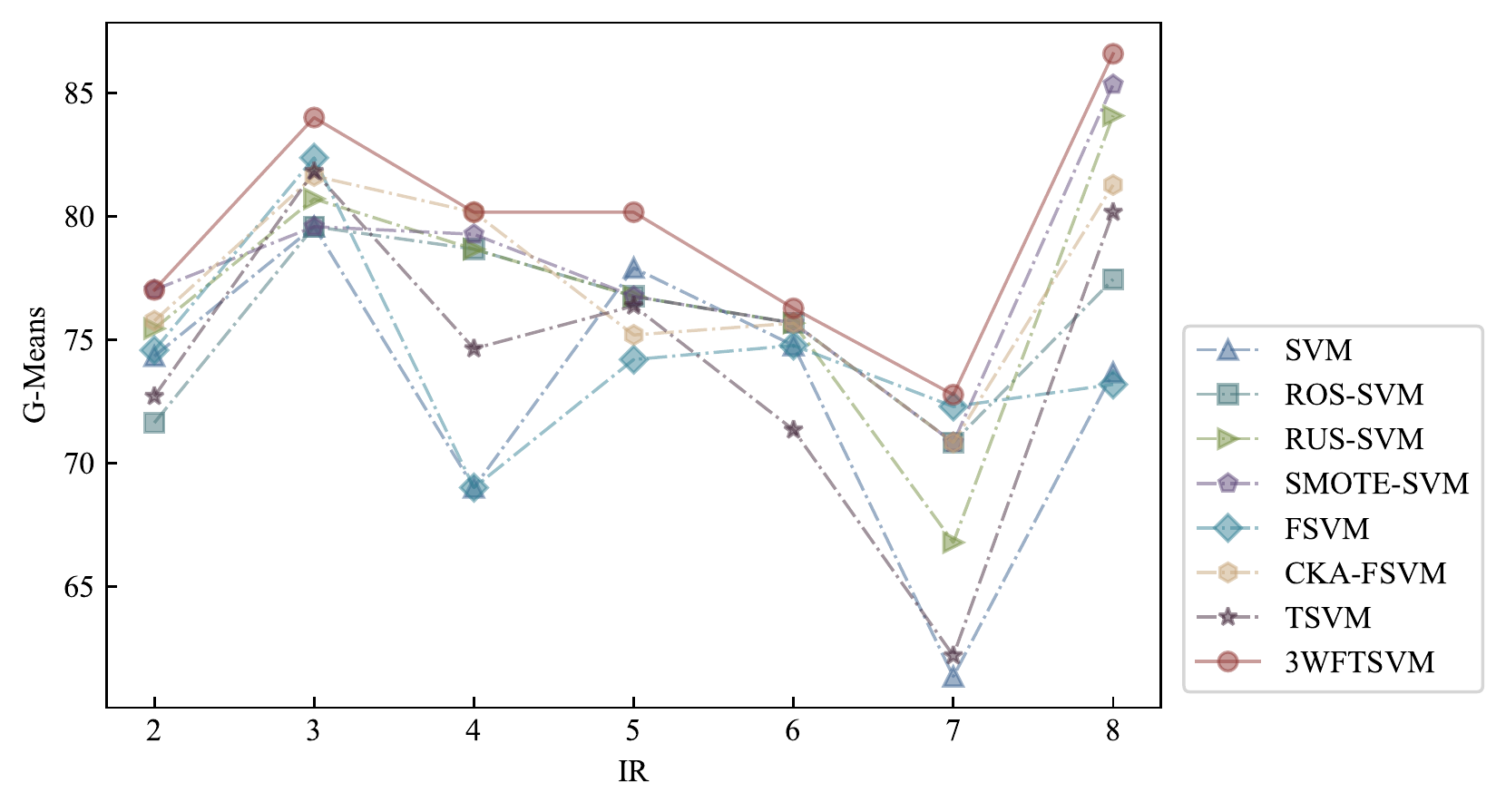}\label{glass2ir}
	}
	\vskip -0pt 
	\subfigure[heart-statlog]{
		\includegraphics[width=8cm]{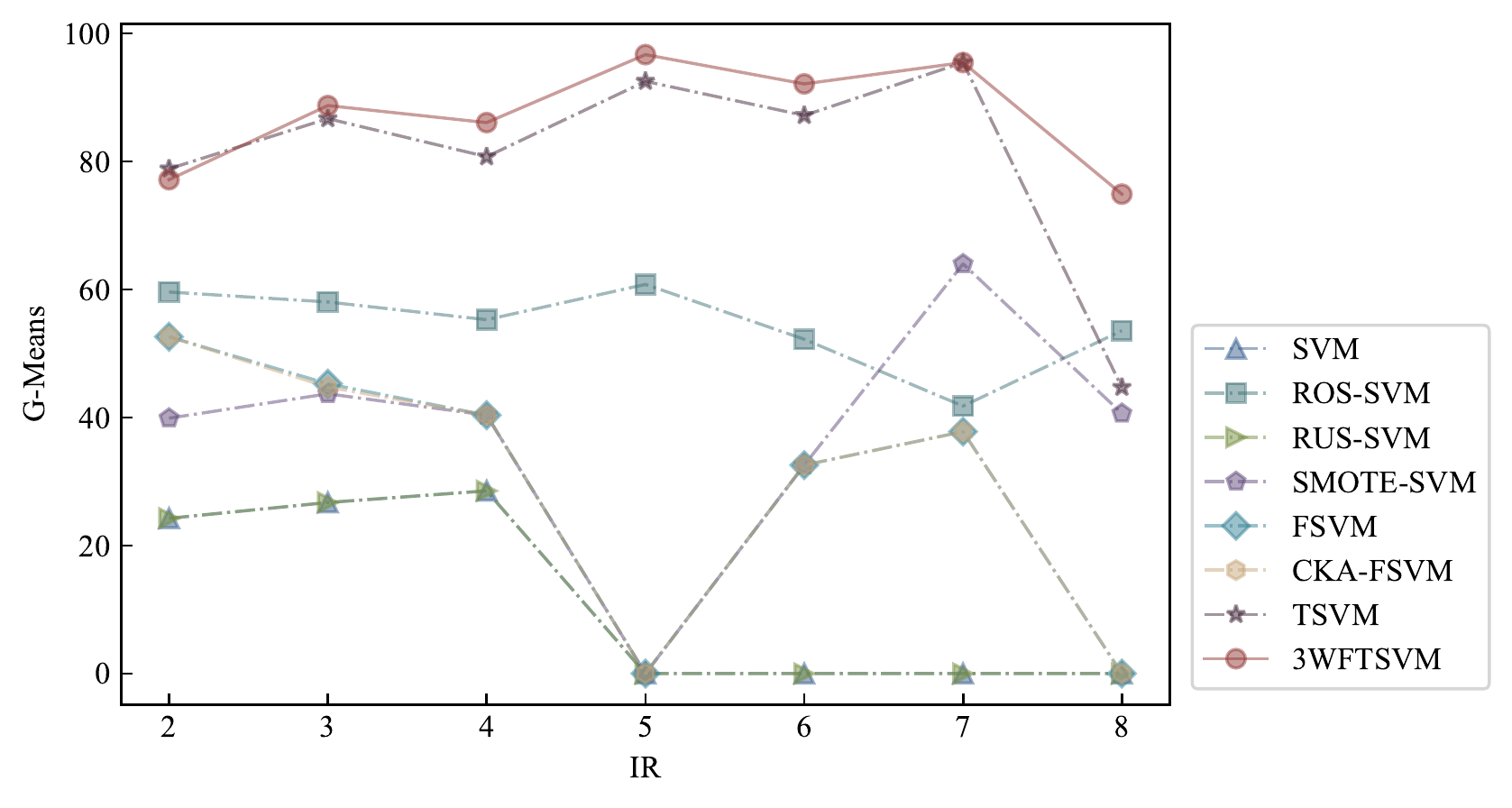}\label{heart-statlogir}
	}
	\hskip -10pt  
	\subfigure[vehicle1]{
		\includegraphics[width=8cm]{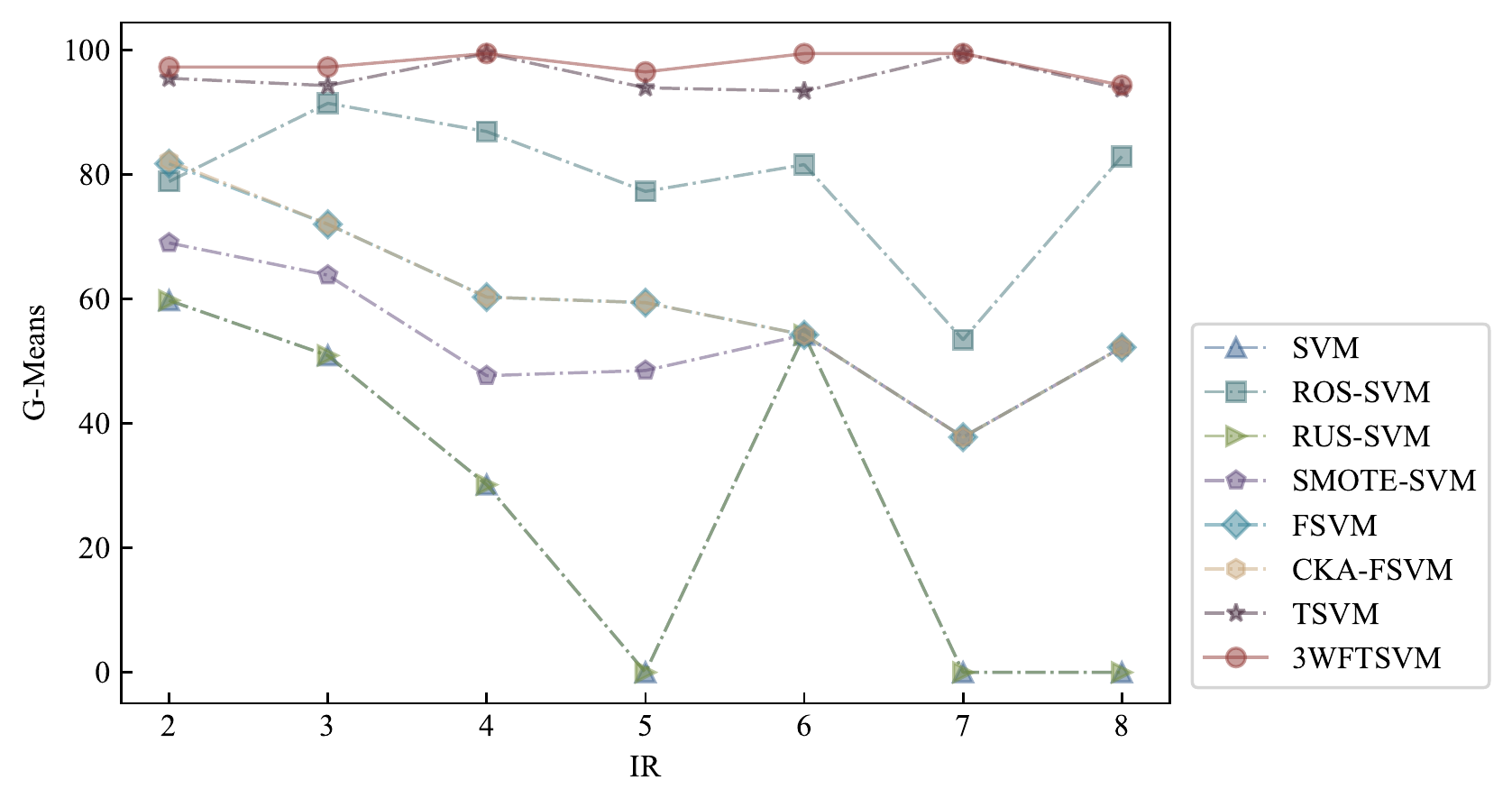}\label{vehicle1ir}
	}
	\vskip -0pt 
	\subfigure[glass6]{
		\includegraphics[width=8cm]{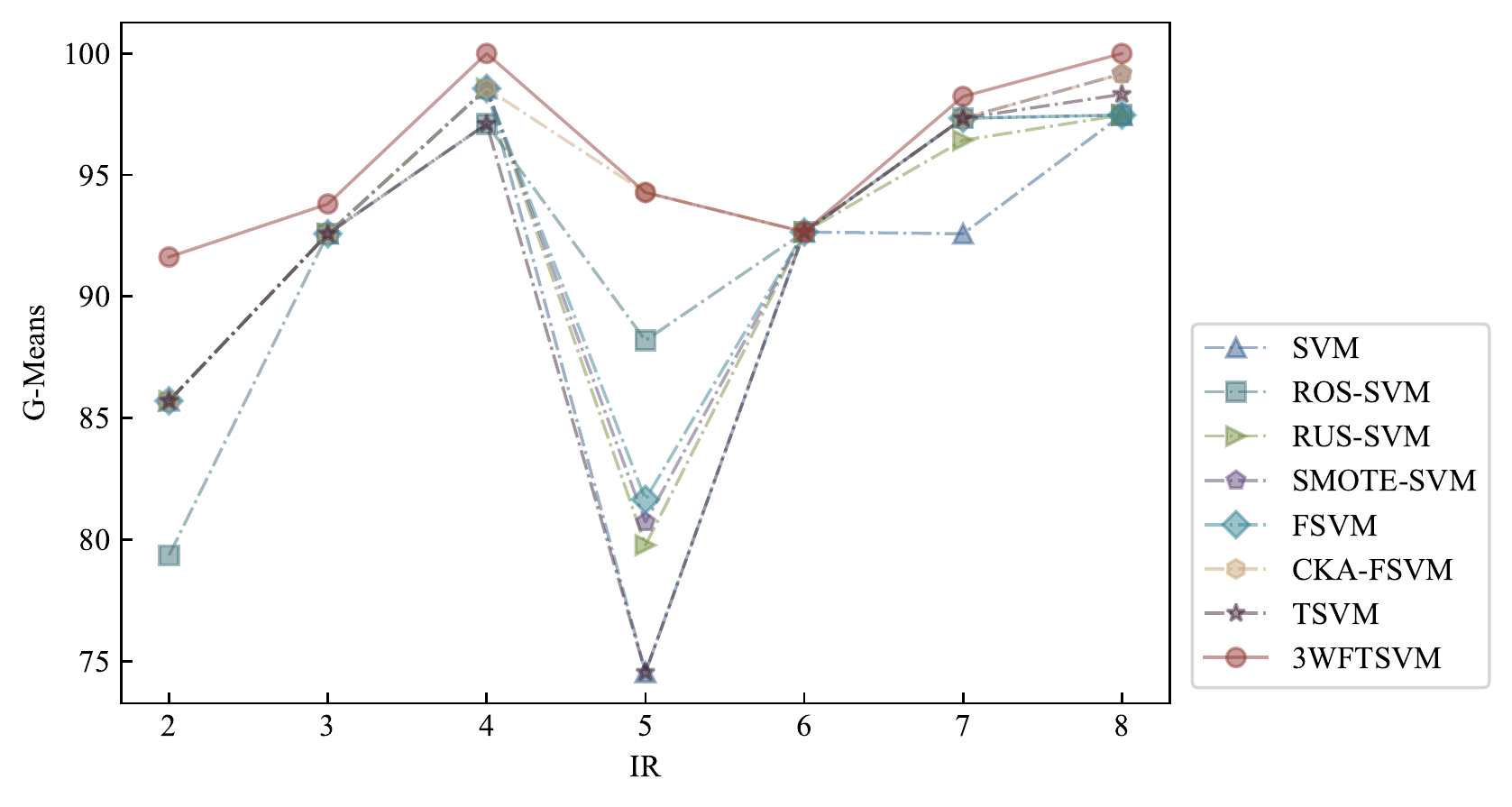}\label{glass6ir}
	}
	\hskip -10pt  
	\subfigure[seeds3]{
		\includegraphics[width=8cm]{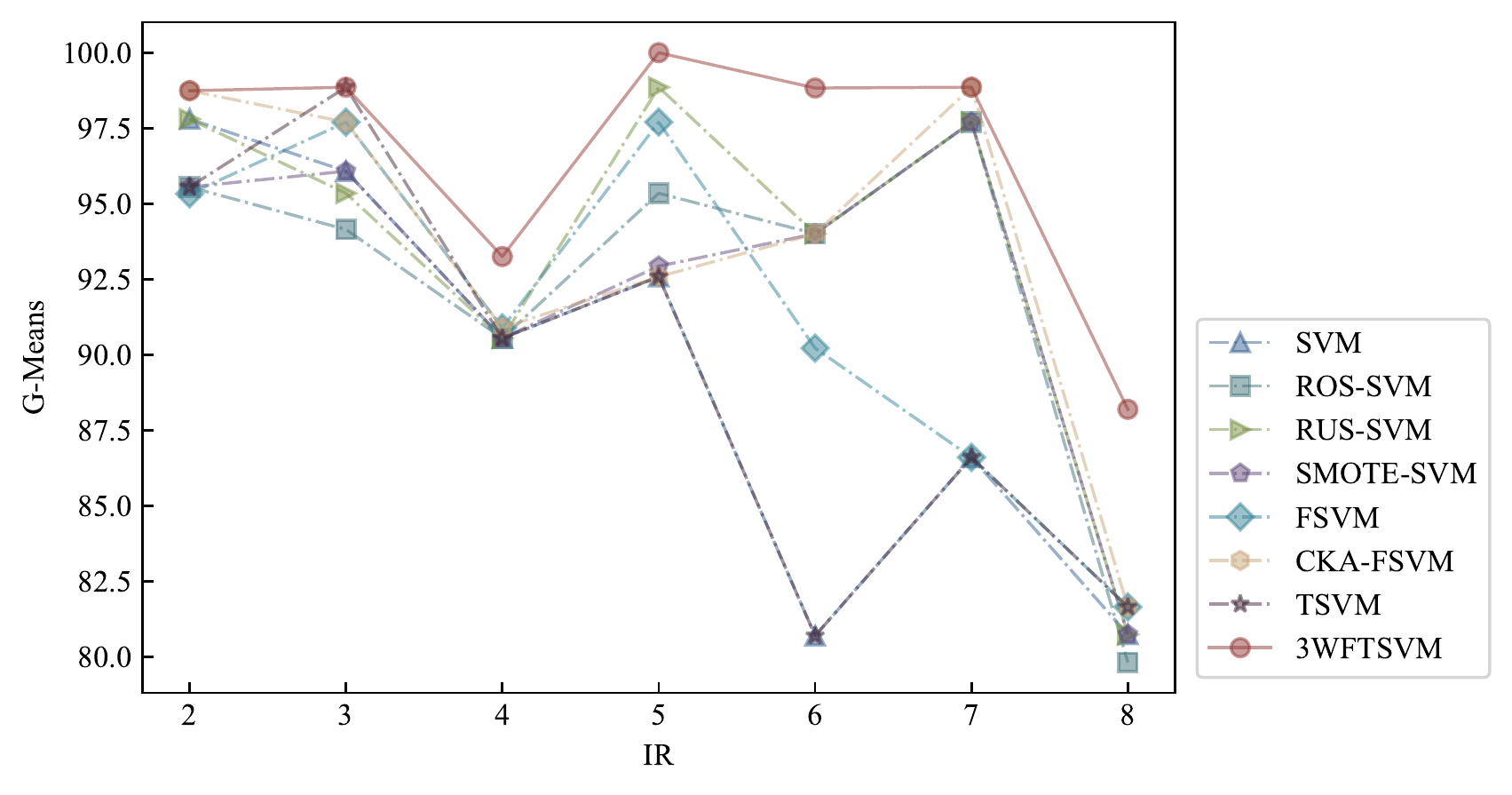}\label{seeds3ir}
	}
	\vskip -0pt 
	\subfigure[breast-cancer-wisconsin]{
		\includegraphics[width=8cm]{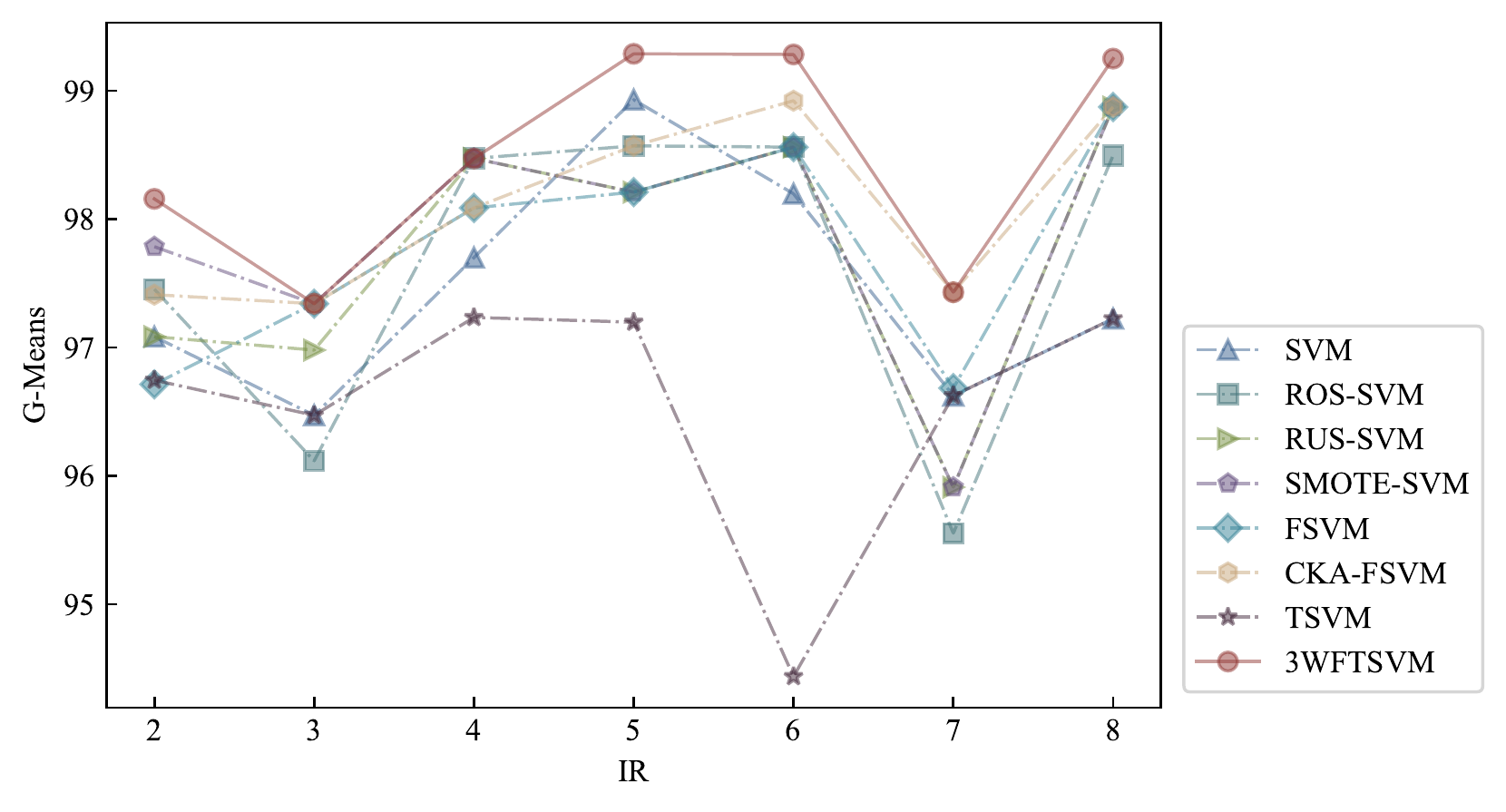}\label{breast-cancer-wisconsinir}
	}
	\hskip -10pt  
	\subfigure[ecoli1346Vs25]{
		\includegraphics[width=8cm]{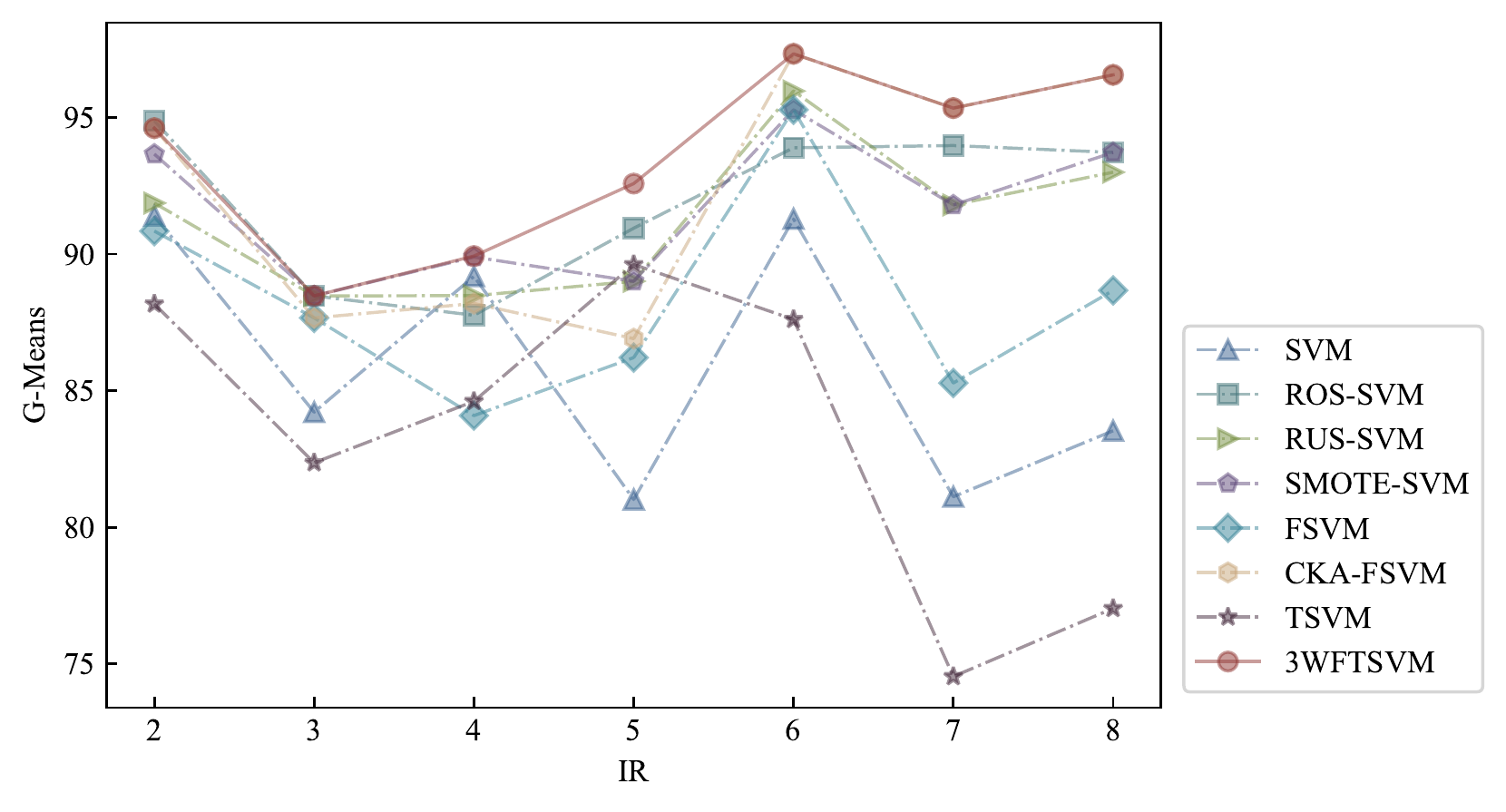}\label{ecoli1346Vs25ir}
	}
	\caption{Comparisons on different imbalanced IRs}
	\label{Compared IR}
\end{figure}

Experimental results are shown in \autoref{Compared IR}. In \autoref{ecoli246Vs35ir}, it can be found that FSVM does not perform well when IR $=3$ and TSVM performs well, but TSVM performs worse than FSVM when IR $=7$, so different algorithms perform differently at different IRs. TWFTSVM outperforms other algorithms at different IRs. When the original dataset is from low IR to medium and high IR, the performance of SVM, RUS-SVM, SMOTE-SVM and CKA-SVM become worse in \autoref{heart-statlogir} and \autoref{vehicle1ir}. When the number of positive samples is small, the G-Means of some algorithms, such as RUS-SVM, will be zero, which is caused by zero accuracies in identifying positive models. Because TWFTSVM has two hyperplanes for positive and negative samples, G-Means can not be zero. TWFTSVM has better generalization than other algorithms. In \autoref{Compared IR}, it can be seen that the G-Means of different algorithms do not decrease when the IR of the dataset increases. Because the original IR of other datasets differs, the G-Means fluctuation of TWFTSVM differs in different sub-graphs. When IR $=5$ in \autoref{glass6ir}, the G-Means of TWFTSVM is far superior to other algorithms. In conclusion, TWFTSVM performers best compared with other algorithms for the same dataset with different IRs.

\begin{table}[htbp]\footnotesize
	\caption{Optimal parameters of TWFTSVM} 
	\label{oppara}
	\resizebox{\linewidth}{!}{
		\begin{tabular}{llllllllll}
			\hline
			Dataset & $\sigma$ & $C_{13}$ & $C_{24}$ & $\alpha$ & Dataset & $\sigma$ & $C_{13}$ & $C_{24}$ & $\alpha$ \\ 
			\hline
			\textbf{Low imbalance}(IR $\leq$ 4.0)          &      &      &       &      &       &               &      &      &       \\
			heart-statlog & 0.01 & 32 & 32 & 0.900 & haberman & 10 & 1 & 0.5 & 0.929 \\
			pima & 10 & 1 & 32 & 0.917 & glass015Vs7 & 1 & 0.5 & 0.5 & 0.875 \\
			liver-disorders & 10 & 4 & 1 & 0.929 & vehicle1 & 1 & 0.5 & 8 & 0.917 \\ 
			zoo0 & 0.01 & 0.5 & 0.5 & 0.875 & vehicle2 & 10 & 32 & 8 & 0.875 \\ 
			glass2 & 1 & 1 & 8 & 0.917 & glass013Vs7 & 0.01 & 0.5 & 0.5 & 0.875\\ 
			ecoli1Vs2 & 0.01 & 1 & 8 & 0.917 & ecoli1346Vs25 & 1 & 1 & 1 & 0.917\\ 
			breast-cancer-wisconsin & 10 & 4 & 8 & 0.917 & glass025Vs7 & 0.1 & 4 & 0.5 & 0.900  \\ 
			wisc & 0.01 & 16 & 0.5 & 0.900 & zoo02356Vs14 & 0.01 & 1 & 1 & 0.917 \\ 
			seeds3 & 10 & 0.5 & 8 & 0.900 & ecoli123Vs456 & 0.1 & 2 & 0.5 & 0.929\\ 
			wine1 & 1 & 4 & 1 & 0.875 & ecoli1 & 0.01 & 8 & 32 & 0.929\\ 
			zoo02346Vs15 & 1 & 0.5 & 0.5 & 0.875 & glass026Vs7 & 0.1 & 8 & 0.5 & 0.875 \\ 
			wine & 1 & 0.5 & 0.5 & 0.875 & zoo01246Vs35 & 0.01 & 1 & 8 & 0.900\\ 
			ecoli1Vs6 & 0.1 & 32 & 32 & 0.875 &  &  &  &  &  \\ 
			\textbf{Medium imbalance} (4.0 < IR $\leq$ 7.0)          &      &      &       &      &       &               &      &      &       \\
			zoo0135Vs246 & 1 & 1 & 0.5 & 0.929 & glass026Vs3 & 1 & 0.5 & 2 & 0.917 \\ 
			glass02Vs3 & 1 & 0.5 & 16 & 0.875 & glass012Vs7 & 0.01 & 0.5 & 2 & 0.875 \\ 
			ecoli126Vs345 & 0.1 & 1 & 4 & 0.875 & ecoli1256Vs34 & 0.1 & 2 & 8 & 0.875 \\ 
			ecoli124Vs6 & 1 & 0.5 & 32 & 0.917 & glass017Vs3 & 0.01 & 0.5 & 2 & 0.917 \\ 
			zoo01456Vs23 & 0.01 & 0.5 & 0.5 & 0.900 & ecoli146Vs3 & 0.01 & 1 & 4 & 0.875 \\ 
			zoo01234Vs56 & 0.01 & 0.5 & 0.5 & 0.875 & glass6 & 1 & 2 & 1 & 0.875 \\ 
			zoo01256Vs34 & 1 & 0.5 & 0.5 & 0.875 &  &  &  &  & \\ 
			\textbf{High imbalance} (IR $>$ 7.0)          &      &      &       &      &       &               &      &      &       \\
			ecoli1Vs4 & 0.01 & 0.5 & 0.5 & 0.875 & ecoli123Vs4 & 0.1 & 0.5 & 4 & 0.875 \\ 
			ecoli1246Vs35 & 1 & 0.5 & 2 & 0.929 & ecoli126Vs4 & 0.01 & 4 & 16 & 0.929 \\ 
			ecoli126Vs3 & 1 & 0.5 & 1 & 0.875 & glass4 & 1 & 0.5 & 0.5 & 0.875 \\ 
			ecoli3 & 0.1 & 0.5 & 2 & 0.875 & ecoli5 & 0.01 & 1 & 16 & 0.875 \\ 
			ecoli1236Vs45 & 0.01 & 1 & 4 & 0.875 & & & & & \\ 
			\hline
		\end{tabular}
	}
\end{table}

\subsection{Parameter discussion}

In TWFTSVM, there are some parameters, i.e. parameter $\sigma$ of Gaussian kernel function, threshold $(\alpha, \beta)$, regularization parameter $C_{1}, C_{3}$ and penalty parameter $C_{2}, C_{4}$ which impacting the classification performance. We present the optimal parameters of TWFTSVM for comparison experiments in \autoref{oppara}. The optimal parameters are different in different datasets, which indicates that different parameters have a greater impact on the experimental results. To investigate the influence of $(\alpha, \beta)$, $C_{1}$, $C_{3}$, $C_{2}$ and $C_{4}$ to TWFTSVM, the classification performances of TWFTSVM on 30 imbalanced datasets are further discussed. The 30 imbalanced datasets include low, medium and high imbalance datasets selected from \autoref{tabledatasets}. When discussing $(\alpha, \beta)$, $\alpha$ is selected from  $\{0.875, 0.900, 0.917, 0.929\}$ and $\beta $ is selected from $\{0.125, 0.100, 0.083, 0.071\}$. When discussing on regularization parameter $C_{1}, C_{3}$ and penalty parameter $C_{2}, C_{4}$ are selected from $\{2^{-1}, 2^{1}, \dots , 2^{4}, 2^{5}\}$.

\begin{figure}[htbp]\footnotesize
	\centering
	\subfigure{
		\includegraphics[width=7.8cm]{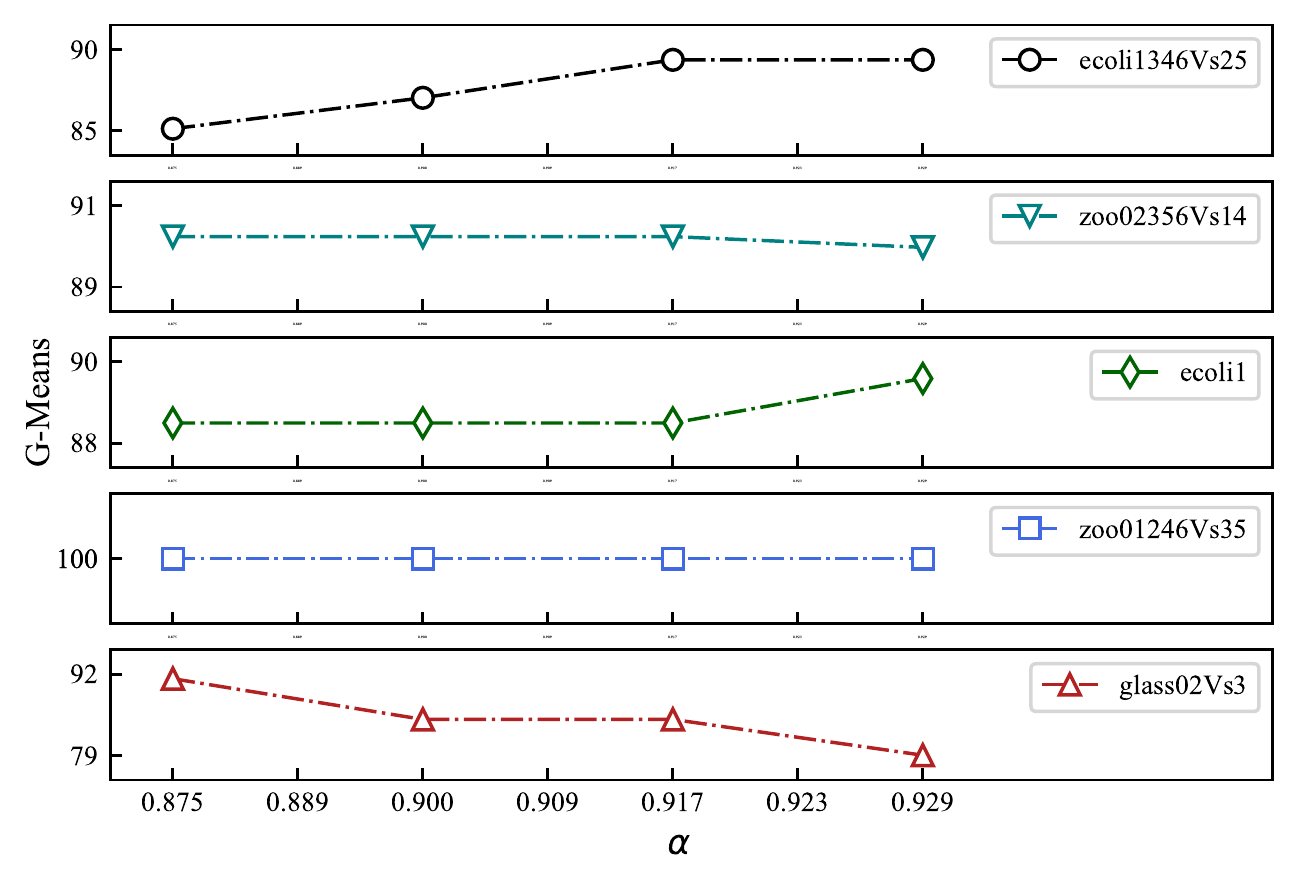} 
	}
	\hskip 5pt
	\subfigure{
		\includegraphics[width=7.8cm]{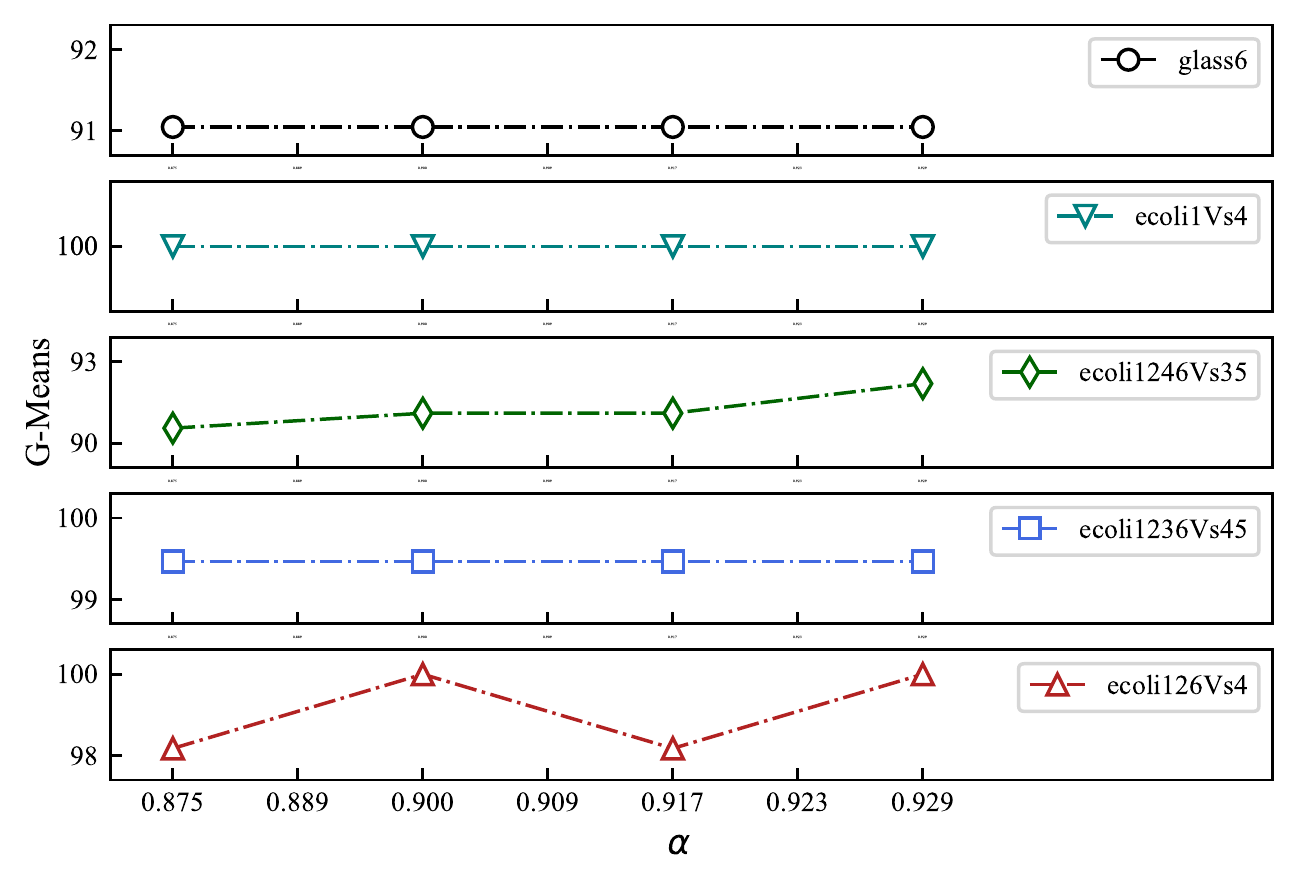}
	}
	\vskip -0pt  
	\subfigure{
		\includegraphics[width=7.8cm]{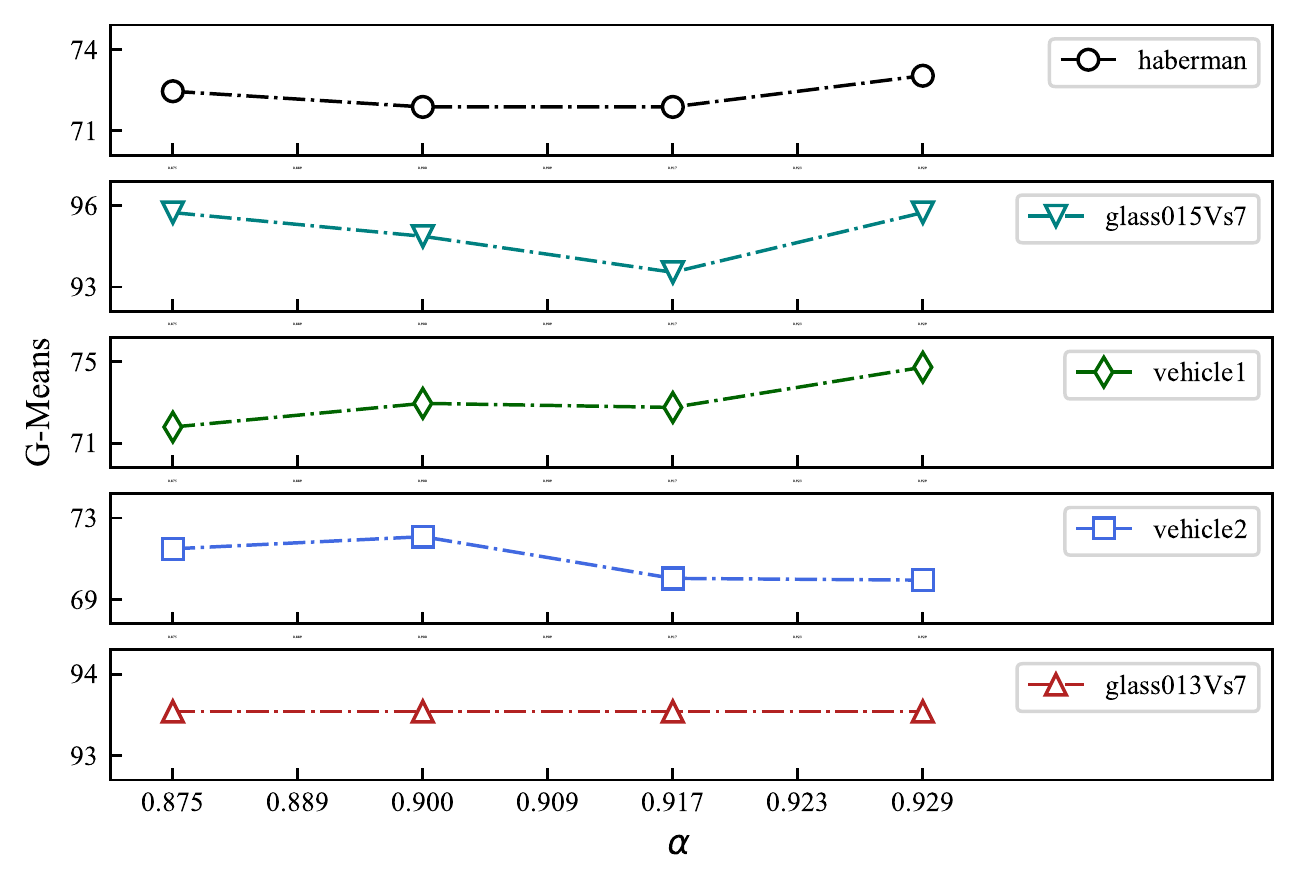}
	}
	\hskip 5pt
	\subfigure{
		\includegraphics[width=7.8cm]{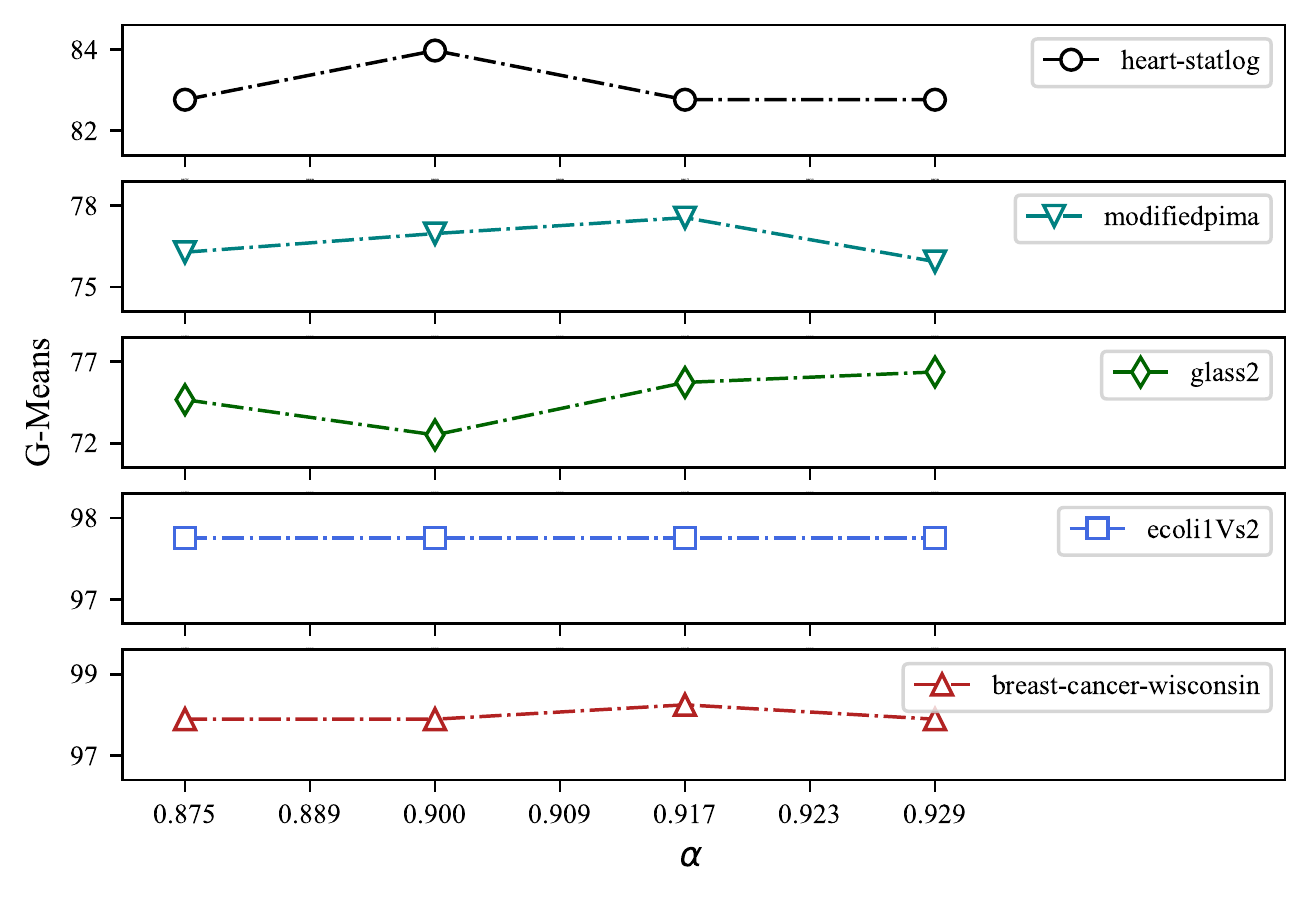} 
	}
	\vskip -0pt  
	\subfigure{
		\includegraphics[width=7.8cm]{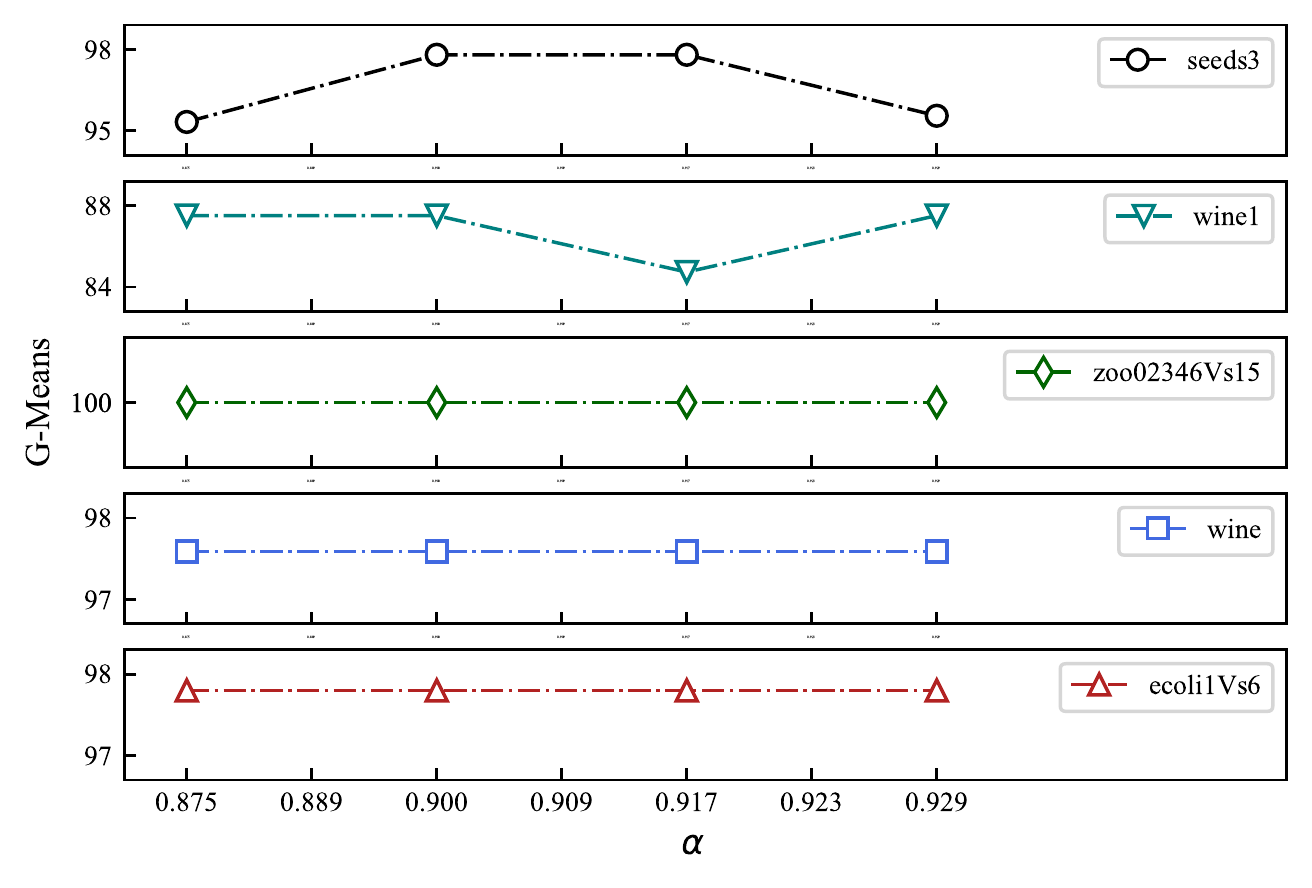}
	}
	\hskip 5pt
	\subfigure{
		\includegraphics[width=7.8cm]{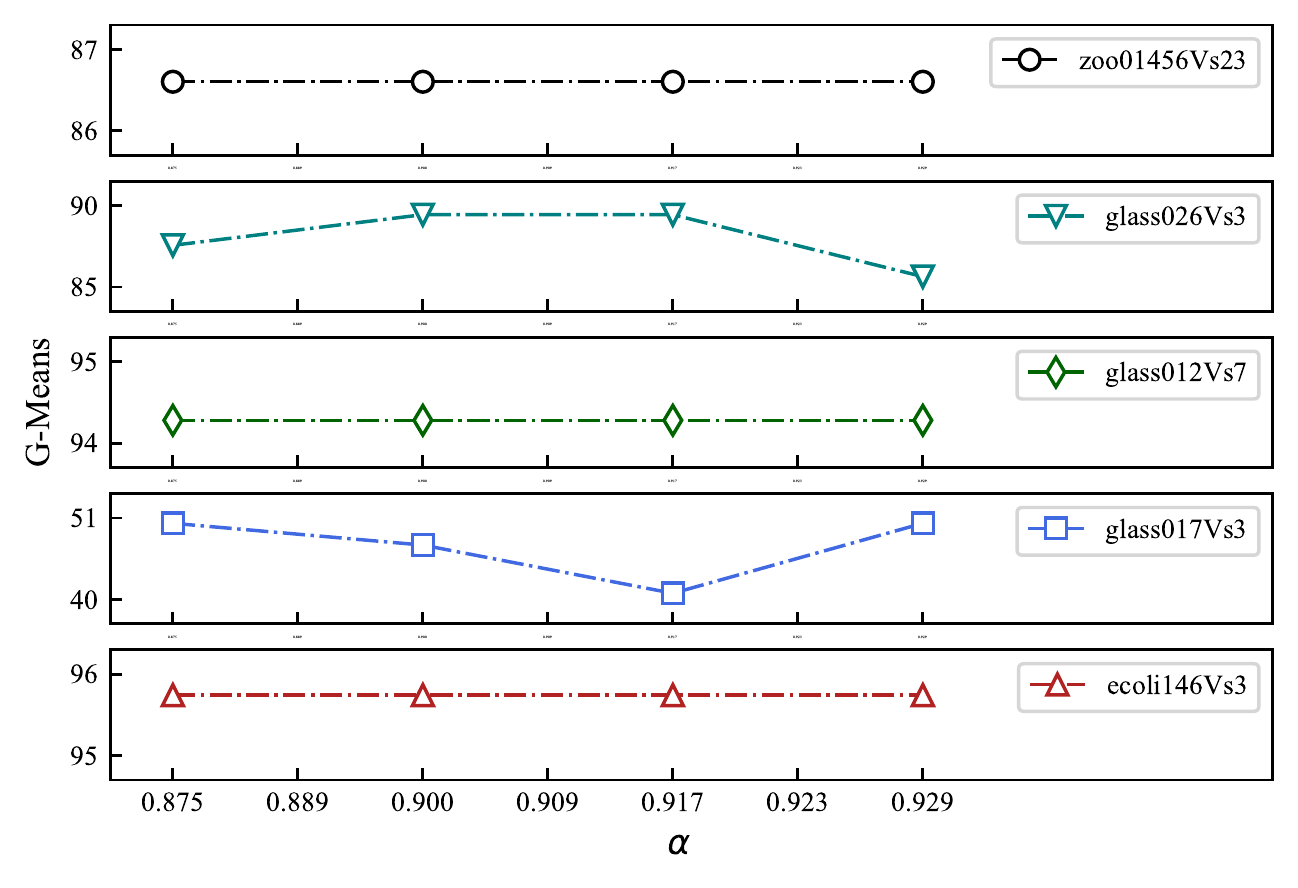}
	}
	\caption{G-Means of TWFTSVM with different $(\alpha, \beta)$}
	\label{datasetsk}
\end{figure}

\subsubsection{Discussion on $(\alpha, \beta)$}

When discussing $(\alpha, \beta)$, we selected the best G-Means of TWFTSVM at different $(\alpha, \beta)$ in the same dataset. \autoref{datasetsk} shows experimental results of different imbalanced datasets in detail. It can be seen that the G-Means of TWFTSVM has significant change on most datasets with different $(\alpha, \beta)$, which demonstrates that different $(\alpha, \beta)$ can influence the classification performance of TWFTSVM. On most datasets, the optimal value of parameter $(\alpha, \beta)$ is located in the set of $\{(0.875, 0.125), (0.900, 0.100), (0.917, 0.083), (0.929, 0.071)\}$. For TWFTSVM, choosing appropriate $(\alpha, \beta)$ is essential.

\begin{figure}[htbp]\footnotesize
	\centering
	\vspace{-25pt}  
	\subfigure[ecoli1Vs2]{
		\includegraphics[width=5.9cm, trim=70 20 30 20,clip]{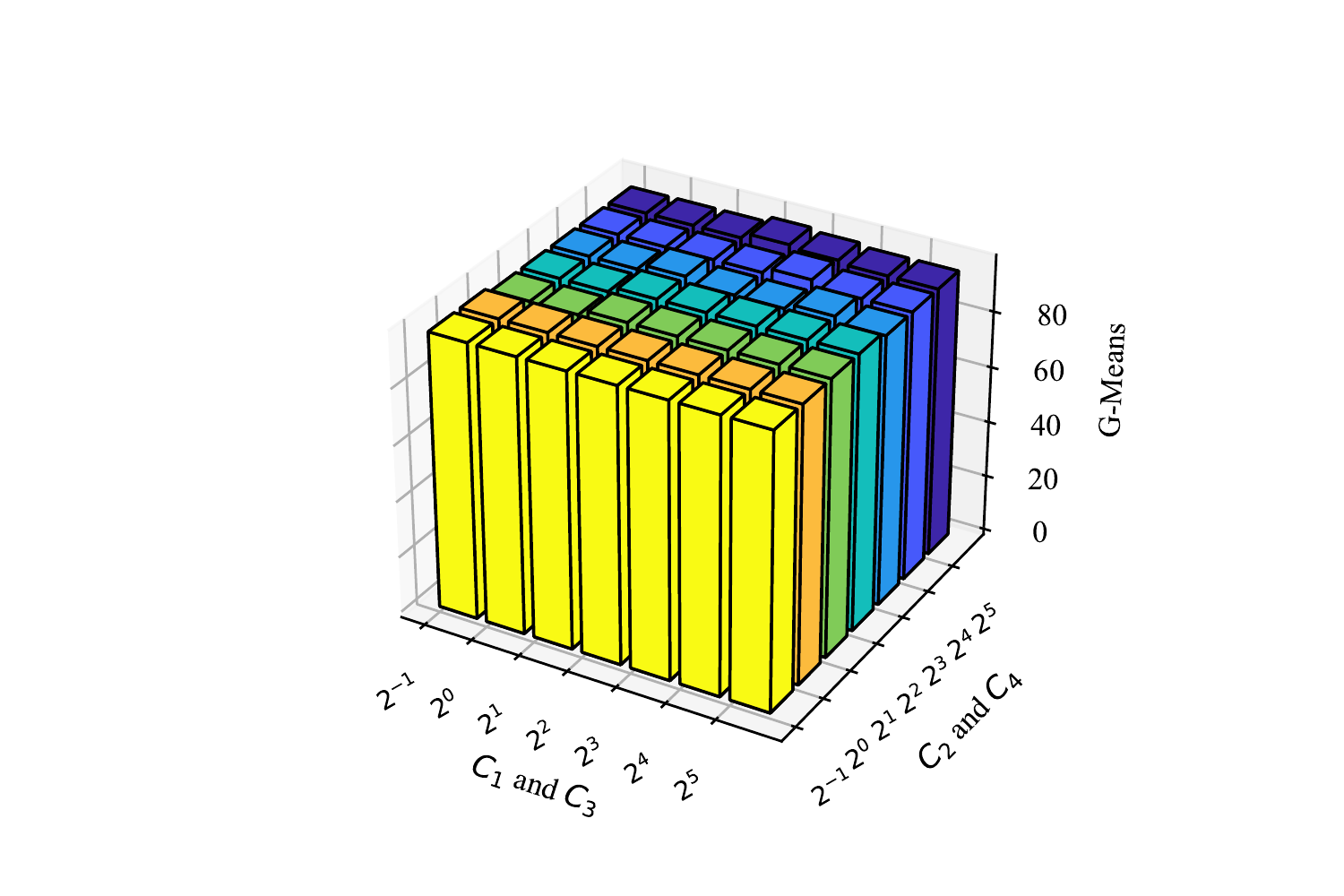}\label{ecoli1Vs2}
	}
	\hskip -30pt 
	\subfigure[ecoli1Vs6]{
		\includegraphics[width=5.9cm, trim=70 20 30 20,clip]{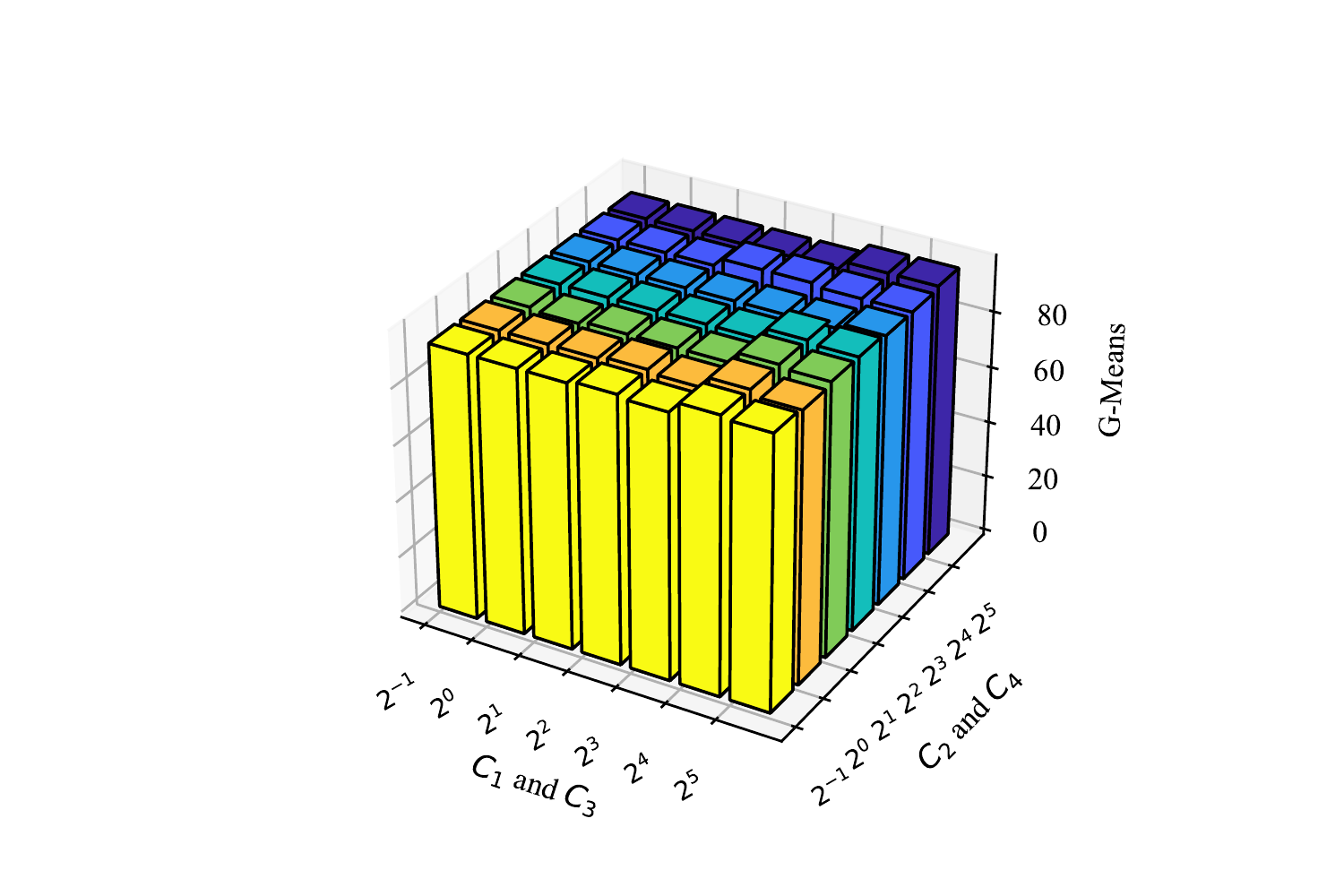}\label{ecoli1Vs6}
	}
	\hskip -30pt 
	\subfigure[glass2]{
		\includegraphics[width=5.9cm, trim=70 20 30 20,clip]{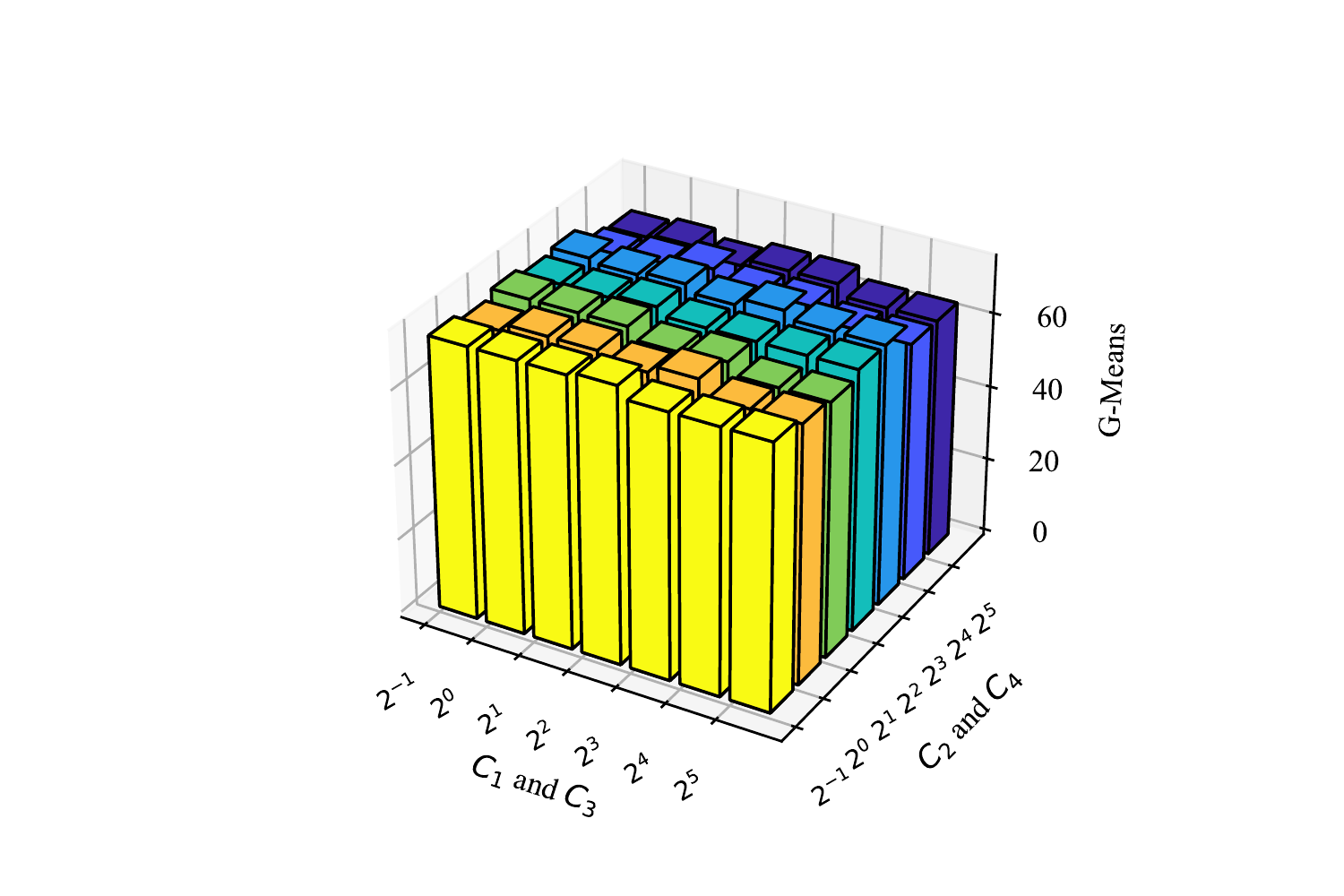}\label{glass2}
	}
	\vskip -15pt  
	\subfigure[glass6]{
		\includegraphics[width=5.9cm, trim=70 20 30 20,clip]{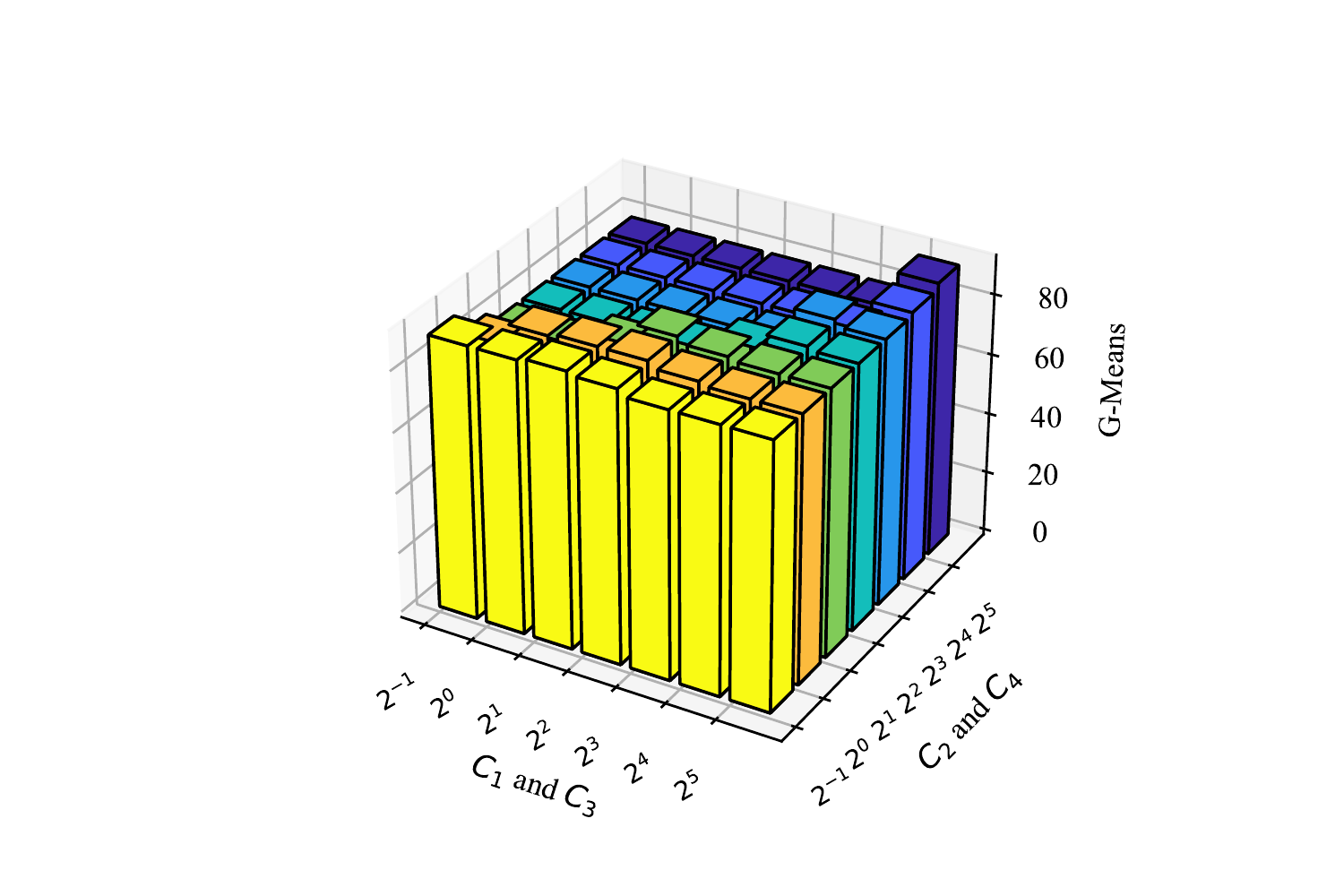}\label{glass6}
	}
	\hskip -30pt 
	\subfigure[glass015Vs7]{
		\includegraphics[width=5.9cm, trim=70 20 30 20,clip]{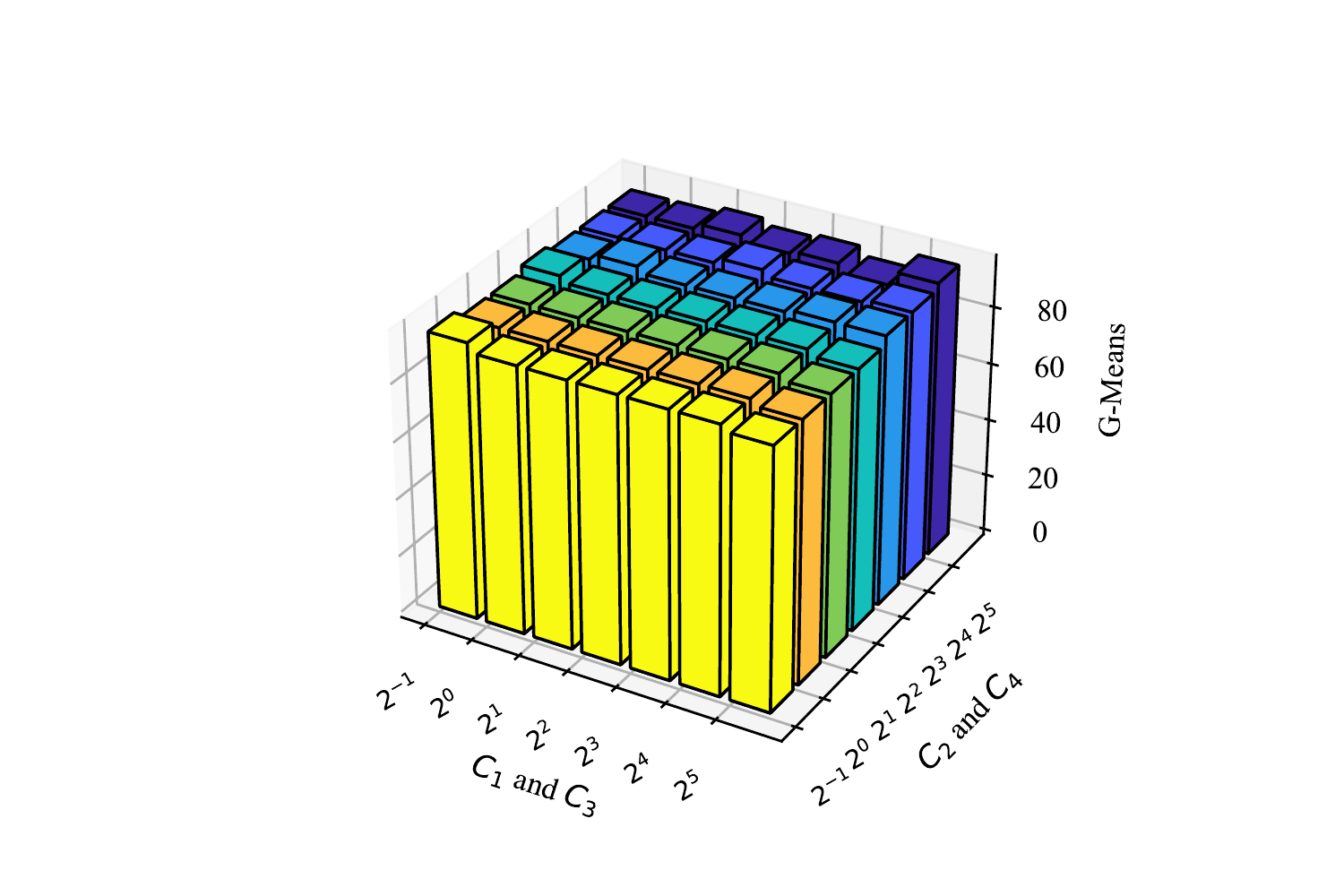}\label{glass015Vs7}
	}
	\hskip -30pt 
	\subfigure[haberman]{
		\includegraphics[width=5.9cm, trim=70 20 30 20,clip]{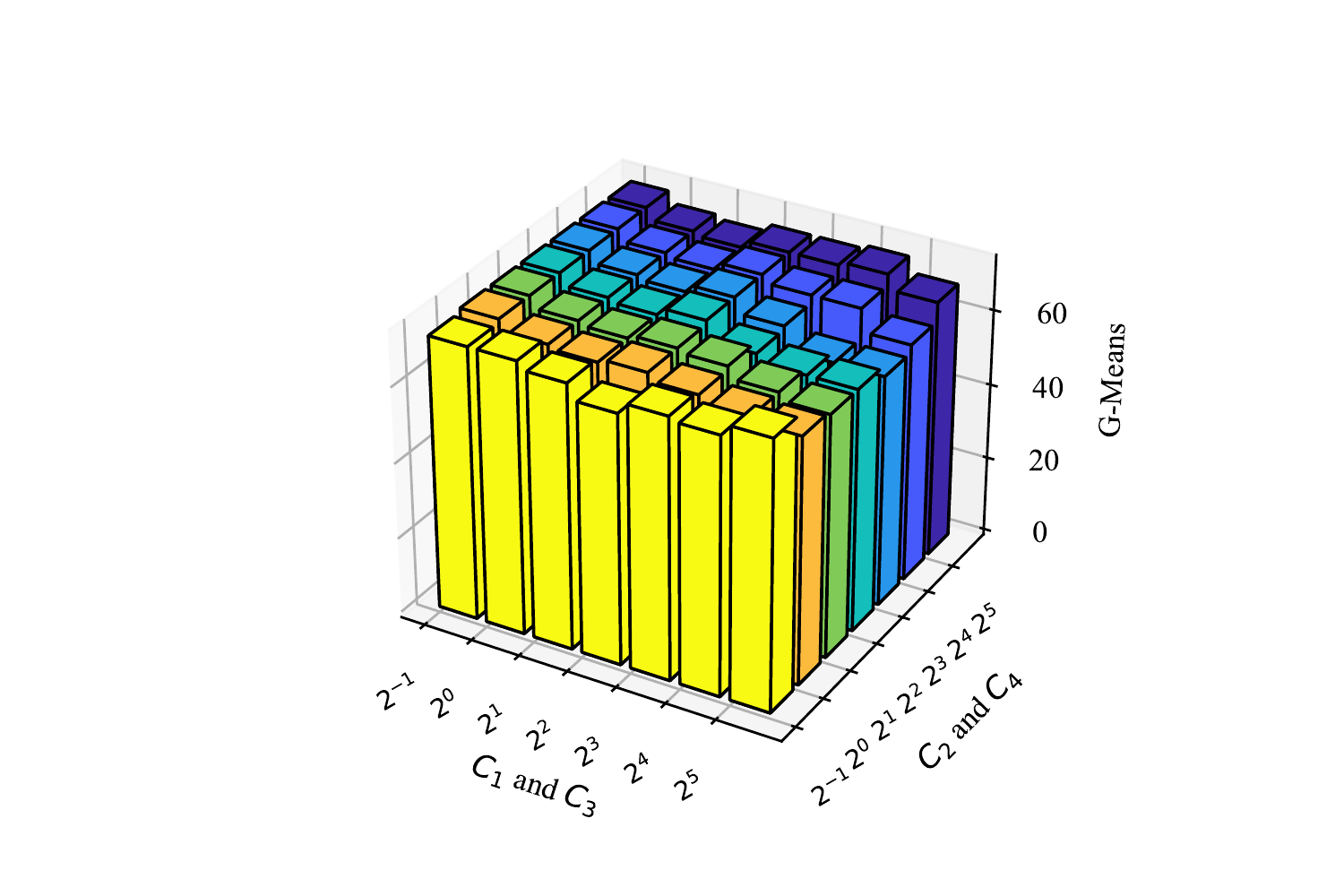}\label{haberman}
	}
	\vskip -15pt  
	\subfigure[heart-statlog]{
		\includegraphics[width=5.9cm, trim=70 20 30 20,clip]{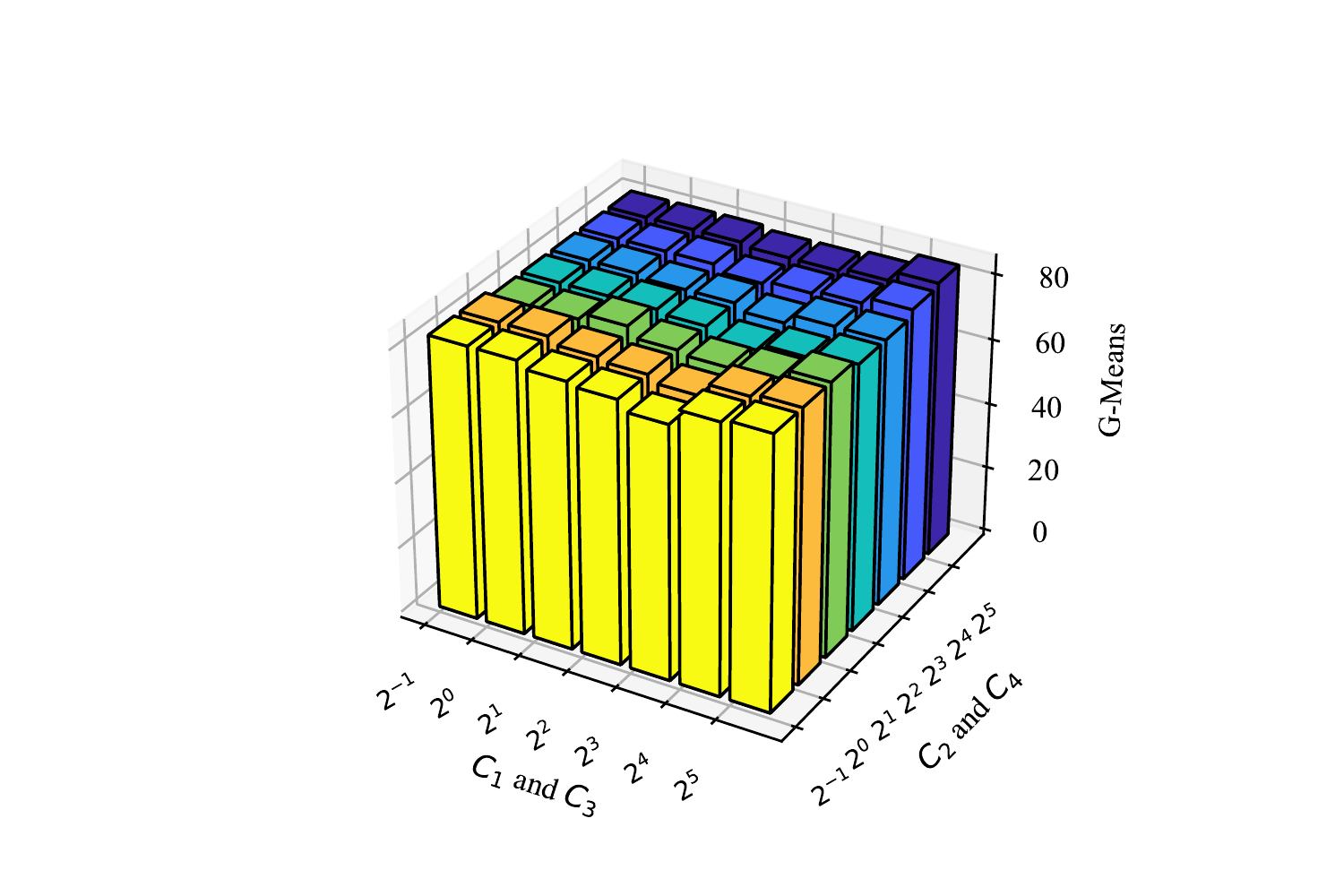}\label{heart-statlog}
	}
	\hskip -30pt 
	\subfigure[pima]{
		\includegraphics[width=5.9cm, trim=70 20 30 20,clip]{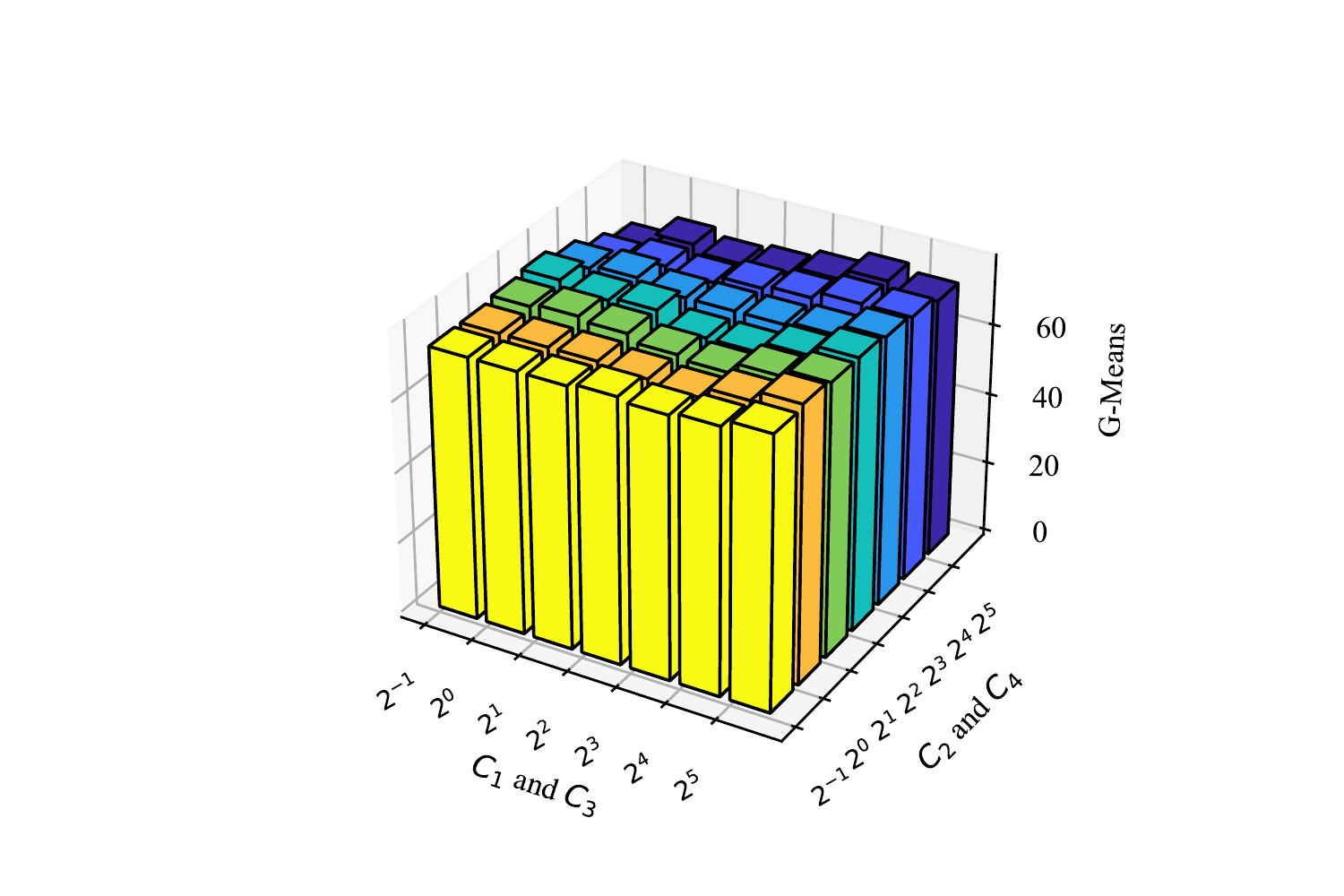}\label{pima}
	}
	\hskip -30pt 
	\subfigure[vehicle1]{
		\includegraphics[width=5.9cm, trim=70 20 30 20,clip]{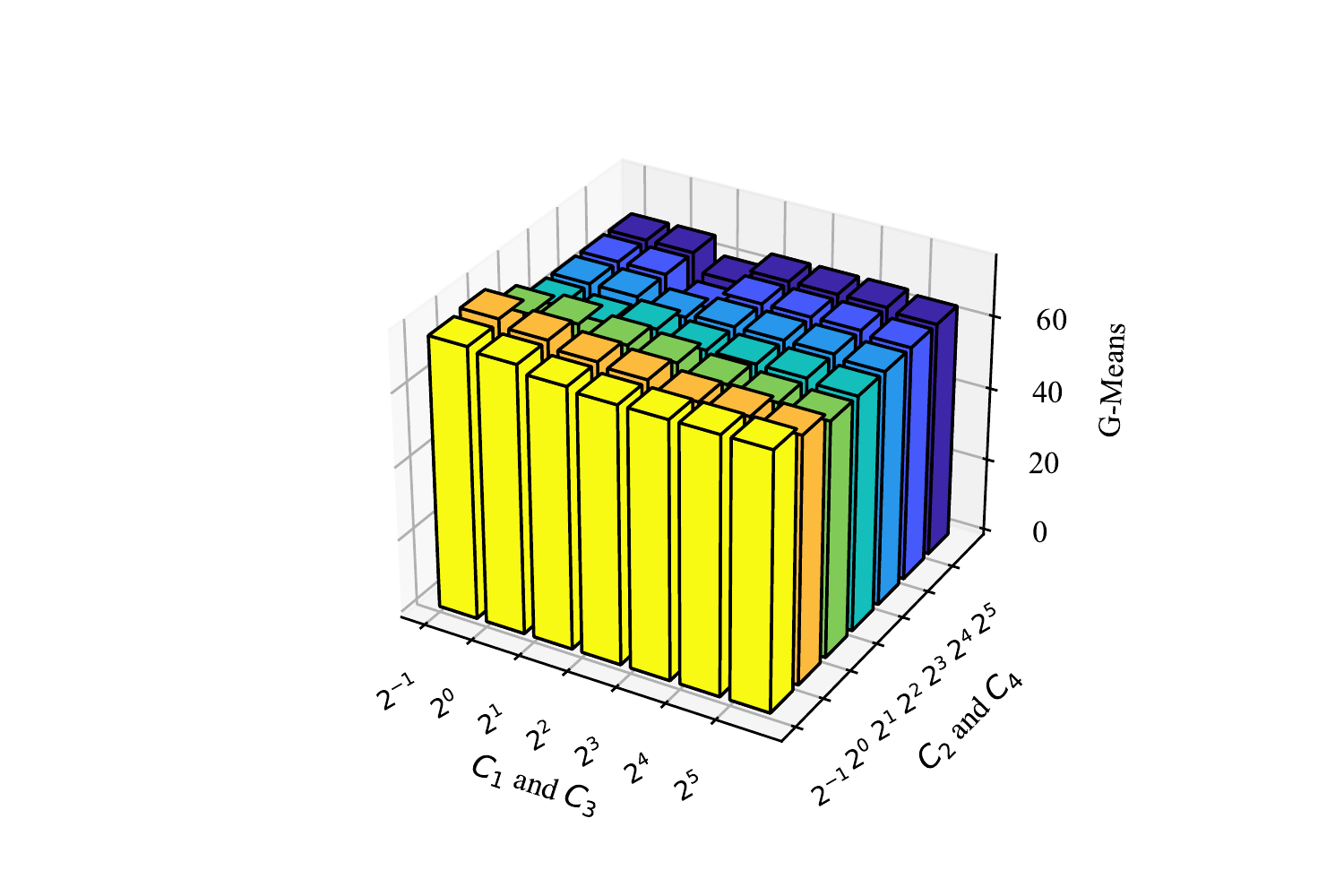}\label{vehicle1}
	}
	\vskip -15pt 
	\subfigure[vehicle2]{
		\includegraphics[width=5.9cm, trim=70 20 30 20,clip]{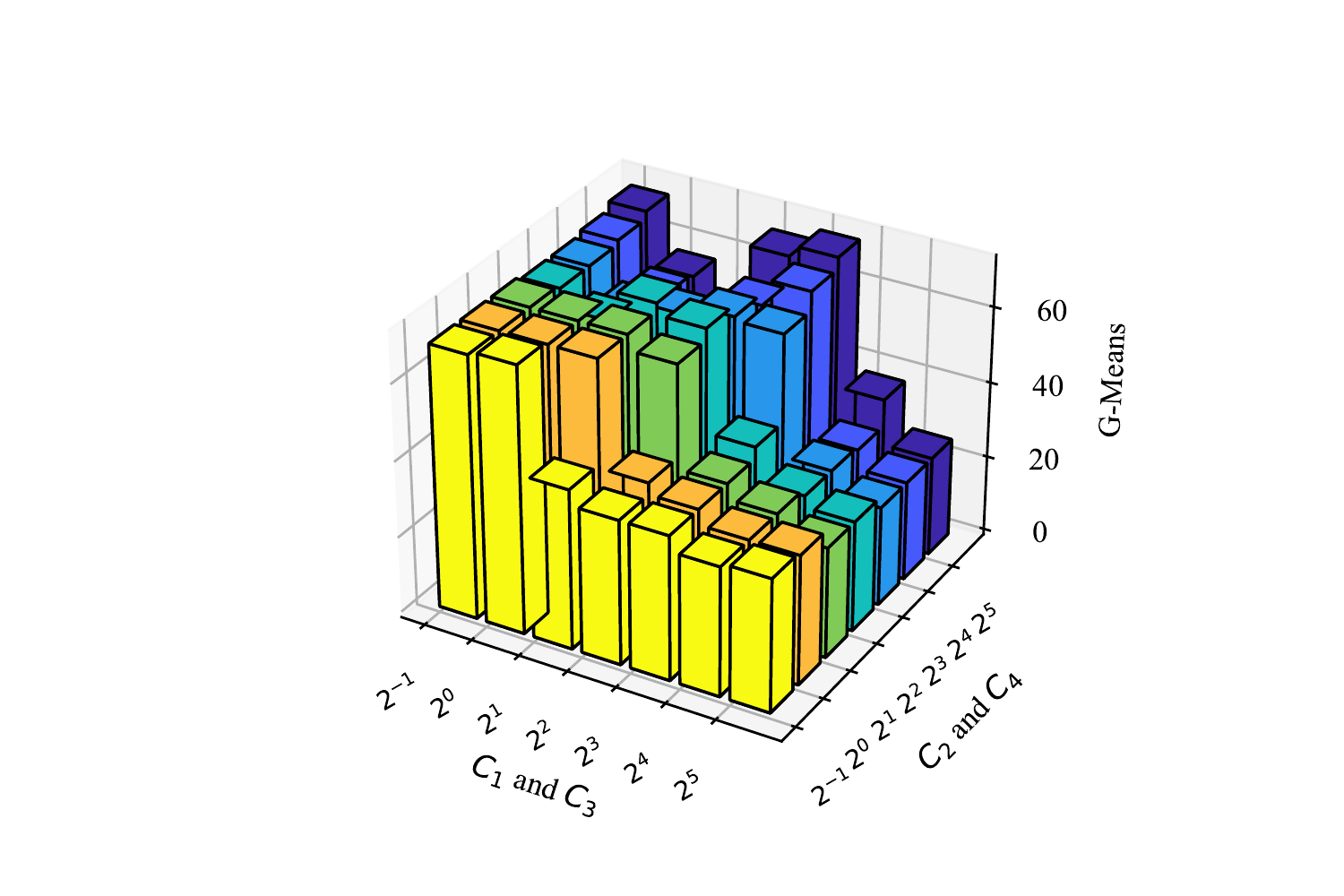}\label{vehicle2}
	}
	\hskip -30pt  
	\subfigure[wine1]{
		\includegraphics[width=5.9cm, trim=70 20 30 20,clip]{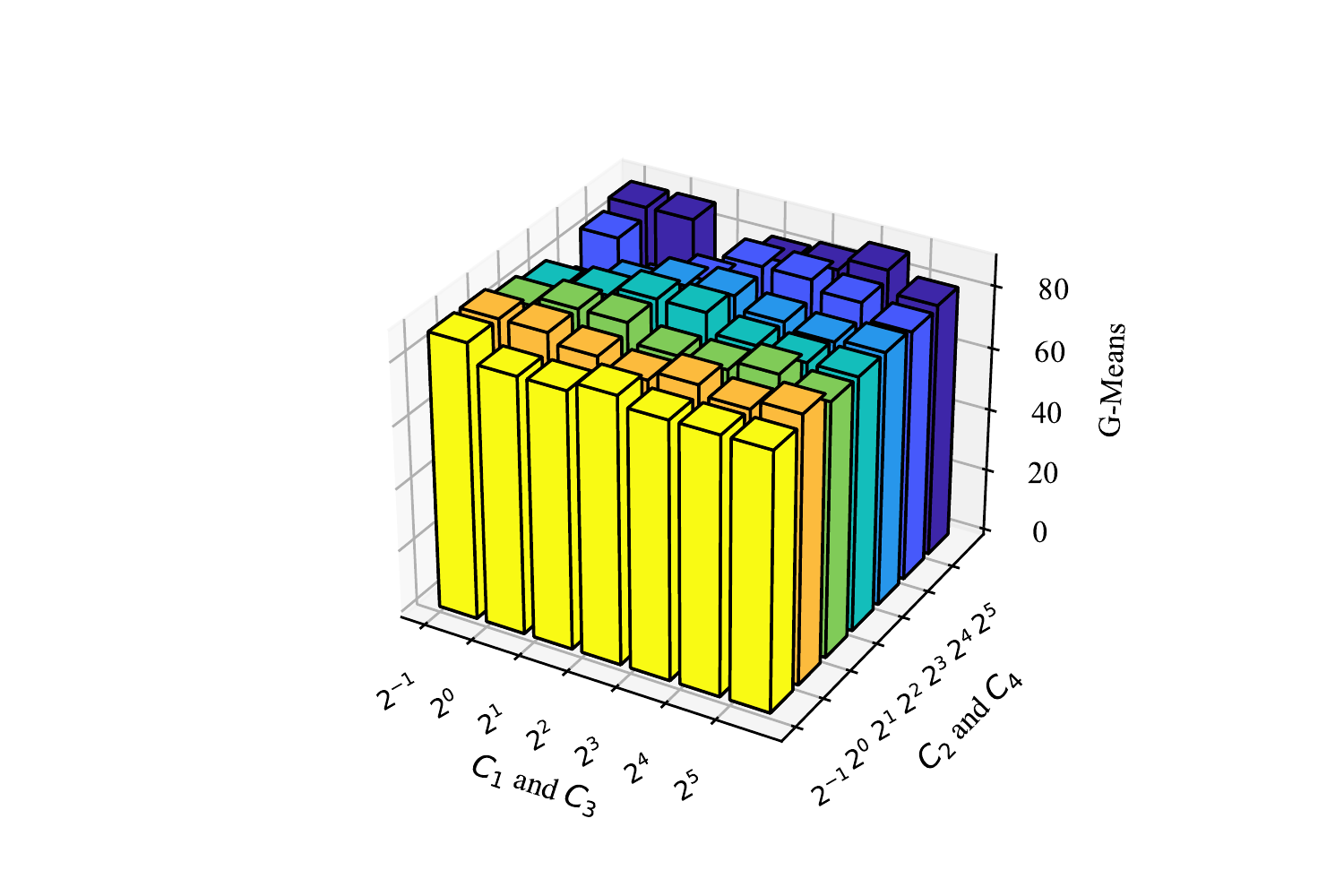}\label{wine1}
	}
	\hskip -30pt 
	\subfigure[zoo01456Vs23]{
		\includegraphics[width=5.9cm, trim=70 20 30 20,clip]{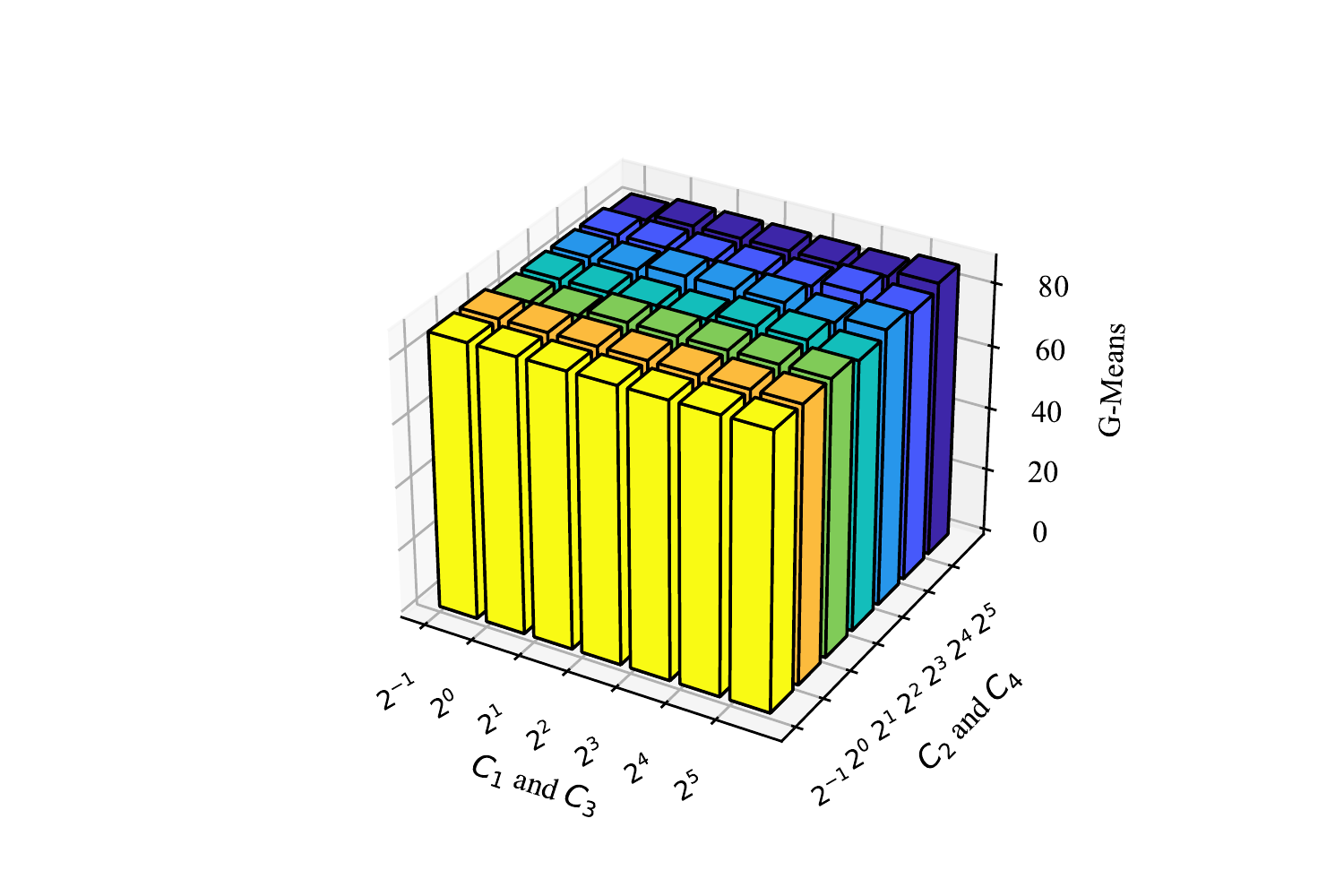}\label{zoo01456Vs23}
	}
	\caption{G-Means of TWFTSVM with different $C_1, C_3$ and $C_2, C_4$}
	\label{datasetc}
\end{figure}

\subsubsection{Discussion on $C_{1}$, $C_3$ and $C_{2}$, $C_4$}

When discussing on $C_{1}, C_{3}$ and $C_{2}, C_{4}$, we selected the best G-Means of TWFTSVM at different $C_{1}, C_{3}$ and $C_{2}, C_{4}$ in same dataset. \autoref{datasetc} shows the experimental results of different imbalanced datasets in detail. Due to space constraints, we have selected half of 30 imbalanced datasets. It can be seen that the G-Means of TWFTSVM has significant change on most datasets with different $C_{1}, C_{3}$ when $C_{2}, C_{4}$ is fixed, and G-Means of TWFTSVM has significant change on most datasets with different $C_{2}, C_{4}$ when $C_{1}, C_{3}$ is fixed. G-Means can be influenced by different $C_{1}, C_{3}$ and $C_{2}, C_{4}$. On most datasets, the best $C_{1}, C_{3}$ is located in the set of $\{2^{-1}, 2^{1}, \dots , 2^{4}, 2^{5}\}$, same as $C_1$ and $C_3$. For TWFTSVM, it is essential to choose appropriate $C_{1}, C_{3}$ and $C_{2}, C_{4}$.

\section{Conclusions and further study }\label{Section5}

In order to develop the research field of three-way decision, this paper combines three-way decision with SVM in machine learning. A new fuzzy twin SVM with three-way decision is proposed to solve CIL problems that classical SVM could not be better. A new three-way fuzzy membership function to assign higher fuzzy membership to positive samples and lower fuzzy membership to negative samples. In practice, the function assigns 1 as fuzzy membership for positive samples and fuzzy membership for negative samples to ensure that positive samples are more important. Then, the function is applied in TWFTSVM, which combines FSVM and TSVM to solve imbalanced problems. The effectiveness of TWFTSVM is evaluated in low, medium, and high IR datasets and different IR datasets derived from the same datasets. Experimental results demonstrate that TWFTSVM outperforms compared algorithms on real-world imbalanced datasets and the same datasets with different IRs. Our future work is focused on optimizing the algorithm for solving TWFTSVM to reduce running time and enhance the structure of TWFTSVM to solve binary classification problems with higher IRs.

\section*{Acknowledgments}
We confirm that there are no known conflicts of interest associated with this publication and our financial support will not affect research outcome.

\vfill


\begin{thebibliography}{10}
	
	\bibitem{Yao2010}
	Y.~Y. Yao.
	\newblock{Three-way decisions with probabilistic rough sets}.
	\newblock{\em Information Sciences}, 180 341-353, 2010.
	
	\bibitem{Yao2016}
	Y.~Y. Yao.
	\newblock{Three-Way Decisions and Cognitive Computing}.
	\newblock{\em Cognitive Computation}, 8 (4) 543-554, 2016.

	\bibitem{Liang2015}
	D.~C. Liang, D. Liu.
	\newblock{Deriving three-way decisions from intuitionistic fuzzy decision-theoretic rough sets}.
	\newblock{\em Information Sciences}, 300 28-48, 2015.

	\bibitem{Zhao2016}
	X.~R. Zhao, B.~Q. Hu.
	\newblock{Fuzzy probabilistic rough sets and their corresponding three-way decisions}.
	\newblock{\em Knowledge-Based Systems}, 91 126-142, 2016.

	\bibitem{Zhang2018}
	Q.~H. Zhang, Q. Xie, G.~Y. Wang.
	\newblock{A Novel Three-way decision model with decision-theoretic rough sets using utility theory}.
	\newblock{\em Knowledge-Based Systems}, 159 321-335, 2018.
	
	\bibitem{Deng2022}
	J. Deng, J.~M Zhan, W.~P Ding, et al. 
	\newblock{A novel prospect-theory-based three-way decision methodology in multi-scale information systems}. \newblock{\em Artificial Intelligence Review}, 2022, https://doi.org/10.1007/s10462-022-10339-6.
	
	\bibitem{Su2023}
	J. Tu, S.~H. Su.
	\newblock{Method for three-way decisions using similarity in incomplete information systems}.
	\newblock{\em International Journal of Machine Learning and Cybernetics}, 2023, https://doi.org/10.1007/s13042-022-01745-x.
	
	\bibitem{Hu2014}
	B.~Q. Hu.
	\newblock{Three-way decisions space and three-way decisions}.
	\newblock{\em Information Sciences}, 281 21-52, 2014.
	
	\bibitem{Liu2013}
	D. Liu, T.~R. Li, H.~X. Li.
	\newblock{Rough set theory: a three-way decisions perspective}. 
	\newblock{\em Journal of Nanjing University: Natural Sciences}, 49(5) 574-581, 2014.
	
	\bibitem{Liu 2022}
	D. Liu, Q.~X. Chen.
	\newblock{A novel three-way decision model with DEA method}.
	\newblock{\em International Journal of Approximate Reasoning}, 148 23-40, 2022.

	\bibitem{Vapnik1995}
	V.~N. Vapnik.
	\newblock {The Nature of Statistical Learning Theory}.
	\newblock {\em Springer-Verlag}, 1995.

	\bibitem{Sun2021}
	G.~Y. Sun, X.~Q. Rong, A.~Z. Zhang, et al.
	\newblock{Multi-scale mahalanobis kernel-based support vector machine for classification of high-resolution remote sensing images}.
	\newblock{\em Cognitive Computation}, 13 (3) 787-794, 2021.

	\bibitem{Mishra2021}
	S.~K. Mishra, V.~H. Deepthi.
	\newblock{Brain image classification by the combination of different wavelet transforms and support vector machine classification}.
	\newblock{\em Journal of Ambient Intelligence and Humanized Computing}, 12 (6) 6741-6749, 2021.

	\bibitem{Do2009}
	H. Do, A. Kalousis, M. Hilario.
	\newblock{Feature weighting using margin and radius based error bound optimization in SVMs}.
	\newblock{\em  Machine Learning and Knowledge Discovery in Databases}, 315-329, 2009.

	\bibitem{Shia2021}
	W.~C. Shia, L.~S. Lin, D.~R. Chen.
	\newblock{Classification of malignant tumors in breast ultrasound using a pretrained deep residual network model and support vector machine}.
	\newblock{\em Computerized Medical Imaging and Graphics}, 87 101-829, 2021.
	
	\bibitem{Wang2020}
	J. Wang, L. Wu, H. Wang, et al.
	\newblock{An efficient and privacy-preserving outsourced support vector machine training for internet of medical things}.
	\newblock{\em IEEE Internet of Things Journal}, 8 (1) 458-473, 2021.
	
	\bibitem{Sun2009}
	Y. Sun, A. Wong, M. Kamel.
	\newblock{Classification of imbalanced data: a review}.
	\newblock{\em International Jouanal of Pattern Recognition and Artificial Intelligence}, 23 (4) 687-719, 2009.
	
	\bibitem{Newby2013}
	D. Newby, A. Freitas, T. Ghafourian.
	\newblock{Coping with unbalanced class data sets in oral absorption models}.
	\newblock{\em Journal of Chemical Information and Modeling}, 53 (2) 461-474, 2013.

	\bibitem{Zhang2016}
	Y. Zhang, J. Zhang, Z. Pan, et al.
	\newblock{Multi-view dimensionality reduction via canonical random correlation analysis}.
	\newblock{\em Frontiers of Computer Science}, 10 (5) 856-869, 2016.
	
	\bibitem{Hupont2019}
	I. Hupont, C. Fernández.
	\newblock{DemogPairs: quantifying the impact of demographic imbalance in deep face recognition}.
	\newblock{\em In: 2019 14th IEEE International Conference on Automatic Face and Gesture Recognition}, Lille in France, 1-7, 2019.
	
	\bibitem{Wang2016}
	Y. Wang, H. Wang, J. Li, et al.
	\newblock{Efficient graph similarity join for information
		integration on graphs}.
	\newblock{\em Frontiers of Computer Science}, 10 (2) 317-329, 2016.
	
	\bibitem{Batista2004}
	G. Batista, R.~C. Prati, M.~C. Monard.
	\newblock{A study of the behavior of several methods for balancing machine learning training data}.
	\newblock{\em In: ACM SIGKDD Explorations Newsletter}, New York in USA, 6 (1) 20-29, 2004.
	
	\bibitem{Yen2009}
	S.~J. Yen, Y.~S. Lee.
	\newblock{Cluster-based under-sampling approaches for imbalanced data distributions}.
	\newblock{\em Expert Systems and Applications}, 36 (3) 5718-5727, 2009.
	
	\bibitem{Bowyer2002}
	K. Bowyer, N. Chawla, L. Hall, et al.
	\newblock{SMOTE: synthetic minority over-sampling technique}.
	\newblock{\em Journal of Artificial Intelligence Research}, 16 (1) 321-357, 2002.
	
	\bibitem{Zong2010}
	Z.~H. Zong, X.~Y. Liu.
	\newblock{On multi-class cost sensitive learning}.
	\newblock{\em Computational Intelligence}, 26 (3) 232-257, 2010.
	
	\bibitem{Cheng2016}
	F. Cheng, J. Zhang, C. Wen.
	\newblock{Cost-sensitive large margin distribution machine for classification of imbalanced data}.
	\newblock{\em Pattern Recognition Letters}, 80 107-112, 2016.
	
	\bibitem{Chawla2003}
	N. Chawla, A. Lazarevic, L.~O. Hall, et al.
	\newblock{SMOTEBoost: improving prediction of the minority class in boosting}.
	\newblock{\em In: Proceedings of 7th European Conference on Principles and Practice of Knowledge Discovery in Databases}, Heidelberg in Berlin, 107-119, 2003.
	
	\bibitem{Seiffrt2010}
	C. Seiffrt, T.~M. Khoshgoftaar, H.~J. Van, et al.
	\newblock{RUSBoost: a hybrid approach to alleviating class imbalance}. 
	\newblock{\em IEEE Transactions on Systems, Man, and Cybernetics, Part A (Systems and Humans)}, 40 (1) 185-197, 2010.
	
	\bibitem{Liu2009}
	X.~Y. Liu, J. Wu, Z.~H. Zhou.
	\newblock{Exploratory undersampling for class-imbalance learning}.
	\newblock{\em IEEE Transactions on Systems, Man, and Cybernetics, Part B (Cybernetics)}, 39 (2) 539-550, 2009.
	
	\bibitem{Freund1997}
	Y. Freund, R.~E. Schapire.
	\newblock{A decision-theoretic generalization of on-line learning and an application to boosting}.
	\newblock{\em Journal of Computer and System Sciences}, 55 (1) 119-139, 1997.
	
	\bibitem{TD2016}
	P. Xu, F. Davoine, H.~B. Zha, T. Denoeux,
	\newblock{Evidential calibration of binary SVM classifiers}.
	\newblock{\em International Journal of Approximate Reasoning}, 72 55-70, 2016.
	
	\bibitem{Lin2002}
	C.~F. Lin, S.~D. Wang.
	\newblock{Fuzzy support vector machines}.
	\newblock{\em IEEE Transactions on Neural Networks}, 13 (2) 464-471, 2002.
	
	\bibitem{Zhou2009}
	M.~M. Zhou, L. Li, Y.~L. Lu.
	\newblock{Fuzzy support vector machine based on density with dual membership}.
	\newblock{\em In: 2009 International Conference on Machine Learning and Cybernetics}, Baoding in China, 674-678, 2009.
	
	\bibitem{Ha2013}
	M.~H. Ha, C. Wang, J.~Q. Chen.
	\newblock{The support vector machine based on intuitionistic fuzzy
		number and kernel function}.
	\newblock{\em Soft Computing}, 17 (4) 635-641, 2013.
	
	\bibitem{Yang2011}
	X. Yang, G. Zhang, J. Lu, et al.
	\newblock{A kernel fuzzy c-means clustering-based fuzzy support vector machine algorithm for classification problems with outliers or noises}.
	\newblock{\em IEEE Transactions on Fuzzy Systems}, 19 (1) 105-115, 2011.
	
	\bibitem{Hang2016}
	J. Hang, J. Zhang, M. Cheng.
	\newblock{Application of multi-class fuzzy support vector machine classifier for fault diagnosis of wind turbine}.
	\newblock{\em Fuzzy Sets Systems}, 297 128-140, 2016.
	
	\bibitem{Wu2014}
	Z. Wu, H. Zhang, J. Liu.
	\newblock{A fuzzy support vector machine algorithm for classification based on a novel PIM fuzzy clustering method}.
	\newblock{\em Neurocomputing}, 125 119-124, 2014.
	
	\bibitem{Wang2005}
	Y. Wang, S. Wang, K.~K. Lai. 
	\newblock{A new fuzzy support vector machine to evaluate credit risk}.
	\newblock{\em IEEE Transactions on Fuzzy Systems}, 13 (6) 820-831, 2005.
	
	\bibitem{Li2015}
	S.~T. Li, C.~C. Chen.
	\newblock{A regularized monotonic fuzzy support vector machine model for data mining with prior knowledge}.
	\newblock{\em IEEE Transactions on Fuzzy Systems}, 23 (5) 1713-1727, 2015.
	
	\bibitem{An2013}
	W. An, M. Liang.
	\newblock{Fuzzy support vector machine based on within-class scatter for classification problems with outliers or noises}.
	\newblock{\em Neurocomputing}, 110 101-110, 2013.
	
	\bibitem{Jiang2006}
	X. Jiang, Z. Yi, J.~C. Lv. 
	\newblock{Fuzzy SVM with a new fuzzy membership function}.
	\newblock{\em Neural Computing and Applications}, 15 (3-4) 268-276, 2006.
	
	\bibitem{Tang2011}
	W.~M. Tang.
	\newblock{Fuzzy SVM with a new fuzzy membership function to solve the two-class problems}.
	\newblock{\em Neural Processing Letters}, 34 (3) 209-219, 2011.
	
	\bibitem{Lin2004}
	C.~F. Lin, S.~D. Wang.
	\newblock{Training algorithms for fuzzy support vector machines with noisy data}.
	\newblock{\em Pattern Recognition Letter}, 25 (14) 1647-1656, 2004.
	
	\bibitem{Wang2020CKA}
	T.~H. Wang, Y.~Z. Qiu, J.~L. Hua.
	\newblock{Centered kernel alignment inspired fuzzy support vector machine}.
	\newblock{\em Fuzzy Sets and Systems}, 394 (2) 110-123, 2020.
	
	\bibitem{Niazmardi}
	S. Niazmardi, A. Safari, S. Homayouni.
	\newblock{A novel multiple kernel learning framework for multiple feature classification}.
	\newblock{\em IEEE Journal of Selected Topics in Applied Earth Observations and Remote Sensing}, 10 (8) 3734-3743, 2017.
		
	\bibitem{Jayadeva2007}
	Jayadeva, R. Khemchandani, S. Chandra.
	\newblock{Twin support vector machines for pattern classification}.
	\newblock{\em IEEE Transactions on Pattern Analysis and Machine Intelligence}, 29 (5) 905-910, 2007.
	
	\bibitem{Shao2011}
	Y.~H. Shao, C.~H. Zhang, X.~B. Wang, et al.
	\newblock{Improvements on twin support vector machines}. 
	\newblock{\em IEEE Transactions on Neural Networks}, 22 (6) 962-968, 2011.
	
	
\end{thebibliography}
\end{document}